# *A Mass-Conserving-Perceptron for Machine-Learning-Based Modeling of Geoscientific Systems*

Yuan-Heng Wang, Hoshin V. Gupta

Department of Hydrology and Atmospheric Science, The University of Arizona, Tucson, AZ

## Abstract

Although decades of effort have been devoted to building *Physical-Conceptual* (PC) models for predicting the time-series evolution of geoscientific systems, recent work shows that *Machine Learning* (ML) based *Gated Recurrent Neural Network* technology can be used to develop models that are much more accurate. However, the difficulty of extracting physical understanding from ML-based models complicates their utility for enhancing scientific knowledge regarding system structure and function. Here, we propose a physically-interpretable *Mass Conserving Perceptron* (MCP) as a way to bridge the gap between PC-based and ML-based modeling approaches. The MCP exploits the inherent isomorphism between the directed graph structures underlying both PC models and GRNNs to explicitly represent the mass-conserving nature of physical processes while enabling the functional nature of such processes to be directly learned (in an interpretable manner) from available data using off-the-shelf ML technology. As a proof of concept, we investigate the functional expressivity (capacity) of the MCP, explore its ability to parsimoniously represent the rainfall-runoff (RR) dynamics of the *Leaf River Basin*, and demonstrate its utility for scientific hypothesis testing. To conclude, we discuss extensions of the concept to enable ML-based physical-conceptual representation of the coupled nature of *mass-energy-information flows* through geoscientific systems.

## Plain Language Summary

We develop a physically-interpretable computational unit, referred to as the *Mass-Conserving Perceptron* (MCP). Networks of such units can be used to model the conservative nature of the input-state-output dynamics of mass flows in geoscientific systems, while Machine Learning (ML) technology can be used to learn the functional nature of the physical processes governing such system behaviors. Testing using data from the *Leaf River Basin* demonstrates the considerable functional expressivity (capacity) and interpretability of even a single-MCP-node-based model, while providing excellent predictive performance and the ability to conduct scientific hypothesis testing. The concept can easily be extended to facilitate ML-based physical-conceptual representation of the coupled nature of *mass-energy-information flows* through geoscientific systems, thereby facilitating the development of synergistic physics-AI modeling approaches.

## Key Points:

- We develop a physically-interpretable unit (Mass-Conserving Perceptron) that can be used as a basic component of geoscientific models.
- Off-the-shelf Machine Learning technology can be used to learn the functional nature of the physical processes governing system behaviors.
- The concept can be extended to facilitate ML-based representation of coupled mass-energy-information flows in geoscientific systems.

## 1. Introduction and Scope

### 1.1 Physical-Conceptual Modeling of the Rainfall-Runoff System

[1] There is a long history of developing computer-based physical-conceptual (PC) catchment-scale rainfall runoff (RR) models in hydrology (e.g., *Burnash et al., 1973; Singh 1988; Boyle 2000; Perrin et al., 2003; Fenicia et al., 2006; Aghakouchak & Habib, 2010*). Such models are constructed based first on a) *perceptual-conceptual understanding* of the nature of the natural system, followed by b) explicit formalization of those perceptions into a *conceptual model*, then progressively translated into c) a *symbolic/mathematical model,* and finally d) a computational model (an algorithm, implemented in computational language, as a computer program) that can be executed on a digital computer. Generally, such RR model development is based on a progressive set of fundamental scientific hypotheses, including the *Conservation Law Hypothesis* ($H_{CL}$), *System Architecture Hypothesis* ($H_{SA}$), *Process Parameterization Hypothesis* ($H_{PP}$), and *Parameter Specification Hypothesis* ($H_{PS}$) (*Gupta et al. 2012*), and may also include other hypotheses that pertain to things outside of the modeled system (see summary in *Gharari et al., 2021*).

[2] In this regard, it may be important to distinguish between a *"Physics-based"* model and a *"Physical-Conceptual"* model. Strictly speaking, a minimum requirement for a model to be called *"physics-based"*, is that its fluxes are driven by potential gradients and perform work against resistances (see definition in *Zehe et al., 2014*). Such representation is not made explicit in *"physical-conceptual"* models of non-linear systems despite the fact that all components of the water balance, including runoff and storage dynamics, do in fact integrate the mass, momentum and energy balance. In practice, such conceptual models reduce the water balance to a changes in mass balance of filling-and-draining buckets, and do not explicitly incorporate representation of the driving gradients.

[3] One aspect of the development of catchment-scale RR models is the attention given to a "*top-down*" approach (*Dooge, 1986; Sivapalan et al., 2003; Young, 2003*) wherein, instead of emphasizing detail in the process representations (characteristic of the "*bottom-up*" approach), the focus is on developing "*efficient*" or "*parsimonious*" model representations, which can be as simple as a linear transfer function that is capable of reproducing key signatures or features of observed hydrological variability (*Sivapalan & Young, 2006; Hrachowitz & Clark, 2017*). However, it is useful to follow the formal steps in model building mentioned above, and represent the system via a directed graph architecture consisting of node and link subcomponents (*Gupta and Nearing, 2014*). This enables a primary focus on the physical principles underlying the dynamics of the system (e.g., conservation and thermodynamic laws) and an appropriate choice of system architecture (selection of state variables and their interconnectedness), while leaving the forms of the process parameterization equations (that determine the state transitions and fluxes) and their associated parameter values to be inferred from the data (*Gharari et al., 2021*).

[4] Occam's Razor (also referred to as the principle of parsimony) is often used to justify top-down modeling approaches, and the hydrological literature has typically emphasized the need for parameter identifiability and interpretability (*Lute & Luce, 2017; Weijs & Ruddell, 2020*). As a consequence, one might argue that PC hydrological models are typically "*lossy over-compressions*" of the available relevant information, consisting of simplified representations that are (therefore) unable to provide the level of predictive performance that might otherwise be possible.

[5] Perhaps in recognition of this, there has been a trend toward introducing more flexibility into the modeling approach. To name just a few, the ''*Framework for Understanding Structural Errors*'' (FUSE) approach explores the use of multiple system parameterizations within a single modeling framework, wherein different models can be viewed as simplifications within a more general architecture (*Clark et al., 2008*), while the ''*Flexible Modeling Approach*'' (FLEX) framework (*Fenicia et al., 2008*) also explores the use of flexible system architectures and parameterizations based on alternative combinations of universal system subcomponents. Such (flexible) representational frameworks enable the method of multiple working hypotheses when describing both individual processes and the overall system (*Clark et al., 2011; Fenicia et al., 2011;* also see

review in *Gharari et al., 2021*), thereby facilitating the examination of multi-model structural adequacy across different locations (*Knoben et al., 2020*).

[6] Meanwhile, recognizing that the problem of developing an "*optimal*" representation of the RR process is inherently multi-objective (*Gupta et al., 1998*), a significant body of literature has sought to identify and extract diagnostic signatures (features) from the system response data (*Gupta et al., 2008; Martinez & Gupta, 2010; Yilmaz et al., 2008; McMillan et al., 2017; McMillan, 2020*). Akin to feature engineering, this strategy seeks to improve model identifiability, guide the specification of parameter values, and help in determining the representational complexity of parameterized processes. Importantly, that research thread explicitly acknowledges structural uncertainty, and has included explorations of Bayesian data assimilation for learning the dynamical input-state-output structure of simple water balance models (*Bulygina & Gupta, 2009*) and lumped catchment-scale RR models (*Bulygina & Gupta, 2011; Nearing & Gupta, 2015*).

[7] More recently, *De la Fuente et al.* (*2023b*) proposed that a multi-representational (as opposed to multi-hypothesis) approach can assist in the process of understanding model structural limitations, and thereby guide the subjective discovery process necessary to achieve model structural improvement. In this regard, it is important to recognize that equations and algorithms are complementary forms of representation, each with their own strengths and limitations (*Gaucherel et al., 2011*), which reflects the reality that there is no fundamentally "*correct*" single representational system. This allows the scientist to frame questions from different perspectives, thereby potentially enriching the interpretations that can be arrived at. As pointed out by a reviewer of this manuscript, Dr. Zehe, one might even stress that reality is always "*sensed*". This arguably is why mankind defined the empirical sciences, at their forefront physics, which rely fundamentally on reproducibility (within uncertainty) of observations/measurements. The measurement devices and gauges to collect reproducible data essentially rely on the validity of conservation laws.

## 1.2 ML-Based Modeling of the Rainfall-Runoff Modeling System

[8] Modern Machine Learning (ML) methods, and especially Deep Learning (DL), bring tremendous versatility to the modeling process (*Kratzert et al., 2018; Shen, 2018; Shen et al., 2018; Reichstein et al., 2019*), and have been shown to achieve unprecedented accuracy in the context of a variety of modelling tasks (see review by *Höge et al., 2022*). The ability of neural network-based (NN) approaches to provide better predictive performance than both traditional data-driven regression-based methods and PC-based modeling has been known since at least the end of last century (*Hsu et al. 1995*), at least in the context of RR modeling. That early work also explored the use of simple recurrent neural networks (RNNs) to emulate system memory and state dynamics (*Hsu et al. 1997*). Recently, however, with the development of gated RNN's, and in particular the long-short-term memory (LSTM) network (*Hochreiter and Schmidhuber, 1997*), the ability of ML to advance the modeling of dynamical hydrological processes has been dramatically demonstrated, not only for catchment-scale RR modeling (*Kratzert et al., 2018; Kratzert et al., 2019; Feng et al., 2020; Lees et al., 2021; Arsenault et al., 2023*), but also for snowpack modeling (*Wang et al., 2022; Duan et al., 2023*), and in many other contexts (*Than et al., 2021; Zhi et al., 2023*) that are relevant to water resource management, such as addressing the potential impacts of changing climate (*Sungmin et al., 2020*).

[9] However, concerns have been raised about the physical interpretability of ML-based models, and considerable attention is now being devoted to addressing this issue (*Montavon et al., 2018; Guidotti et al., 2018; Samek et al., 2019; Molnar et al., 2020; Molnar, 2021*); see also *Fleming et al.,* (*2021*) and *McGovern et al.,* (*2019*) in the hydrological and meteorological contexts respectively. In particular, RR modelers are now exploring strategies for embedding PC principles into ML-based approaches with a view to enhancing the "*geoscientific awareness*" (*Jiang et al., 2020*) of data-driven modeling, and thereby providing the ability to efficiently combine prior physics-based knowledge with the ability of ML-based to learn from data. These approaches include, for example, incorporating NNs as postprocessors or wrappers to physics-based models (*Jiang et al., 2020; Frame et al., 2021*), or as strategies for learning the empirical forms of process equations embedded in such models (*Rasp et al., 2018; Bennett & Nijssen, 2021*). It has been recently proposed that this emergent field be referred to as *"differentiable hydrology"* (*Shen et al., 2023*) due to its ability to exploit the

strengths of automated differentiable programming (*Baydin et al., 2018*). Strategies span a broad range, from restricting (regularizing) the learnable functional space of NN-based models based on physical principles, to incorporating learnable NN-based functions into physics-based models thereby affording them greater structural flexibility and thereby diminishing their degree of "*over-compression*". Overall, this can be seen a very positive development as it acts to dissolve the artificial barriers between two modeling communities (*Tsai et al., 2021; Feng et al., 2022*).

[10] In the context of this paper (see Section 2), an interesting and notable development is the work of *Hoedt et al. (2021)* who developed and tested a "*mass-conserving*" LSTM, wherein the internal cell-state values across each node in the network are redistributed by implicitly assuming that the state variables represent physically meaningful quantities (e.g., soil moisture). In this regard, the cell state of a LSTM can be treated similarly to an additive stock variable (such as mass and energy), only if the change in the cell state is based on the sum/balance of additive fluxes. Overall, the challenges to interpretation of the internal system dynamics remain, arguably due to the fact that LSTM network was originally developed in the context of natural language processing, and so research still needs to be done to express its internal architecture in a way that aligns with the characteristic dynamical properties of hydrologic and other Earth Science systems. Because the LSTM cell-states represent abstract features in some latent space, extra analyses must be performed to determine how they are informationally related to actual physical quantities (*Lees et al., 2022*). Interesting and creative developments have included the application of metric-based approaches (*Jiang et al., 2022*) and the design of reduced complexity "*HydroLSTM*" architectures (*De la Fuente et al., 2023a*) that seek to facilitate the process of deciphering what has been learned by LSTM-type models.

## 1.3 The Complementary Nature of Physics-Based and ML-Based Modeling

[11] An important observation is that both PC- and RNN-based models can be abstractly represented as directed graphs consisting of recurrent nodes (representing sequentially evolving informational features or states) and links (representing informational relationships, transitions or fluxes), and this applies also to the aforementioned representational approaches used for RR modeling (*Gupta & Nearing, 2014; Saket et al., 2014*). *Figure 1* shows a very simple such graph that we will refer to as a *"HyMod-like"* architecture (*Boyle 2000*), and that is similar to ones that have been widely used to model the catchment-scale input-state-output dynamics of water movement associated with the RR process. Here, there are three recurrent nodes (visually indicated using rectangular boxes) that can be physically interpreted as representing an aggregate soil-moisture state ($x_t^{sm}$), a channel routing state ($x_t^{rt}$) and a groundwater state ($x_t^{gw}$), connected by links that can be interpreted as representing different pathways by which water can move through the system via faster-moving surface-water fluxes ($Z_o(t)$ from the soil moisture tank and $Z_{gw}(t)$ from the surface routing tank) or slower-moving groundwater fluxes ($Z_{gw}(t)$). Because fluxes of mass (and energy) through a physical system are expected to obey conversation principles, appropriate regularizing constraints must be imposed when implementing this directed graph as a computational algorithm, so that our understanding of the fundamental scientific principles underlying the evolution of hydrological systems is properly represented. This idea underlies much of the current exploration of hybrid Machine Learning and Physical Conceptual (ML-PC) approaches to model development in the geosciences (*Jiang et al., 2020; Bennett & Nijssen, 2021; Hoedt et al., 2021; Höge et al., 2022*).

[12] Conversely, the directed graph depicted in *Figure 1* can also be viewed from the ML perspective as representing a deep (multi-layer) NN with a single recurrent node (informational cell-state) in the first hidden layer and two recurrent nodes (informational cell-states) in the second hidden layer, connected by links that represent different pathways by which <u>*information*</u> can move through the system. From this perspective, rather than (necessarily) representing dynamical flows of "*physical*" quantities such as mass and/or energy, the nodes and links of the graph can instead be viewed as representing dynamical flows of "*informational*" quantities through the system. In this context, depending on our goals, we can choose whether or not to impose additional regularizing constraints (such as requiring conservation principles to hold) on the system. Imposing such constraints will necessarily reduce the capacity/flexibility of the system and thereby its ability to learn

from data (*Frame et al., 2023*), but can potentially enhance its interpretability, and possibly (when done correctly) even improve the ability of the system to extrapolate beyond the range of the training data. One other remark that is worth making is that the "*information-flow*" directed graphs employed in ML architectures typically utilize fully-connected (many-to-many) representations of the linkages between hidden units (nodes) and with the input variables, which can complicate the process of interpreting the trained model (*Rai, 2020*), whereas the mass/energy flow directed graphs employed in PC modeling tend to be more sparsely connected (based on theory/prior knowledge).

## 1.4 Goals and Objectives of This Paper

[13] The aim of this paper is to propose and demonstrate a specific type of system/network "*node*" that can potentially form the basis for "*interpretable*" ML-based learning of RNN-type directed graph representations of mass-conserving physical systems. This node, hereafter referred to as a *Mass Conserving Perceptron* (MCP), incorporates several desirable features including:

  (i) Recurrence, so that the dynamical state-variable evolution of system memory can be represented.
  (ii) Ability to impose conservation constraints at the nodal level.
  (iii) Ability to learn unobserved losses of mass/energy from each node.
  (iv) LSTM-like gating, so that the state-variable time-constants can be dynamically adjusted based on current context; and
  (v) Ability to learn the forms of the flux equations governing the behaviors of the system.

[14] To be clear, this current paper does *not* investigate the use of large data sets or complex directed graph (network) architectures to build/learn geoscientific models using ML-technologies. Instead, it is focused on fundamental development and testing of the MCP concept, with the idea that "*geoscientifically interpretable*" ML-type networks (directed-graph-type models) can then be constructed using MCP nodes as the fundamental units. This will enable such networks to be both programmed and trained end-to-end using standard ML-development differentiable programming environments such as PyTorch in Python (*Paszke et al., 2019*), so that geoscientific investigation can thereby benefit from the progressive rapid development of ML technologies.

[15] As a proof of concept, we conduct preliminary testing by applying the MCP unit to catchment-scale RR modeling of the *Leaf River Basin* in Mississippi (*Sorooshian et al., 1983*). We show that the MCP facilitates development of ML-based models that remain consistent with prior physical understanding regarding the nature of the physical system underlying the data generating process, and that enable meaningful interpretation of internal system behaviors and the physical processes governing the dynamics of both unobserved system losses and observed system fluxes. Importantly, relatively complex input-state-output behaviors can be learned and represented using only a single MCP node, suggesting the possibility of parsimonious interpretable representations of physical (mass/energy conserving) systems that are flexible enough to be directly learn the nature of poorly understood process relationships from data.

[16] Of course, this ability comes with some cost – which in this case means that considerable care must be exercised during model development and training. We explore and discuss some of these issues in this paper, but no doubt there is much more that needs to be understood through future work. One specific issue is that it seems beneficial to employ a strategy of progressive training, whereby more complex system architectures inherit the initial values of the weights (parameters) from their trained parent models. A benefit of this strategy is that it facilitates viewing the model as an "*assemblage of hypotheses*" (*Clark et al., 2008*), and enables an understanding of how performance is affected (improves, deteriorates and/or remains unchanged) as additional components (nodes and/or gates) are added to the MCP-based representation of the system, thereby representing a systematic approach to hypothesis testing (*Gong et al., 2013; Nearing et al., 2020*).

[17] Overall, we seek to contribute to the discussion regarding how to bridge the gap between PC models and ML-based models, and thereby address the conflict between a model's predictive accuracy and the possibility of understanding its behaviors – sometimes referred to as the "*accuracy-interpretability*" dilemma (*Florez-*

*Lopez & Ramon-Jeronimo, 2015*) and which is related to the tradeoff between predictive accuracy and model structural complexity (*Nearing, 2014; Höge et al., 2018*). Clearly, one can either start from the side of theory – begin with a (possibly over-compressed) PC model and use data-driven approaches to guide the process of incorporating further (relevant) structural complexity into the model architecture, or start from the side of the data – begin with a generic ML-type of formulation (such as a gated NN such as an LSTM) and use theory-based ideas to progressively regularize the architecture in a direction that becomes more readily interpretable while maintaining a suitable level of performance.

[18]  The manuscript is organized as follows. Section 2 introduces the structure of the generic *Mass Conserving Perceptron* (MCP) and discusses how it can be augmented/altered to facilitate hypothesis testing. Importantly, we discuss various implementation details that enable the forms of the gating functions to be expressed using standard ML technologies and learned using standard differentiable programming ML frameworks. Section 3 describes the modeling environment setup for all the experiments, including the use of RR data, data splitting, loss function used for training, procedure for evaluating model performance, and various PC- and ML-based models used for performance benchmarking. Sections 4 and 5 present and discuss the experimental results. Finally, Section 6 summarizes our findings, and discusses our vision of how the approach can be further developed to enable interpretable coupled mass-energy-information flow ML-based modeling of geoscientific systems.

## 2. The Mass-Conserving Perceptron

[19]  To be clear, in the context of physics and thermodynamics, we define a "process" as giving rise to temporal changes in the system states, reflected by the existence of changing state variables (*Reggiani et al., 1998,1999*). In the context of hydrology, such processes include wetting and drying, reflected in increases and decreases of corresponding storage states in some soil control volume or groundwater control volume. Of course, the process of drying can be realized in several ways through net mass losses via evaporation, seepage or runoff production. In combination with conservation laws, this framework leads to balance equations for mass, momentum and energy, etc. The formulation of closure relation describing the related fluxes leads to a closed equation system, that can be solved using numerical methods.

### 2.1 Simple Representation of a Mass-Conserving System

[20]  Equation (1) shows the conservation equation that can be used to describe the mass balance dynamics of a simple system (**Figure 2a**) where $u(t)$, $x(t)$, and $o(t)$ denote the input flux (mass/time), internal state (mass), and output flux (mass/time) respectively at time $t$. In catchment-scale RR modeling, for example, these can represent the time-evolution of soil moisture (mass storage) and discharge (mass output flux) in response to precipitation (mass input flux).

$$\frac{dx(t)}{dt} = u(t) - o(t) \qquad (1)$$

[21]  Notice that Eqn. (1) is the mass balance formulation used in hydrology community in the absence of sinks and sources. By applying the discrete explicit Euler method, we can integrate Eqn. (1) to obtain Eqn. (2), where the state $X_{t+\Delta t}$ at time $t + \Delta t$ can be expressed as an update to the internal state $X_t$ at time $t$, obtained by adding the total input flux $U_t = \int_t^{t+\Delta t} u(t)dt$ and subtracting the total output flux $O_t = \int_t^{t+\Delta t} o(t)dt$ that occurred during the time interval $\Delta t = (t + \Delta t) - t$ and (for simplicity) assuming that $\Delta t = 1$ temporal unit (such as a day). The integral yields the difference of $U_{t+dt} - U_t$, and $O_{t+dt} - O_t$ that can be replaced by $U(t')dt$, while $t'$ is element of $[t, t + dt]$ and assuming $t = t'$.

$$X_{t+1} = X_t + U_t - O_t \qquad (2)$$

[22]  To proceed, we assume that the magnitude of the output $O_t$ depends on the value of the state $X_t$ through some process parameterization relationship such as $O_t = f_O(X_t|\Theta_O)$ whose behavior is controllable by adjusting its parameters $\Theta_O$. However, as shown by Eqn. (3), this parameterization can be re-expressed in terms

of a time-varying "gating function" $G_t^O = G^O(X_t|\Theta_O) = \left(\frac{f_O(X_t|\Theta_O)}{X_t}\right)$ where $\lim_{X_t \to 0} \left(\frac{f_O(X_t|\Theta_O)}{X_t}\right) \to G^O(X_t = 0|\Theta_O) = G_{X_t=0}^O$ is a finite value, and where $0 \leq G_t^O \leq 1$. Accordingly, $G_t^O$ (hereafter referred to as the "*Output gate*") acts as a time-varying "*output conductivity*" term that acts to control the rate at which mass is released from the state $X_t$ to form the output flux $O_t$. Notice that the state $X_t$ (e.g. soil moisture) is in units of "*quantity*" while the flux $O_t$ (e.g. streamflow) is in units of "*quantity per unit time*".

$$O_t = f_O(X_t|\Theta_O) = \left(\frac{f_O(X_t|\Theta_O)}{X_t}\right) \cdot X_t = G^O(X_t|\Theta_O) \cdot X_t = G_t^O \cdot X_t \qquad (3)$$

[23] In the particular case that $G_t^O = \kappa_O$, where $\kappa_O$ is a constant, this representation describes the behavior of a mass-conservative "*linear*" reservoir, whereas if $G_t^O$ can vary with time it describes the behavior of a "*non-linear*" reservoir. Importantly, the time-varying behavior of $G_t^O$ can be represented by any linear or non-linear function with any degree of complexity. Further, $G_t^O$ could also depend on other quantities besides the current state $X_t$, as long as those quantities can serve as informational surrogates for the thermodynamic forces (e.g. pressure head) driving the flux; for example, we could have $G_t^O = G^O(X_t, U_t|\Theta_O)$ so that gating depends simultaneously on the current state $X_t$ and the system input $U_t$. More generally, $G_t^O$ could even depend on past inputs $U_{t-j}$, state values $X_{t-j}$, or outputs $O_{t-j}$, for $j = 1,2,...$ etc., or other variables as might be appropriate. The only restriction is that $0 \leq G_t^O \leq 1$ to ensure the conservative nature of the system, and we will also generally expect the functional dependence of $G_t^O$ on other quantities to be monotonic non-decreasing (in keeping with the principles of thermodynamics); see *Zehe et al. (2014)*, *Loritz et al. (2018)*, and re-emphasised in *Gharari et al. (2021)*.

[24] By substituting Eqn. (3) into Eqn. (2), and by including another "*input conductivity*" gating function $G_t^U$ (hereafter referred to as the "*Input gate*") we obtain Eqn. (4), which can be rewritten as Eqn. (5):

$$X_{t+1} = (1 - G_t^O) \cdot X_t + G_t^U \cdot U_t \qquad (4)$$

$$X_{t+1} = G_t^R \cdot X_t + G_t^U \cdot U_t \qquad (5)$$

where $G_t^R = 1 - G_t^O$ represents the fraction of the past state $X_t$ that is "*remembered*" by the system from any given time step to the next (hereafter referred to as the "*Remember gate*"). Note that in RR modeling it is common to assume that $G_t^U = 1$ to represent the simple case where all of the precipitation flux $U_t$ is directly added to the system state $X_t$ (see e.g., *Gupta & Sorooshian, 1985*) so that no infiltration excess process occurs that causes mass flux to bypass the system storage, or to be lost prior to entering the system control volume. However, it is also possible for $G_t^U$ to be time varying and positive but less than 1 ($0 \leq G_t^U \leq 1$) to represent either infiltration excess (due to ponding) or saturation excess (due to the soil being unable to accept more moisture) processes, by including appropriate functional representations. In such cases, we will need to include additional process parameterization equations to represent such behaviors, which equations will be accompanied by their own sets of parameters. Regarding other examples of $G_t^U \neq 1$, we refer the readers to Système Hydrologique Européen (SHE; *Abbott et al., 1986*), Hydrologiska Byråns Vattenbalansavdelning (HBV; *Bergström, 1992*) and FLEX-Topo (*Savenije, 2010*) for more information.

[25] Here, we intentionally refer to $G_t^U$, $G_t^O$ and $G_t^R$ as "*gating*" functions, to draw attention to the similarity of the representation introduced above to that employed in the description of gated RNNs. Further, the "*Remember gate*" can be considered analogous to the so-called "*Forget gate*" in the classical LSTM formulation – but we choose this particular naming convention to be consistent with the role of this gate in determining how much of the past state is retained (i.e., remembered) at each time step.

### 2.2 Extending the Mass-Conserving System to Represent Unobserved Loss

[26] To accurately model a typical environmental system, the formulation provided above is not sufficient, since it does not include any representation of unobserved mass (energy) flux losses $L_t$ from (parts of) the system. For example, in catchment-scale RR modeling, evapotranspirative mass flux loss can occur in response to time-evolving atmospheric demand. Incorporating the additional loss term $L_t$ into Eqn. (2) we obtain $X_{t+1} =$

$X_t + U_t - O_t - L_t$. As with the mass output flux, we can assume that the magnitude of $L_t$ depends on the value of the state $X_t$ (representing available mass), $L_t$ can represent unobserved transpiration loss or a sink term where for instance roots take the water from the internal of the soil, while evaporation is flux across the upper boundary. Further, it can also depend on other variables $D_t$ (can be more than one) that are informative about the "*demand*" factors that act to drive the unobserved loss of mass from the system. For example, in catchment-scale RR modeling, data regarding "*potential evapotranspiration*" and/or "*near surface air temperature*" can be considered to be informative regarding the demands for moisture imposed by the vegetation and atmosphere. Mathematically, this can be expressed through a process parameterization relationship $L_t = f_L(X_t, D_t | \Theta_L)$ whose behavior is controllable by adjusting its parameters $\Theta_L$, and that can be re-expressed in terms of a time-varying "*Loss gate*" conductivity $G_t^L = \left(\frac{f_L(X_t, D_t | \Theta_L)}{X_t}\right)$ where $\lim_{X_t \to 0} \left(\frac{f_L(X_t, D_t | \Theta_L)}{X_t}\right) \to G^L(X_t = 0 | L) = G_{X_t=0}^L$ is finite such that $0 \leq G_t^L \leq 1$ controls the rate at which mass is released from the state $X_t$ to form the loss flux $L_t = G_t^L \cdot X_t$.

[27] Eqns. (4) and (5) can therefore be expanded to include the unobserved loss, giving:

$$X_{t+1} = (1 - G_t^O - G_t^L) \cdot X_t + G_t^U \cdot U_t \qquad (6)$$

$$X_{t+1} = G_t^R \cdot X_t + G_t^U \cdot U_t \qquad (7)$$

where the *Remember* gate is now expressed as $G_t^R = 1 - G_t^O - G_t^L$ to satisfy the mass-conservation principle.

## 2.3 Summary of the Basic Mass-Conserving Perceptron (MCP) Unit

[28] The overall set of mathematical equations that describe this generic single state-variable, mass-conserving system with unobserved loss are summarized in *Figure 2b*. We hereafter refer to this unit, with its specific architecture, as a "*Mass-Conserving Perceptron*" (MCP). As demonstrated later, MCP-based network architectures can be trained to estimate the time evolution $L_t$ of unobserved losses even though actual direct data regarding that loss is unavailable. This is in part because the conservation constraint $G_t^R + G_t^O + G_t^L = 1$ makes it possible for observational data regarding the system output to actually provide relevant information regarding the unobserved loss, and in part due to the ability to exploit relevant "*contextual*" information provided by (observed) data $D_t$ about the "*demand*" factors driving the unobserved losses (e.g., potential evapotranspiration or air temperature, etc.).

[29] Note that functional forms of the *Input, Output,* and *Loss* gates $G_t^U$, $G_t^O$ and $G_t^L$ can, flexibly, be either *specified* using theory (prior knowledge) or *learned* from data as parameterized functions (sub-models) during the model development process. Further, and this point cannot be stressed strongly enough, they can be expressed using *any ML (or other) architecture/technology of choice*. This includes methods such as clustering/classification, simple linear or non-linear regression, shallow or deep Artificial Neural Networks (ANNs) – including RNNs, gated RNNs, LSTMs, and even attention-based Transformer mechanisms, etc. – as long as they can be trained via stochastic gradient descent (SGD; *Bottou, 2010*) or any appropriate parameter estimation (weight training) optimization methodology. And by implementing suitable regularizing constraints on the architectures of choice, prior knowledge can be used to constrain/guide the learning process.

[30] In the experiments reported here we begin by representing the gates using simple *Sigmoid* squashing functions. We then explore more complex ANN-based representations implemented using *ReLU* (Rectified Linear Unit) non-linear activations at the hidden nodes and *Sigmoid* squashing to constrain the result on [0,1]. The latter approach is effectively similar to the piecewise linear approximation approach used by *Gharari et al. (2021)*.

## 2.4 Extending the Basic MCP Unit to Encode and Test Additional Hypotheses

[31] As illustrated by *Figure 2b*, the MCP formulation (and extensions of it) can also be used to facilitate the encoding and testing of a variety of hypotheses regarding the nature of the system, thereby helping to enhance physical realism and improve its ability interpretably represent the input-state-output mapping. For example, one can readily explore: (i) whether an input-bypass mechanism is needed, (ii) whether (and how) the

mass/energy conservation constraint should be relaxed, and/or (iii) whether there are benefits to learning an input bias-correction. While the basic MCP unit does not include such (physically-interpretable) components (i.e., input gate $G_t^U = 1$ and mass-relaxation gate $G_t^{MR} = 0$ ), they are readily incorporated so that corresponding hypotheses can be tested (*Gong et al., 2013; Nearing et al., 2020*), thereby enabling exploration of different ideas about how the catchment may be functioning (see Section 4).

[32] The input gate $G_t^U$ controls the fraction of incoming mass (energy) entering the system state. If at any time $t$, we have $G_t^U = 1$ then there is no other flow path except through the system state, while $G_t^U = 0$ represents the opposite extreme where the incoming mass (energy) must completely bypass the system state by some other flow path; in colloquial ML terminology, this is analogous to a "*skip*" connection. Note that this input-bypass gate will be particularly relevant when examining multi-node and flow path network architectures (to be reported in future work). For simplicity, all single node architectures tested in this paper use $G_t^U = 1$.

[33] Similarly, a mass relaxation gate ($G_t^{MR}$) can be added to explore the possibility of unobserved bi-directional mass exchanges with the environment. In this case, $G_t^{MR}$ is allowed to vary over positive and negative values to represent unobserved mass (energy) leaving and entering the system respectively. In catchment-scale RR modeling this can be used to represent state-dependent sub-surface groundwater fluxes across the system boundary (Section 4.5.1).

[34] Finally, one can choose to include an input bias-correction gate $G_t^{UBC}$; in catchment-scale RR modeling this can be used to correct for precipitation under-catch (by setting $G_t^{UBC} > 1$) or over-catch ($0 < G_t^{UBC} < 1$) by the measurement device (Section 4.5.2).

## 2.5 Issues Related to Implementation

[35] All of the models developed and tested in this study, whether based on the aforementioned MCP node or involving other ML architectures (time-delay ANNs, LSTMs, etc.), were implemented in Python version 3.7 using the PyTorch open-source ML differentiable programming framework (*Paszke et al., 2019*). This allowed us to take advantage of the advanced programming resources that PyTorch makes available, including use of optimization algorithms that use backpropagation (*Rumelhart, 1986*) to facilitate efficient parameter learning.

[36] In the process of developing model architectures using the MCP node, it is necessary to train the parameters $\Theta_O$ and $\Theta_L$ etc., that determine the behaviors of the various gating functions (note that <u>all</u> model parameters are associated with the gates). To achieve this using PyTorch, we formulate the gates in a manner that facilitates efficient learning of the parameters via automated backpropagation. For example, the *Output* gate is represented as $G_t^O = \kappa_O \cdot \sigma(S_t)$ with $\kappa_O$ further expressed as $\kappa_O = \exp(c_O)/\Psi$, and where $S_t = f_S(X_t)$ is a time-varying feature that depends on the state $X_t$ in some fashion determined by a parameterized (learnable) feedforward artificial neural network (ANN), the *Sigmoid* function $\sigma(S_t) = (1 + \exp(-S_t))^{-1}$ is used to squash the result onto the range $[0,1]$, and $\kappa_O$ determines the maximum possible conductivity value. The value for the term $\Psi$ is discussed below.

[37] In the relatively simple case that we use a simple linear function $S_t = a_O + b_O \cdot X_t$ for $f_S(\cdot)$, this results in three parameters/weights ($a_O, b_O, c_O$) to be trained, where all three can vary unconstrained on the real number line. The parameters $a_O$ and $b_O$ control the location and steepness of the gating function (see ***Figure S1a*** and ***S2***) while $c_O$ determines the maximum conductivity rate ($\kappa_O$) which is required to be less than $1$. The reason for learning $c_O$ instead of directly training on $\kappa_O$ is that the exponentiation operation $exp(c_O)$ ensures that the value of $\kappa_O$ remains positive.

[38] The other gating functions are parameterized using a similar strategy. As expressed earlier, the functional forms of the various gates can be made to be as simple or complex as desired using a variety of ML architectures (here we restrict our investigation to ANNs), and can depend on (more than one) other informationally relevant variables such as the lagged input and output data values, or other relevant variables such as those that determine unobserved losses (for example, in catchment-scale RR modeling the behavior of the *Loss* gate can depend on potential evapotranspiration or temperature data).

[39] Next, we discuss the relevance of the term $\Psi$. An important implementational issue is to ensure that the mass (energy) conservation principle holds, which requires that the sum of the (positive valued) *Output*, *Loss*, and *Remember* gates must equal 1.0 (i.e., $G_t^O + G_t^L + G_t^R = 1$). To achieve this, we set $\Psi = \exp(c_O) + \exp(c_L) + \exp(c_R)$ which is equivalent to implementing the *SoftMax* function on the gates:

$$\kappa_O = \exp(c_O)/\Psi = \frac{\exp(c_O)}{\exp(c_O) + \exp(c_L) + \exp(c_R)} \tag{8}$$

and similarly, $\kappa_L = \exp(c_L)/\Psi$ and $\kappa_R = \exp(c_R)/\Psi$. Note that this formulation requires that one additional parameter $c_R$ be trained. Further, this step is only necessary when the total number of gates exceeds two (as in the MCP with *Output*, *Loss* and *Remember* gates); when there is no *Loss* gate then $G_t^R = 1 - G_t^O$, and since the *Output* gate is constrained to remain between 0 and 1, so too will the *Remember* gate.

[40] Finally, note that the inputs to the parameterized (learnable) gating functions (e.g., $X_t$ in $S_t = f_S(X_t)$) are only required to be "*informational*", as opposed to having specific units of mass. In other words, they act to *inform* the gates about how open or closed they should be at any point in time (i.e., how large the conductivity values should be), and in the context of the MCP formulation their units are not important. As such, backpropagation-based training of the MCP is greatly facilitated by *standardizing* such inputs (e.g., by subtracting their long-term means and dividing by their standard deviations) so that the values of the associated gating parameters are more easily initialized (typically independently and randomly on $[-1, +1]$ and close to zero) and do not drift very far from their initial points (*Wasserman, 1989*). This will be further discussed in Sections 4 to 6.

## 3. Experimental Setup & Methods

### 3.1 The Leaf River Catchment

[41] In this paper, we explore the expressive capacity of the MCP node unit, and its potential for use in spatially lumped catchment-scale modeling of the RR process. All of the experiments reported in this paper were conducted using the *Leaf River* data set (compiled by the US National Weather Service), which consists of 40 years (WY 1949-1988) of daily data from the humid, 1944 $km^2$, *Leaf River Basin* (LRB) located near Collins in southern Mississippi, USA. The dataset has been extensively used by the hydrological community for model development and testing and consists of cumulative daily values of observed mean areal precipitation ($PP$; mm/day), potential evapotranspiration ($PET$; mm/day), and watershed outlet streamflow ($QQ$; mm/day). Hereafter, $PP$ represents the mass input driver to the system, $QQ$ represents the mass output target to be predicted, and $PET$ is treated as being informative about the energy input driver to the system that helps to determine how much evapotranspiration actually occurs ($AET$; mm/day).

### 3.2 Data Splitting

[42] During model development, it is important to use only a portion of the available data $\mathcal{D}$ for making decisions about the choices of model structure and parameter values (the model representation), while retaining a separate portion for testing the validity of those choices. In this study, we adopt a version of the robust data allocation method proposed and tested by (*Zheng et al., 2022*), that partitions the data so as to ensure distributional consistency of the observational streamflow records across three subsets of the data to be used for *training* ($\mathcal{D}_{train}$), *selection* ($\mathcal{D}_{select}$), and *testing* ($\mathcal{D}_{test}$). $\mathcal{D}_{train}$ is used for learning values for the parameters/weights. $\mathcal{D}_{select}$ is used for making decisions about the network structure/architecture and for selecting values for the network hyper-parameters (i.e., for selecting among alternative structural hypotheses, including numbers of nodes, pattern of connectivity, and complexity of the functional forms for the gates). Finally, $\mathcal{D}_{test}$ is used for independent evaluation of the generalization ability of the trained model.

[43] For our experiments, we selected the $\mathcal{D}_{train}:\mathcal{D}_{select}:\mathcal{D}_{test}$ partitioning ratio to be 2:1:1 respectively. Overall, given a total of 14,610 time-steps/days in the 40-year LRB dataset, this results in the training subset consisting of 7,306 timesteps, and the selection and testing subsets consisting of 3,652 time-steps each.

### 3.3 Metrics used for Training and Performance Assessment

[44] We trained each model architecture by initializing from 10 different random seeds (see Section 3.4) and selected the one with the highest $KGE_{ss}$ score on the *selection* set. The primary metrics used for model training and performance assessment were the scaled version ($KGE_{SS}$; *Knoben et al., 2019; Khatami et al., 2020;* Eqn. (9)) of the Kling-Gupta Efficiency ($KGE$; *Gupta et al., 2009*; Eqn. (10), and its components shown in Eqns. (11)-(13):

$$KGE_{ss} = 1 - \frac{(1-KGE)}{\sqrt{2}} \tag{9}$$

$$KGE = 1 - \sqrt{((\rho-1)^2 + (\beta-1)^2 + (\alpha-1)^2)} \tag{10}$$

$$\alpha^{KGE} = \frac{\sigma_s}{\sigma_o} \tag{11}$$

$$\beta^{KGE} = \frac{\mu_s}{\mu_o} \tag{12}$$

$$\rho^{KGE} = \frac{Cov_{so}}{\sigma_s \sigma_o} \tag{13}$$

where $\sigma_s$ and $\sigma_o$ are the standard deviations, and $\mu_s$ and $\mu_o$ are the means, of the corresponding data-period simulated and observed streamflow hydrographs respectively and, similarly, $Cov_{so}$ is the covariance between the simulated and observed values. Ideal values for $\alpha^{KGE}$, $\beta^{KGE}$ and $\rho^{KGE}$ are $1$, resulting in maximization of the $KGE$ performance metric (indicating perfect model performance). Note that $KGE_{ss}$ rescales the original $KGE$ metric so that a value of $KGE_{ss} = 0$ corresponds to a performance no better on average than use of the long-term observed system output (here streamflow) as the simulated/predicted value at every time step:

[45] To ensure a robust evaluation of the interannual variability of model performance under time-varying hydroclimatic conditions, we plot the 40-year distribution of year-wise <u>annual</u> $KGE_{ss}$ scores, and display these using Box-and-Whisker Plots. So, instead of the conventional and poorly informative approach of assessing long-term average model performance using the overall (integrated/median) $KGE_{ss}$ score computed over some overall evaluation period, we are able to "*distributionally*" assess how the model performs across hydro-climatically different years (e.g., dry, medium, and wet). In particular, we report various percentiles (including $min$, 5%, 25%, $median$, 75%, and 95% values) for $KGE_{ss}$ and the three associated $KGE$ components. As shown by the results (Sections 4 and 5) this is a powerful approach to performance evaluation, because models with essentially the same average long-term performance can display quite different interannual (hydroclimatic) distributions of performance. In particular, performance can be much poorer during drier periods that are characterized by low amounts of precipitation throughout the year.

### 3.4 Training Procedures and Hyperparameter Selection

[46] To minimize the potential effects of state initialization errors, we implemented a model spin-up period. Given no clear consensus on how to optimally achieve this (*Ajami et al., 2014*) we followed *Seck et al., (2015)* and *Shen et al., (2022*) and repeated the first water year data (WY 1949) three times sequentially at the start of the overall 40-year simulation period, thereby initializing the model states using a "*three-year*" spin-up. The state variable values at the end of this spin-up period were then used to initialize the 40-year simulation runs. This process, involving a total of $43 = (40 + 3)$ years of data, including spin-up, was performed for every potential model architecture and set of parameter values.

[47] For parameter training we use the ADAM gradient-based optimization algorithm (*Kingma & Ba, 2014*). The loss function ($KGE$) and its gradient were computed using the streamflow values/timesteps assigned to the training subset. Every potential model architecture was independently trained $10$ times using independent random initializations of the parameters/weights, using the seeds (2925, 9998, 2025, 2525, 3410, 9899, 5555, 2520, 2828, 3140). From these, the model achieving the "*best*" value of the $KGE$ metric on the *selection* data subset was chosen, and its overall (generalization) performance assessed using the *testing* data subset.

[48] Each of the model architectures was trained for several thousand epochs (passes through the entire training data subset) as indicated in *Table S1*, for each of the $10$ random parameter/weight initializations, using a single full-batch of the data. The parameter set with the best *selection* period performance was then either further fine-tuned for an additional number of epochs or used to initialize the next stage of model training. Consistent with the parsimonious nature of the model architectures examined here, the ADAM learning rate was initialized at $2.5 \times 10^{-2}$ for the first $300$ epochs, and then reduced to $1.25 \times 10^{-2}$.

[49] Given the goal of maintaining physical interpretability of the model architectures developed here, no transformation was applied to the input data. However, because it is impossible to know, at the outset, what an appropriate scaling for the nodal states $X_t$ should be when used as informational variables in the gating functions (their range of variation cannot, in general, be known a priori), we adopted the following strategy. First, we "*pre-trained*" each model architecture *without* applying any scaling to the information flows, by initializing all parameters on $[-1,1]$, and using a single fixed random seed. The means and standard deviations of the 40-year simulated state time series obtained for this pre-trained model architecture were then used for information flow scaling (normalization) during the regular training of the model using $10$ different random seeds as mentioned above. During subsequent fine-tuning of the model architecture, the means and standard deviations used for information flow scaling were updated based on the results of the previous training step. Although the issue of variable scaling is an ongoing research topic (*Shao et al., 2020*), our experience suggests that fine-tuning was only really important when exploring the mass-relaxation hypothesis in Section 4.5.1 (see details in *Table S1* of the supplementary materials).

### 3.5 Performance Benchmarks

[50] As benchmarks for comparison, we trained several data-based models that are *not* constrained to be mass-conserving, and that treat the available data only as "*informational*" (in the sense that the mass or energy status of any variable is not considered to be relevant). These include models based on the linear time-series ARX architecture, non-linear time-series models based on the time-delay feedforward ANN architecture, and recurrent models based on the RNN and gated LSTM architectures. Accordingly, because these models are not strongly constrained to obey physical principles (such as mass conservation), they represent a variety of potential benchmarks that are indicative of the possible performance that is practically achievable given the information content of the available data set (*Gong et al., 2013; Nearing et al., 2020*). Note that the specific model architectures selected as performance benchmarks are restricted to ones that only have access to the same information as is available to the MCP-based models (see below), namely current observed values of precipitation ($U_t$) and potential evapotranspiration ($D_t$), and some variable that summarizes the current state of the system.

[51] As with the MCP-based models, all of the benchmark models were implemented and trained using PyTorch. The ARX-based, RNN-based and LSTM-based models were trained for 2000 epochs and the ANN-based models were trained for 5000 epochs, using $KGE$ as the training criterion, ADAM as the optimization algorithm, and a learning rate of $1.25 \ x \ 10^{-2}$. Each input feature was normalized to $[0,1]$ by dividing by its maximum value. Ten versions of every model architecture were trained (using the $10$ random seeds mentioned previously), and the version with the best *selection* period performance was retained.

#### 3.5.1 Linear ARX Time-Series Model Benchmark

[52] Linear ARX time-series models (*Box & Jenkins, 1970*) are arguably among the simplest ways to generate data-based predictions of dynamical system behaviors and are popular both for practical application and as performance benchmarks (*Fang et al., 2017; Lute and Luce 2017; Ehsani et al., 2022*). Here, the streamflow discharge $O_t^{obs}$ was regressed on the current observed values of precipitation ($U_t$) and potential evapotranspiration ($D_t$) and on one past-lagged value of simulated discharge ($O_{t-1}^{sim}$), which provides indirect information about the current state of the system.

### 3.5.2 Time-Delay Artificial Neural Network (ANN) Model Benchmarks

[53] Feedforward time-delay ANN models can be used to model nonlinear input-output relationships such as those occurring in the transformation of rainfall to runoff (*Hsu et al., 1995*), and can be viewed as more complex, non-linear versions of the ARX data-based regression approach. ANN models use one or more hidden layers arranged in series, with each consisting of several nodes that extract different time-varying "*features*" from the data/features presented at the previous layer, by applying a linear transformation followed by a non-linear activation function. The features at the last hidden layer are then combined linearly to generate the predicted output. Here we use a relatively simple fully-connected feed-forward architecture with only one hidden layer consisting of $N_H$ nodes using *Sigmoid* activation functions. The inputs ($O_{t-1}^{sim}$, $U_t$ and $D_t$) to the ANN are the same as in the linear ARX model (Section 3.5.1). Models with different levels of complexity/capacity are constructed by varying the number of nodes $N_H$ in the hidden layer.

### 3.5.3 Recurrent Neural Network (RNN) Model Benchmarks

[54] RNN models are an important data-based approach to modeling dynamical systems (*Rumelhart et al., 1985*). Their potential advantage over the linear ARX and non-linear ANN model structures is their ability to store and utilize information from long time periods in the past, with their cell-states serving as "*informational*" analogues to the mass/energy "*states*" in dynamical physical-conceptual models. Here, we use RNN's having only a single hidden layer, and vary the number of nodes $N_H$ in the hidden layer to achieve different levels of complexity/capacity. Due to their recurrent state-variable structure which keeps track of the current "*state*" of the dynamical system, input variables to the RNN include only current time-step precipitation $U_t$ and potential evapotranspiration $D_t$.

### 3.5.4 Long Short-Term Memory Network (LSTM) Model Benchmarks

[55] LSTM models are more complex gated forms of RNNs (*Figure 2c*). They have been shown to be ideal for data-based modeling of dynamical systems (*Kratzert et al., 2018*). The power and capacity of LSTM-based models comes from the fact that their cell-states are subjected to three dynamical (context-dependent) "*gating*" operations that facilitate the learning of long-term dependencies (*Hochreiter & Schmidhuber, 1997*). Here, we use LSTMs having a single hidden layer with varying numbers of nodes $N_H$, and search over different sequence lengths of 1, 7, 15, 30, 60, 90, 120, 150, 180, 210, 240, 270, 300, 330, 360 and 390 days to find the optimal sequence length for use in computing the cell-states. As with the RNN models, due to their recurrent state-variable structure, input variables to the LSTM include only current time-step precipitation $U_t$ and potential evapotranspiration $D_t$.

## 4. Experimental Results

### 4.1 Behavioral Expressivity of a Single MCP Node

[56] For our first experiment, we implement four potential models of the catchment-scale input-state-output dynamics of the LRB, in which the model architecture consists of a *single* MCP node. This experiment investigates the behavioral expressivity achievable by a single MCP node enabled by the learnable gating mechanism. The difference between these four MCP-based models is the level of complexity assumed for the *Output* (streamflow-flux) and *Loss* (evapotranspiration-flux) gating functions that control the system dynamics. The four architectures are listed below (where the term "*Constant*" refers to not varying with time):

a) **$MC\{O_\kappa L_\kappa\}$**: Both the *Output* and *Loss* gates are constant, so that neither gate generates context-dependent fluxes ($G_t^O = \kappa_O$, $G_t^L = \kappa_L$).

b) **$MC\{O_\kappa L_\sigma\}$**: The *Output* gate is constant, but the *Loss* gate depends on <u>scaled</u> Potential Loss $\widetilde{D}_t$ via context-dependent *Sigmoid* function gating ($G_t^O = \kappa_O$, $G_t^L = \kappa_L \cdot \sigma(b_L + a_L \cdot \widetilde{D}_t)$).

c) **$MC\{O_\sigma L_\kappa\}$**: The *Loss* gate is constant but *Output* depends on the <u>scaled</u> cell state $\widetilde{X}_t$ via context-dependent *Sigmoid* function gating ($G_t^O = \kappa_O \cdot \sigma(b_O + a_O \cdot \widetilde{X}_t)$, $G_t^L = \kappa_L$).

d) $MC\{O_\sigma L_\sigma\}$: The *Output* and *Loss* gates <u>both</u> use context-dependent *Sigmoid* function gating depending on the scaled cell state $\tilde{X}_t$ and scaled Potential Loss driver $\tilde{D}_t$ respectively ($G_t^O = \kappa_O \cdot \sigma(b_O + a_O \cdot \tilde{X}_t)$, $G_t^L = \kappa_L \cdot \sigma(b_L + a_L \cdot \tilde{D}_t)$).

[57] Remember from Section 2.5, that instead of directly tuning $\kappa$, we actually tune on a corresponding trainable parameter $c$ (which can take on unbounded values) and then compute $\kappa = exp(c)/\Psi$ which is physically required to be strictly non-negative. Accordingly, the "*Constant*" gates each have only a single tunable parameter $c$ (where $\kappa = exp(c)/\Psi$ is analogous to the output conductivity parameter of a simple linear reservoir), whereas the context-dependent "*Sigmoid*" gates each have three tunable parameters ($a$, $b$ and $c$) and enable the value of the gates to change, through their dependence on the time-varying contextual information provided. Further, as discussed previously (Section 3.4), the "*information-flows*" into these gates (here the cell-state $X_t$ and Potential Loss driver $D_t$) were <u>scaled</u> to ensure that the corresponding parameter values ($b_O, a_O, b_L, a_L$) tend to remain within or close to the range $[-1,1]$ and can be properly trained using the ADAM optimizer.

[58] In practice, this method of implementing model training worked very well. In the supplementary materials (*Figure S3*), we show that several metrics computed on the *training*, *selection* and *testing* data subsets track each other very closely (especially for the cases with time-varying *Output* gate), as is required to ensure that generalization performance can be expected to be good (i.e., overfitting does not occur). Once the loss function converged during training, the largest $KGE_{SS}$ difference of 0.04 on these data subsets was obtained for the *Constant-Output* gate architecture $MC\{O_\kappa L_\sigma\}$, while for the context-dependent *Output* and *Loss* gate architecture $MC\{O_\sigma L_\sigma\}$ the $KGE_{SS}$ difference was only ~0.01. Further, even though the optimizer was run for a full 5,000 epochs, most of the training occurred within about 250 epochs (as should be expected for this simple, low-dimensional, gradient descent optimization problem).

[59] The main result here is the demonstration of behavioral expressivity achievable by a single MCP node due to the learnable gating mechanism. *Figure 3a* shows the quantile box-plots of the distributions of $KGE_{SS}$ performance for each year in the total 40-year data set achieved by each of the four architectural choices.

- The simplest constant gating architecture $MC\{O_\kappa L_\kappa\}$ achieves a median $KGE_{SS}$~0.64, and performance is very poor (and even negative) for some of the years.
- Context-dependent *Sigmoid Loss* gating $MC\{O_\kappa L_\sigma\}$ improves overall performance only a small amount (notably on the "*worst*" and "*above median*" years).
- A statistically significant performance gain (median $KGE_{SS}$~0.78) is achieved with context-dependent *Sigmoid Output* gating $MC\{O_\sigma L_\kappa\}$.
- Further gains are achieved by $MC\{O_\sigma L_\sigma\}$ (median $KGE_{SS}$~0.85) when <u>both</u> *Output* and *Loss* gating are context-dependent, and notably the inter-quartile distance (inter-annual variability) is considerably reduced.
- Overall, $MC\{O_\sigma L_\sigma\}$ obtains the highest $KGE_{SS}$ for 30 of the 40 water years in the dataset.

[60] More detailed analysis reveals that the constant gating architecture $MC\{O_\kappa L_\kappa\}$ struggles to achieve good values of mass balance ratio ($\beta^{KGE}$) and variability ratio ($\alpha^{KGE}$), and that context-dependent gating is necessary for good values (close to 1.0) of these statistics to be obtained (supplementary materials, *Figure S3*).

- Overall, we see that context-dependent gating enables the model to dramatically improve its representation of input-state-output dynamics, particularly on the dry-to-average water years.

[61] Finally, *Figure 4* compares model-generated output hydrographs with observed streamflow data for the smallest, median, and largest flow-peak water years (similar results for the smallest/median/largest annual-flow volume years appear in the supplementary materials *Figure S4*), and the performance on different percentile years is summarized in *Table S2*.

- With constant *Output* gating the simulations are very flashy due to too much streamflow being generated during low-flow periods (consistent with the metrics indicating $\beta^{KGE} > 1$).
- The use of context-dependent *Output* and *Loss* gating results in significant behavioral improvement, especially during low-flow periods, which is noteworthy since it is achieved using only a single state variable (cell-state).

[62] Focusing on the best performing $MC\{O_\sigma L_\sigma\}$ architecture, we further note that $\beta^{KGE} \sim 1$ which indicates that that the model is doing an excellent job at partitioning incoming precipitation into streamflow and evapotranspirative fluxes, and that $\alpha^{KGE} \sim 1$ which indicates that it is reproducing the full range of variability of flows. This means that its overall performance is effectively limited by the inability to drive $\rho^{KGE} \to 1$ (we have $\rho^{KGE} \sim 0.88$). Further improvement in performance will, therefore, be largely due to better characterization of hydrograph timing and shape.

[63] To conclude this section, although only a parsimonious *single* cell-state architecture has been used to model the dynamics of the catchment system, the learnable context-dependent gating mechanism enables a considerable behavioral expressivity, to the extent that median annual $KGE_{ss}$ exceeding $0.85$ can be achieved.

### 4.2 Imposing a Physical Constraint on Computed Loss

[64] One way the aforementioned model architectures may be considered physically inconsistent is that they do not constrain the computed Loss ($L_t = G_t^L \cdot X_t$) to be less than or equal to Potential Loss $D_t$. We therefore implemented an additional constraint to ensure that $L_t \leq D_t$ and ran tests to see if inclusion of this physical regularizing constraint in $MC\{O_\sigma L_\sigma\}$ would improve performance and/or physical interpretability of the results (*Read et al., 2019*). The "*physically-constrained*" *Loss* gate is now defined as $G_t^{L^{con}} = G_t^L - ReLU(G_t^L - \frac{D_t}{X_t})$ and the new architecture is denoted $MC\{O_\sigma L_\sigma^{con}\}$ (where "*con*" signifies constrained). From *Figure 3a* we see that median $KGE_{ss}$ drops slightly from $0.85$ to $0.84$ and worst year performance declines from $0.44$ to $0.30$.

- Overall, for the LRB, inclusion of the physical constraint has a small impact on performance of the $MC\{O_\sigma L_\sigma\}$ model.

### 4.3 Physical Interpretability of the Trained Model

[65] Since the $MCP$ node is conceptually isomorphic to component units used in ODE-based modeling of mass conserving systems, the trained model can be physically interpreted (*Jiang et al., 2022*). *Figure 3b* shows the form of the learned *Output* gate function, $G_t^O = f_O(X_t | \theta^O)$, where $\theta^O = \{a_O, b_O, c_O\}$, for the $MC\{O_\sigma L_\sigma^{con}\}$ model.

- We observe a steep threshold-like transition from $G_t^O = 0$ when $X_t < \sim 590 \, mm$ to a constant maximum value of $G_t^O = 0.048 \, mm/mm$ when $X_t > \sim 800 \, mm$.

[66] This physically reasonable result indicates that average soil moisture content $X_t$ of the catchment must exceed $\sim 590 \, mm$ before any runoff is generated, and that lateral hydraulic conductivity is more or less constant at $\kappa_O = 0.048 \, mm/mm$ (4.8%) when the average soil moisture content $X_t$ of the catchment exceeds $\sim 800 \, mm$.

[67] In the current study, the output gate can be interpreted as being analogous to lateral hydraulic conductivity. The saturated value of the output gate (0.048 mm/mm on the daily basis) is within a reasonable range for commonly known soil types. Here, the value of the output gate is assumed to depend on the internal cell state via a non-linear non-decreasing sigmoid type of relationship and we ask the model to learn the properties of this relationship. It might be beneficial in future to further constraint this relationship and to consider the impacts of spatial distribution of soil moisture on lateral hydraulic conductivity (output gate) and to examine possible changes in the associated runoff mechanism.

[68] Further, *Figure 3e* shows how the value of the *Output* gate, and the corresponding cell state, vary with time.

- The on/off nature of *Output* gating is clearly visible, showing how the system generates large pulses of streamflow in response to precipitation inputs, while the cell state $X_t$ is seen to vary largely between $\sim 640 \, mm$ to $\sim 770 \, mm$, consistent with the humid nature of the LRB.

[69] Note that whereas the value of parameter $\kappa_O$ is physically interpretable as a conductivity rate, the values of parameters $a_O$ and $b_O$ are only interpretable inasmuch as they control the "*shape*" of the gating function (location and steepness along the $X_t$ axis). This highlights the fact that it is not ultimately the "*parameters*" of a model, but instead the nature/shape of the gating/conductivity function that determines how fluxes are generated, and that is important for understanding the behavior of the system. While it may indeed be possible that the values of parameters $a_O$ and $b_O$ could (in principle) be related to material and geometric properties of the catchment, we do not examine that issue here and leave it for future investigation via a large-sample catchment study. Further, the smooth nature of the gating function is due entirely to its structure being restricted (regularized) by use of a *Sigmoid* non-linearity. In the next section (Section 4.4), we relax this regularizing assumption to examine whether more general/flexible shapes would result in improved model performance, and in further insights into the process of flux generation within the catchment.

[70] Similarly, *Figure 3c* shows the form of the learned *Loss* gate function, $G_t^L = f_L(D_t|\theta^L)$, where $\theta^L = \{a_L, b_L, c_L\}$.

- We see a smooth upward-curving functional form starting from $G_t^L = 0.001 \, mm/mm$ when $D_t = 0 \, mm$ and increasing to $G_t^L = 0.032 \, mm/mm$ when $D_t = 8 \, mm$.

[71] *Figure 3f* shows how the value of the *Loss* gate varies with time, in a somewhat damped response to the corresponding Potential Loss driver $D_t$.

[72] Finally, *Figures 3d and 3g* show the functional form and time-series evolution of the "*Remember*" gate (computed as $G_t^R = (1 - G_t^O - G_t^L)$ to ensure mass balance of inputs, storage, and outputs of the MCP cell).

- Consistent with the humid nature of the LRB, we see that $G_t^R > \sim 0.985 \, mm/mm$ more than 95% of the time (*Figure 3f* and yellow regions of *Figure 3c*) due to the energy-limited nature of the LRB system (water availability exceeds energy-based demand).
- As can be seen from the blue regions of *Figure 3d*, the system only loses significant amounts of its stored water when <u>both</u> the water storage is large ($X_t > \sim 700 \, mm$) and the energy-based demand is significant ($D_t > \sim 7 \, to \, 8 \, mm$).

## 4.4 Learning More Complex Forms for the Context-Dependent Gating Functions

### 4.4.1 Gating Functions with Increased Functional Flexibility

[73] Next, we relax the restriction that the gating functions can only take on forms permitted by passing the output of a simple linear function $S_t = a + b \cdot I_t$ through the *Sigmoid* non-linearity $\sigma(S_t)$, where $I_t$ represents the contextual "*information flux*" used to determine the value of the gate $G_t$. To do this, we instead define the intermediate variable $S_t$ as the output of a more general piecewise linear function $S_t = a + \sum_{j=1}^{N_H} b_j \cdot SeLU(I_t - b_j^*)$. The gating function now has the form of a single-layer feedforward ANN having $N_H$ hidden nodes that are parameterized as $SeLU$ activation units. This ANN has the ability to learn any general $I_t \rightarrow G_t$ mapping (where $0 \leq G_t \leq 1$) for sufficiently large $N_H$. The flexibility/capacity of this more general gating function can be varied simply by varying the number of hidden nodes $N_H$. The functional flexibility of this formulation is conceptually illustrated in *Figure S1b*.

[74] Results using this more flexible ANN-based function for the *Output* gate are shown in *Figure 5a* and *Table S3*, for increasing values of $N_H$, while retaining the *Sigmoid*-based form for the *Loss* gate. This architecture is

referred to as $MC\{O_{AN}L_\sigma^{con}\}$, where subscript $A$ indicates ANN-based gating, and subscript $N$ indicates the number of hidden nodes (larger values indicate more functional flexibility).

- The results indicate that only a very tiny degree of $KGE_{ss}$ improvement is obtained by increasing the *Output* gate complexity.

- Model $MC\{O_{A5}L_\sigma^{con}\}$, with $N = 5$, has (perhaps) a bit tighter distribution of annual $KGE_{ss}$ and improved performance on the poorer (outlier) years, compared to the previous best result $MC_1\{O_\sigma L_\sigma^{con}\}$.

- There seems to be little overall benefit to increasing the flexibility of the *Output* gate since the properties of $KGE_{ss}$ distribution including median year value (0.84), interquartile range (~0.09 to 0.10) stays mostly the same for all cases.

- In contrast, there seems to be some benefit to increasing the flexibility of the *Loss* gate (***Figure 5b & Table S3***; architecture $MC\{O_\sigma L_{AN}^{con}\}$), while retaining the *Sigmoid*-based form for the *Output* gate. Specifically, the performance on outlier years improves significantly when $N \sim 4\ to\ 5$, with the worst year $KGE_{ss}$ increasing from 0.30 to 0.54 and 0.53 respectively. The $N_H = 4$ case appears to be best since it has the smallest interquartile range (0.09).

[75] A further test of all possible combinations of $N_H = 1 \rightarrow 5$ for the *Output* and *Loss* gates (***Figure 5b***) resulted in no significant improvement, while incurring considerable additional expense in the form of many more parameters to be trained (***Figure S5 & Table S4*** in the supplementary materials).

- Overall, we found only limited (marginal) benefit to implementing more complex/flexible forms for the gating functions in the case of this single-node $MCP$ representation of the dynamics of the LRB.

- Given only current-time-step cell state as the source of contextual information determining the behavior of the *Output* gate, it appears that a simple monotonically non-decreasing *Sigmoid*-shaped function is sufficient to capture the dynamics of streamflow.

- Meanwhile a small degree of improvement is obtained by allowing more flexibility in the form of the *Loss* gate. For instance, the $KGE_{ss}$ median of $MC\{O_{A2}L_{A5}^{con}\}$, $MC\{O_{A4}L_{A5}^{con}\}$, and $MC\{O_{A5}L_{A5}^{con}\}$ increases from 0.84 to 0.85. The more flexible *Loss* gate also results in better worst year and low percentile year performance (***Table S4***).

[76] Nonetheless, these results illustrate how the flexibility engendered by a machine-learning-based approach can be used to explore various hypotheses regarding the functional nature of gating processes that give rise to internal and external fluxes within a dynamical system. This allows us to leverage model complexity and predictive performance more properly (*Weijs and Ruddell, 2020*) while pursuing a meaningful parsimonious approach to interpretable scientific discovery (*Gupta and Nearing, 2014*).

### 4.4.2 Gating Functions with Increased Context-Dependence

[77] Next, we examine the benefit of providing each gating function with <u>multiple</u> sources of information for establishing the context-dependence. To keep things relatively simple, we allow each gating function to depend on only two contextual sources of information $I1$ and $I2$ via the formulation $G_t = \kappa \cdot \sigma(S_t)$ where $S_t = a + b_1 \cdot I1 + b_2 \cdot I2$. This formulation adds only one additional trainable parameter to each gating function. For the *Output* gate, we select $I1 = \tilde{X}_t^1$ and $I2 = \tilde{X}_{t-1}^1$, so that the gating can depend on (scaled) values of the cell state and both the current and previous time steps, thereby allowing the model to take advantage of short-term time-delay effects within the catchment. For the *Loss* gate, we select $I1 = \tilde{D}_t$ and $I2 = \tilde{X}_t^1$, so that the gating can depend on both the (scaled) *Potential Loss* and the (scaled) current value of the cell state. This allows the system loss to depend on both environmental demand (indicated by *Potential Loss*) and the current availability of water (indicated by the cell state). The results of these architectural modification are shown in ***Figure 5c***, where the subscript (+) indicates gating dependence on multiple sources of information. As before,

we compare the results to the baseline $MC\{O_\sigma L_\sigma^{con}\}$ model and summarize the performance of percentile years in *Table S5*. We see that:

- Adding additional context dependence to only the *Output* gate ($MC\{O_{\sigma+} L_\sigma^{con}\}$) generally improves model performance for years that were already <u>above</u> the median level of performance. Although median annual $KGE_{ss}$ remains at the same level of ~0.84, values for the 75[th] and 95[th] percentile years improve significantly from 0.87 to 0.90 and 0.91 to 0.93 respectively. Further, although performance for the entire distribution becomes wider than for $MC\{O_\sigma L_\sigma^{con}\}$, the 5[th] percentile year of $KGE_{ss}$ increases from 0.48 to 0.51, and both models share the same 5[th] percentile year of $KGE_{ss} = 0.78$ indicating that performance on below-median years is quite similar.

- In contrast, adding additional context dependence to only the *Loss* gate ($MC\{O_\sigma L_{\sigma+}^{con}\}$) generally improves model performance for years that were <u>below</u> the median level of performance. Performance for the 5[th] percentile year improves significantly, from 0.48 to 0.70, and performance for the 25[th] percentile year improves from 0.78 to 0.80.

- When we allow additional context dependence in <u>both</u> the *Output* and *Loss* gates ($MC\{O_{\sigma+} L_{\sigma+}^{con}\}$) we see general overall improvement in the entire distribution of both <u>above</u> and <u>below</u> median years (although the median remains more or less the same). The median of $KGE_{ss}$ increases from 0.84 to 0.85 compared to $MC\{O_{\sigma+} L_\sigma^{con}\}$ and $MC\{O_\sigma L_{\sigma+}^{con}\}$.

[78] Similar to the exercise reported in Section 4.4.1, we test all possible combinations of $N_H = 1 \rightarrow 5$ for the *Output* and *Loss* gates with multiple sources of information (*Figure S6 & Table S6*).

- Using ANN-based gating for both the *Output* and loss gates maintains the performance of above average years (as compared to less-complex architectures) while achieving improvement in the performance of below average years (so that all water years exhibit at least 0.50 yearly $KGE_{ss}$ skill).

[79] Overall, these results suggest the importance of incorporating information regarding the availability of water in the catchment (internal system storage) when computing the unobserved losses, and of using previous time-lagged information to determine gating behavior. More generally, these preliminary results suggest that further investigation is warranted into what kinds of context-dependence can lead to improved Gating behavior. The methodology demonstrated here can facilitate an efficient (and parsimonious) way to improve the overall predictive accuracy of a model by using machine-learning to explore various alternative hypotheses regarding dependence of the *Output* and *Loss* fluxes on conditions within the system.

## 4.5 Relaxing the Mass Conservation Requirement

[80] Given that many real-world systems (including catchment systems) are governed by conservation laws, it is reasonable to suppose that including a capability for the model to learn about non-observed interactions with the environment could lead to improved generalization performance. Here, we investigate whether relaxing the requirement for strict mass conservation might lead to performance improvements (*Hoedt et al., 2021; Nearing et al., 2021*).

### 4.5.1 Allowing for Unobserved Exchanges of Mass with the Environment

[81] First, we investigate the possibility that non-observed interactions with the environment (that cause unobserved mass to be added or removed from the system) could lead to improved generalization performance (*Perrin et al., 2003*). This capability was added to the baseline $MC\{O_\sigma L_\sigma^{con}\}$ architecture as follows. Recall that, to ensure conservation of mass, the *Remember* gate is computed from the learned *Output* and *Loss* gates as $G_t^R = (1 - G_t^O - G_t^L)$. We now add an additional learnable "Mass Relaxation" ($MR$) gate ($G_t^{MR}$). This results in a modified *Remember* gate, represented as $G_t^R = (1 - G_t^O - G_t^L - G_t^{MR})$.

[82] The flux generated through the MR gate is formulated as $q_t^{MR} = G_t^{MR} \cdot abs(X_t - c_{MR})$ where $-1 < G_t^{MR} < 1$ and $c_{MR} > 0$. The $abs(.)$ operation on $(X_t - c_{MR})$ measures the "*potential energy*" driving the exchange of mass with the environment, while the "*sign*" of $G_t^{MR}$ determines the direction of the exchange.

Notice that the parameter $c_{MR}$ is in the units of the cell state and represents the equilibrium cell state at which there is no mass exchange with the environment. It should normally be set to be larger than $0$ but we can also allow $c_{MR}$ to be negative by removing the positive constraint over the parameter, in which case there will always be water flowing out of the node. Also, another extreme case would be when a value of $c_{MR} > max(X_t)$ is obtained through training, which would correspond to water always flowing into the node.

[83] To ensure trainability, and so that mass is properly conserved, the new gating scheme is implemented as $G_t^{MR} = f_t^{MR} - ReLU(f_t^{MR} - G_t^R)$ where $f_t^{MR}$ is the mass relaxation (MR) gate and $G_t^{MR}$ is the adjusted MR gate value which ensure that it is impossible to lose more water from the system than the total cell state available in the node. The range of $G_t^{MR}$ varies on $[-1,1]$. Positive $G_t^{MR}$ means that water is flowing out of the cell state, and negative $G_t^{MR}$ means that water is flowing into the cell state.

[84] We investigated two type of MR gating functions where the difference lies in whether or not $f_t^{MR}$ depends on the internal cell state.

   a) *Cell-state-dependent MR gating* (denoted $\boldsymbol{MC\{O_\sigma L_\sigma^{con} M_\sigma^R\}}$): Here, we have $f_t^{MR} = \kappa_{MR} \cdot Tanh\left(a_{MR} \cdot \left(\tilde{X}_t - \tilde{c}_{MR}\right)\right)$ where $\kappa_{MR}$, $a_{MR}$ and $\tilde{c}_{MR}$ are trainable parameters, and $\tilde{X}_t$ is the scaled cell state time series. The value of $\kappa_{MR}$ is constrained to lie on $[0,1]$. In this study, we set $a_{MR}$ to be larger than $0$. Hence, when $X_t > c_{MR}$, we have $0 < f_t^{MR} \leq \kappa_{MR}$ which means that water is flowing out of the node, and when $X_t < c_{MR}$, we have $-\kappa_{MR} \leq f_t^{MR} < 0$ which means that water is flowing into the node. The interpretation is opposite if the $a_{MR}$ is set to be less than $0$. Furthermore, if $a_{MR}$ is set to be a very large quantity, we can approximate, arbitrarily closely, the case of constant gating.

   b) *Cell-state-independent (time-constant) MR Gating* (denoted $\boldsymbol{MC\{O_\sigma L_\sigma^{con} M_I^R\}}$): Here, we have the simpler case, formulated as $f_t^{MR} = \kappa_{MR} \cdot Sign(X_t - c_{MR})$, where $0 < \kappa_{MR} < 1$. In this case, when $Sign(X_t - c_{MR}) > 0$, we have $0 < f_t^{MR} \leq \kappa_{MR}$ which means that water is flowing out of the node, and when $Sign(X_t - c_{MR}) < 0$, we have $-\kappa_{MR} \leq f_t^{MR} < 0$ which means that water is flowing into the node. The parameter $c_{MR}$ is normally set to be positive but can be negative if this constraint is removed. In fact, one could even treat the fluxes as learnable "*constant inputs*" or "*constant outputs*" to the cell state, so no gating operation is required. This however is the trivial case; no relevant attempt is made with this regard.

[85] We investigate these two types of MR gating, with or without imposing the constraint on $c_{MR}$ parameter to be positive, which results in a total of 4 cases. The *cell-state-dependent* and *cell-state-independent cases* with unconstrained $c_{MR}$ are denoted as $\boldsymbol{MC\{O_\sigma L_\sigma^{con} M_\sigma^R\}}$ and $\boldsymbol{MC\{O_\sigma L_\sigma^{con} M_I^R\}}$ respectively, and the corresponding cases with $c_{MR}$ constrained to be positive are denoted as $\boldsymbol{MC\{O_\sigma L_\sigma^{con} M_{\sigma r}^R\}}$ and $\boldsymbol{MC\{O_\sigma L_\sigma^{con} M_{Ir}^R\}}$ respectively. (*Figure 6a* and *Table S7*).

- As summarized in *Table S7*, all four cases result in significant performance improvements for the *lower percentile* (drier) years, with $KGE_{SS}^{worst}$ ($KGE_{SS}^{5\%}$) increasing from $0.30$ ($0.48$) to at least $0.46$ ($0.52$) and to as high as $0.60$ ($0.64$) for $\boldsymbol{MC\{O_\sigma L_\sigma^{con} M_\sigma^R\}}$ with unconstrained $c_{MR}$. Meanwhile, skill for the higher percentile years remains mostly the same.

[86] We further analyze the internal nodal structure and its behavior using $\boldsymbol{MC\{O_\sigma L_\sigma^{con} M_\sigma^R\}}$ with unconstrained $c_{MR}$ which appears to be the best case for the MR gate implementation.

- We find that $c_{MR} \sim 783$ which means that the MR gate shown in *Figure 6b* tends to remove water from the system when the cell state is larger than this threshold value. In fact, the Markovian state update based on mass balance can be described as the cell state at current timestep (*Figure 6d*) plus the difference between incoming precipitation flux and the outgoing *Loss* gate flux (*Figure 6c*), and plus the incoming mass relaxation flux (*Figure 6e*).

[87] *Figure 6f* compares the log-transformed hydrograph of $MC\{O_\sigma L_\sigma^{con} M_\sigma^R\}$ against the default $MC\{O_\sigma L_\sigma^{con}\}$ for the dry WY 1952 where the annual flow peak value is the smallest of the 40 years.

- Incorporating the MR gate improves low flow performance with $KGE_{SS}$ for this year improving from 0.30 to 0.71. Clearly, the MR gate is adding water into the system during this dry year since the cell state never reaches the $c_{MR}$ threshold during the year.

- Overall, the result with MR gating supports the hypothesis of an unobserved mass exchange with the environment.

- Meanwhile, the use of MR gating does not affect the long-term water balance which is likely due to the ability of other model components to compensate for the incoming mass (*Figure S7*).

### 4.5.2 Allowing for Bias Corrections to the Input Flux

[88] An alternative, physically-realistic, way for to account for violations of mass conservation is to assess whether systematic observational biases might exist in the input data. For example, in the context of a catchment system, the measured precipitation might be subject to systematic biases such as wind-induced undercatch (*Adam and Lettenmaier, 2003*), frequency of snowfall (*Yang et al., 1999*), or other unforeseeable random factors affecting the measurement process (see review in *Ehsani and Behrangi, 2022*).

[89] Here, we implemented a piecewise-linear "*bias-correction*" (BC) adjustment function ($G_t^{UBC}$), where the nature of this function was learned, jointly along with the other learnable components of the model. We implemented two types of adjustment functions including piecewise-linear and piecewise-quadratic. The piecewise-linear and piecewise-quadratic BC gates are formulated as $G_t^{UBC-PL} = \sum_{j=1}^{N_H} \omega_j \cdot ReLU\left(\widetilde{U}_t - \sigma(\gamma_j)\right)$ and $G_t^{UBC-PQ} = \sum_{j=1}^{N_H} \omega_j \cdot ReLU\left(\widetilde{U}_t - \sigma(\gamma_j)\right) + \gamma_0$ respectively, where $\gamma_j$ and $\omega_j$ are learnable parameters and $\widetilde{U}_t$ is the precipitation scaled by the maximum value $U^{max}$ within the 40 years. The bias-corrected precipitation is then written as $U_t^{BC-PL} = U_t + G_t^{UBC-PL} \cdot U^{max}$ or $U_t^{BC-PQ} = U_t \cdot G_t^{UBC-PQ}$ depending on the type of adjustment function used. These BC gate adjustment functions were implemented on the baseline $MC\{O_\sigma L_\sigma^{con}\}$ architecture, with different numbers of piecewise sections ($N_H = 1 \rightarrow 5$) to progressively increase the flexibility of the model.

[90] The performance of the resulting piecewise-linear input bias-corrected models $MC\{O_\sigma L_\sigma^{con} B_{LN}\}$ is shown in *Figure 5d*.

- As $N_H = 1 \rightarrow 4$, there is a slight reduction in the interquartile range, $KGE_{SS}^{5\%}$ improves by ~0.01 and the worst year performance improves by ~0.02.

- Overall, we observe only minor overall improvement compared to the baseline model $MC\{O_\sigma L_\sigma^{con}\}$ without bias correction (*Table S8*).

- No significant improvement was obtained using the smoother piecewise quadratic functions (see supplementary materials *Figure S8* and *Table S9*), for which larger numbers of parameters had to be trained.

- Overall, the learned piecewise-linear input-bias correction function indicates that trained model <u>adds</u> water to the system when the recorded precipitation value exceeds ~130 mm indicating the possibility of systematic rain gage undercatch above this threshold (see *Figure S9*).

### 4.6 Summary and Discussion

[91] To summarize:

(i) **Training:** We encountered no challenge to training various lumped implementations of MCP-based catchment-scale input-state-output models for the LRB, with only second-decimal-place differences in

$KGE_{SS}$ performance computed independently on the *training*, *selection* and *testing* subsets of the available data.

(ii) **Behavioral Expressivity:** Due to its high degree of conceptually-realistic behavioral expressivity, allowing the MCP to learn time-varying context-dependent *Sigmoid*-based gating for the output and loss fluxes increased the median 40-year $KGE_{SS}$ performance from ~0.64 (baseline with time-constant gating) to ~0.85, and significantly reduced the 40-year interannual variability of performance. Constraining the computed evapotranspirative loss flux to not exceed potential demand had negligible impacts on overall performance. Further, almost "*perfect*" values of streamflow mass balance ratio ($\beta^{KGE}$~1) and variability ratio ($\alpha^{KGE}$~1) were obtained, and high temporal cross-correlation ($\rho^{KGE}$~0.88) indicates excellent characterization of hydrograph timing and shape.

(iii) **Conceptual Interpretability:** Importantly, the parsimonious $MCP$-based model shows a very high degree of conceptual interpretability, learning functional forms for the *Output* and *Loss* gating functions that are physically reasonable. For example, the model learned a very low value for the output conductivity when soil moisture content $X_t < $ ~59 $cm$ increasing sharply to a constant value of $0.048\ mm/mm$ (4.8%) when $X_t > $ ~80 $cm$, indicating that soil moisture must exceed ~59 $cm$ before runoff is generated in response to precipitation inputs. Further, the magnitude of $X_t$ varies mainly between $64 - 77\ cm$, and the *Remember* gate exceeds $0.985\ mm/mm$ most of the time, consistent with the humid, energy-limited, nature of the LRB, so that the system loses significant amounts of water only when *both* the water storage is large ($X_t > $ ~70 $cm$) and the energy-based demand is high ($> 7 - 8\ mm$).

(iv) **Functional Flexibility:** Enabling the MCP to learn more flexible ANN-based forms for the *Output* gating function provided only marginal overall benefits but doing so for the *Loss* gate significantly improved performance on the worst and low percentile (drier) years. However, providing the *Output* gating function with *multiple sources of contextual information* improved performance for years with <u>above</u> median performance, while doing so for the *Loss* gate *significantly* improved performance for years with <u>below</u> median performance. Notably, the combined effect was that performance on every year exceeded $KGE_{SS} = 0.5$.

(v) **Hypothesis Testing:** The test for un-observed state-driven mass exchanges with the environment (mass-relaxation) resulted in significant performance improvements for below median years, and supported the hypothesis that the LRB may tend to lose water through the subsurface when the soil moisture storage exceeds ~78 cm, and gain water otherwise. Incorporating this learnable process into the model significantly improved the representation of dry-year baseflows while not changing the long-term water balance. Meanwhile, the test for systematic observational biases in the input data suggested that systematic rain gage undercatch may be occurring when the recorded precipitation values exceed ~130 $m$.

[92] Overall, these results illustrate how the flexibility engendered by a machine-learning-based approach based in the MCP can be used to explore various hypotheses regarding the functional nature of a dynamical system, thereby pursuing a data-based approach to interpretable scientific discovery (*Gupta and Nearing, 2014*). They suggest that interpretable input-state-output models of lumped catchment-scale hydrology can be learned directly from data using parsimonious MCP-based model network architectures consisting of only a very few nodes (say $1 - 5$; here we only used $1$) trained using off-the shelf ML technologies. In particular, the flexibility by which gating behaviors can be represented using a variety of different informational/attentional/context-based mechanisms provides considerable room for hypothesis-based discovery of how catchments function.

## 5. Benchmarking Against Purely Data-Driven Models

[93] As a final step, we compare the performance of four of the previously discussed single-node MCP-based models with that of several purely data-based model benchmarks. These four *MCP*-based models include:

   a) $MC\{O_\sigma L_\sigma^{con}\}$ from section 4.2, which uses simple *Sigmoid*-function gating with ET-loss constrained to not exceed PET,

   b) $MC\{O_{\sigma+} L_{\sigma+}^{con}\}$ from section 4.4.2, which augments the aforementioned model to provide additional context dependence in *both* the *Output* and *Loss* gates,

   c) $MC\{O_\sigma L_\sigma^{con} M_\sigma^R\}$ from section 4.5.1, which allows for unobserved exchanges of mass with the environment using cell-state-dependent mass relaxation gating,

   d) $MC\{O_{\sigma+} L_{\sigma+}^{con} M_\sigma^R\}$ (an additional model), which simultaneously combines additional context dependence (section 4.4.2) with cell-state-dependent mass relaxation gating (section 4.5.1).

[94] As benchmarks (see Section 3.5), we use non-mass-conserving model architectures (*ARX, ANN, RNN* and *LSTM*) that treat the available data as being only "*informational*". The results are reported in ***Figure 7*** and ***Table S10***. From left to right the first four box-plots show 40-year distributions of annual $KGE_{SS}$ performance for the *MCP*-based single-node $MC\{O_\sigma L_\sigma^{con}\}$, $MC\{O_{\sigma+} L_{\sigma+}^{con}\}$, $MC\{O_\sigma L_\sigma^{con} M_\sigma^R\}$ and $MC\{O_{\sigma+} L_{\sigma+}^{con} M_\sigma^R\}$ models. These are followed by the simplest data-based benchmark model, namely $ARX$. After that are shown three versions of the *ANN*-based model having $N_H = 1, 2$ and 6 hidden nodes respectively, three versions of the *RNN*-based model having $N_H = 1, 2$ and 7 hidden nodes respectively, and three versions of the *LSTM*-based model having $N_H = 1, 2$ and 5 hidden nodes respectively. In these latter cases $N_H = 1$ corresponds to the simplest possible model in that class (i.e., having only one hidden node), $N_H = 2$ corresponds to how performance improves when the model capacity (number of nodes) is doubled, and the third case corresponds to the "best" model performance achieved when the number of hidden nodes was varied between *1* and *10*.

- First, we see that the four single-node *MCP*-based models compare favorably to all of the non-mass-conserving data-based models tested here. In terms of median annual performance and 40-year distribution of annual performance there is arguably not much difference between $MC\{O_{\sigma+} L_{\sigma+}^{con}\}$, $MC\{O_\sigma L_\sigma^{con} M_\sigma^R\}$ and $MC\{O_{\sigma+} L_{\sigma+}^{con} M_\sigma^R\}$, all of which show better dry year performance than the simpler $MC\{O_\sigma L_\sigma^{con}\}$ model. Without more detailed testing, focused on other aspects of model performance, it is difficult to discriminate between these three architectures.

- Next, we see that the simple linear data-based $ARX$ model performs very poorly, whereas the corresponding non-linear $ANN(N_H)$ models, that use the same predictive input information ($U_t$, $D_t$ and $O_{t-1}^{sim}$), have much better distributions of annual performance, with performance improving progressively as the number of hidden nodes is increased from 1 to 6. These results suggests strongly that use of a non-linear model structure is beneficial. Of course, providing the *ARX* and *ANN* models with additional past-lagged values of the aforementioned inputs could potentially result in much better performance, but that would then be providing them with access to information that is not available to the *MCP*-based models.

[95] We now move on to the *RNN*- and *LSTM*-based models, which have the ability to store and utilize information from the past, with their cell-states serving as "*informational*" analogues to the mass/energy "*states*" in dynamical physical-conceptual models.

- We see, overall, that the *RNN*-based models achieve similar performance to their *ANN* counterparts, while performance of the *LSTM*-based models is much better. Arguably the superior performance of the *LSTM*-based models is due to two things – the functional flexibility afforded by the context-dependent gating functions, and the lack of a restriction for each *LSTM*-node (and more generally the entire *LSTM* network) to obey mass-conservation. Note however, that at least two hidden *LSTM*-nodes ($N_H = 2$) are required to achieve performance that is superior of any of the *MCP*-based models (***Figure S10***).

[96] In conclusion, the four single-node *MCP*-based models compare favorably (in terms of distribution of annual $KGE_{SS}$ performance) with that of the various purely data-based non-mass-conserving model benchmarks tested here. Whereas the best "overall" model performance was achieved by the *LSTM*-based architecture, this was achieved using a larger number ($N_H = 5$) of memory cell-states (hidden nodes), while the *MCP*-based models provided arguably comparable performance with only a single memory cell-state, while also obeying physical-conservation principles, being much more interpretable, and being more easily usable for scientific hypothesis testing.

[97] Of course, a more comprehensive head-to-head comparison would necessarily include use of *MCP*-based network architectures with larger numbers of cell-states, and *LSTM*-based architectures that are constrained to be mass conserving *(Hoedt et al., 2021*; *Frame et al., 2023)*, but we leave that for future work. The main purpose of this simple benchmark study was to demonstrate that, by judiciously combining physical principles with machine-learning technologies when designing model/network architectures, it should be possible to develop physically-interpretable models with performance approaching that of purely data-based models.

## 6. Conclusions

### 6.1 Summary & Discussion

[98] This study has proposed the structure for a physically-interpretable *Mass-Conserving Perceptron* (MCP) as a basis for modeling the behavior of geoscientific systems using *Machine Learning* technologies and has explored its use as the basis for physical-conceptual modeling of catchment-scale rainfall-runoff dynamics. Importantly, *MCP*-based models can be fully implemented using off-the-shelf machine learning technologies and can therefore be easily expanded to take advantage of ongoing technical developments.

[99] The *MCP* node facilitates construction of interpretable model architectures that are functionally isomorphic to those used in *ODE*-based modeling of mass-conserving systems; it can be viewed as a certain type of reduced-complexity representation of generic gated recurrent neural networks. Our main goal in this paper was *not* to test the performance of *MCP*-based models on large data sets consisting of multiple catchments, but instead to test and demonstrate the expressivity and interpretability of the MCP node and demonstrate its potential for use in scientific hypothesis testing.

[100] A major feature of the *MCP* unit is imposition of mass conservation at the "*nodal*" level (i.e., the fundamental unit of a directed graph architecture). In any representation of a dynamical system using neural-network type formulations, one can potentially impose mass conservation (or violations thereof) at the *overall network* level, at the *individual nodal* level, and/or *both*. The *MCP* facilitates construction of models/networks where the investigator can control how and where the conservation principle is applied. A further feature is the ability to learn the appropriate (parsimonious) degree of functional complexity and context dependence for the gating functions.

[101] Applied to the *Leaf River Basin*, we found that models constructed using only a *single MCP*-node can demonstrate very high degrees of functional expressivity, such that excellent input-state-output performance (median year $KGE_{SS}^{median} \geq 0.85$ and worst year $KGE_{SS}^{worst} \geq 0.50$) can be achieved using only a single cell state to represent the moisture-storage dynamics of the basin. These models facilitated a variety of hypotheses tests (gating structure, gating context-dependence, input-bias and unobserved mass exchange with the surrounding environment) regarding the functioning of the *LRB.* Further, MCP-based models were found to achieve predictive performance that is comparable to that achievable using non-mass-conserving, purely-data *ARX, ANN, RNN* and *LSTM* (gated-*RNN*) approaches, when fed with the same input information.

[102] Overall, an *MCP*-based modeling approach can facilitate relatively easy exploration of hypotheses regarding the nature of the data-generating process (the physical system), while seeking to balance model complexity with interpretability while maintaining a desirable level of predictive accuracy. Given that physical-conceptual hydrologic models are traditionally constructed with multiple state variables and flow paths, the MCP computational unit can be utilized as the basis for the development of spatially-lumped catchment-scale

hydrologic models (*Wang and Gupta, 2024*), and potentially extended to distributed hydrologic model representations based on the concept of the *Hydrologic Response Unit* (HRU; *Flügel, 1995*), and the *Representative Elementary Watershed* (REW; *Reggiani et al., 1998, 1999, 2000*). Given also that conventional data-based ML models can often be severely pruned (removing nodes and links by as much as 90% or more) after training, after which the reduced-complexity network architectures then re-trained to obtain comparable or even higher performance (see the "*lottery ticket hypothesis*"; *Frankle and Carbin 2019*), it seems plausible that ML technologies could be used to develop physical-conceptual geoscientific system representations that are <u>not</u> "*lossy-over-compressions*" of the information provided by the data. We propose that the *MCP* can form the future basis for such developments.

[103] This study also revisits the issue of what it means for a model to be "*better*". While we have mainly used a single metric (annual $KGE_{ss}$) for model evaluation, this is clearly insufficient given the inherent multi-objective nature of the problem and the importance of explicitly recognize the roles of input data uncertainty and model error. Further, as is well known, a tension that is inherent to the model development process is the trade-off that can exist between the desire for improved "*accuracy and precision*" and the need for better "*physical-conceptual interpretability*".

### 6.2 Towards ML-based Modeling of Coupled Mass-Energy-Information Systems

[104] If it has not become apparent by now, an analogous *Energy Conserving Perceptron* (ECP) can be constructed to represent "*energy*" flows through a system. By properly combining *MCPs* and *ECPs* within a directed graph architecture, it would be possible to model coupled-mass-energy systems that are governed by both mass and energy conservation principles – examples include *Snowpack Accumulation and Melt Models*, and *Land-Surface Models*). More precisely, *MCP's* and *ECP's* would actually be used in conjunction with non-conservative "*Information Flow Perceptron's*" (IFP's) that are not mass-energy constrained, to learn physical-conceptual representations of the coupled nature of <u>mass-energy-information flows</u> through geoscientific systems. Here, for both clarity and brevity, we have restricted the presentation to mass-conservation, but the extensibility and potential for future development is obvious and will form the basis of our future work.

### 6.3 Further Extensions

[105] In the context of catchment-scale modeling, although we have shown that single-*MCP*-node architectures can achieve surprisingly good level of performance for the climatically-humid *Leaf River Basin*, future large catchment-sample investigations (*Gupta et al., 2014*) should explore a variety of different *MCP*-based network architectures for their suitability in different hydro-climatic regimes, and to investigate the bare minimum model complexity required at different locations. The knowledge gained thereby could inform the development of generic continental-scale network architectures useful for transferring learning between different locations and hydroclimatic-zones (*Jiang et al., 2020*). Ultimately, one might hope to construct models that do not require the kinds of massive networks that were used in previous studies (e.g., *Krazert et al., 2019*; *Nearing et al., 2021*). In ongoing work (to be reported separately), we are exploring the interpretability and functional expressivity of directed graph models for catchment-scale rainfall runoff modeling constructed using networks of *MCP* nodes.

[106] Further, although "*parameter optimization*" is currently the main focus of model development using ML technologies, we reiterate our view (see *Gharari et al., 2021*) that the focus of model development should instead be shifting towards "*function space optimization*". This points towards the potential for technologies such as symbolic regression (*Udrescu & Tegmark, 2020*) to be used for discovering plausible forms for the gating functions (*Klotz et al., 2017; Feigl et al., 2020*), with the further potential of being eventually able to regularize the model search to be consistent with physical-conceptual information regarding hydro-climatic context (*Feigl et al., 2022*).

[107] Finally, we acknowledge the suggestion of reviewer Dr. Yang that the problem of learning a dynamical systems model from data while respecting physical principles can also be viewed as a model-based offline Reinforcement Learning (RL) problem (*Levine et al., 2020*), in which the RL agent learns an optimal policy from

a fixed dataset of previously collected state-action-reward-next state tuples, as opposed to actively interacting with the environment to collect new data. In our hydrological context, offline RL can refer to learning an optimal policy from past climate runoff and discharge data. Future work may wish to explore this perspective, the advantage being that many RL algorithms have been developed to model/control/solve Markov decision processes (MDPs) with incomplete information (i.e., Partially Observable Markov Decision Processes (POMDPs)), where the dynamics of the environment are assumed to satisfy the Markov property, but some of the information of the system is unobserved. Such methods could be useful for more reliably inferring the "*memory*" of a catchment and how such catchments behave for a given memory and forcing at any given time step. In this regard, it might be worth exploring whether the MCP can be implemented in the context of a RL algorithm to learn a more physically meaningful architecture for the system (*Liu and Wang, 2021; Engin and Isler, 2023; Banerjee et al., 2023*).

## Acknowledgments

The authors would like to thank the WRR editorial team including Marc Bierkens (Editor), Juliane Mai (Associate Editor), and reviewers *Dr. Yang Yang* and *Dr. Erwin Zehe* plus one anonymous reviewer for taking the time to provide incredibly constructive comments. The first author (YHW) would like to thank the late *Thomas Meixner*, as well as *Jennifer Mcintosh, Martha Whitaker, Eyad Atallah, Dale Ward*, Ty Ferré, Jim Yeh, and *Chris Castro*, for their support, and acknowledge the TA and outreach assistantship support provided by *the Department of Hydrology and Atmospheric Sciences* and the *University of Arizona Data Science Institute* during the final two years of his Ph.D. study, which made the finalization of this work possible. We also thank *University of Arizona Data Science Institute* for providing HPC computation resources. The second author (HVG) acknowledges partial support by the *Australian Centre of Excellence for Climate System Science* (CE110001028), the inspiration and encouragement provided by members of the *Information Theory in the Geosciences* group (geoinfotheory.org), and support for a 4-month research visit to the *Karlsruhe Institute of Technology,* Germany provided by the *KIT International Excellence Fellowship Award* program.

## Open Research

The code for the mass-conserving perceptron and the 40-year Leaf River rainfall-runoff data can be downloaded at https://zenodo.org/records/10002551

**Figure 1:** The illustration of HyMod-Like architecture

**Figure 2:** Direct-graph representation for: a) a simple physically-based "*bucket*" model; b) the generic Mass Conserving *Perceptron* (MCP); and c) a *Long-Short-Term-Memory* network (LSTM)

**Figure 3:** a) Box and whisker plots of the 40-year distributions of annual $KGE_{ss}$ performance metric values for various single-node MCP architectures ($O$ and $L$ indicate *Output* and *Loss* gates respectively; subscripts $\kappa$ and $\sigma$ indicate *constant* and context-dependent *Sigmoid* gating respectively; superscript $con$ indicates that the *actual Loss* is constrained to not exceed *Potential Loss*). Each "box" delimits the 25th and 75th percentiles, central red line indicates the median, whiskers indicate the 5th and 95th percentiles, red "+" symbols indicate outlier years with poor $KGE_{ss}$ skill relative to other years, and the yellow "dot" indicates the year with worst $KGE_{ss}$ skill. Subplots (b) through (g) show results for the $MC\{O_\sigma L_\sigma^{con}\}$ model architecture only. Subplots (b) through (d) show the learned forms of the *Output*, *Loss* and *Remember* gating functions respectively. Subplot (e) shows time-evolution plots of the *Output* gate (solid black line) and corresponding cell state (dashed grey line). Subplot (f) shows time-evolution plots of the *PET-constrained-Loss* gate (solid black line), *unconstrained-Loss* gate (dashed blue line), and corresponding Potential Loss (solid grey line). Subplot (g) shows time-evolution plots of the *Remember* gate (solid black line) and corresponding cell state (dashed grey line).

**Figure 4:** Observed (red dots) and model-simulated (solid black lines) hydrographs for three selected years. Top row shows a dry year (WY 1952), middle row shows a median year (WY 1953), and bottom row shows a wet year (WY 1974) based on annual flow peak. From left to right the columns show results for the $MC\{O_\kappa L_\kappa\}$, $MC\{O_\kappa L_\sigma\}$, $MC\{O_\sigma L_\kappa\}$ and $MC\{O_\sigma L_\sigma\}$ model architectures respectively.

**Figure 5:** Box and whisker plots of the 40-year distributions of annual $KGE_{ss}$ performance metric values for various single-node MCP architectures. In each subplot, the leftmost boxplot represents the baseline architecture $MC\{O_\sigma L_\sigma^{con}\}$ that we are trying to improve upon. Subplots (a) & (b) show use of progressively more flexible (ANN-based) forms for the *Output* and *Loss* gating functions respectively, achieved by increasing the number $N_H$ of hidden nodes. Subplot (c) shows use of increased context dependence with simple *Sigmoid-based* gating. The $MC\{O_{\sigma+} L_\sigma^{con}\}$ architecture uses both current and past *cell states* as contextual information for the *Output* gate, but only current *Potential Loss* for the *Loss* gate. The $MC\{O_\sigma L_{\sigma+}^{con}\}$ architecture uses current *Potential Loss* and cell state as contextual information for the *Loss* gate. The $MC\{O_{\sigma+} L_{\sigma+}^{con}\}$ architecture uses both current and past *cell states* as contextual information for the *Output* gate, and current *Potential Loss* and *cell state* as contextual information for the *Loss* gate. Subplot (d) shows use of piecewise-linear input bias correction, with number of piece-wise segments indicated by the notation $B_{Ln}$ where $n$ is the number of segments. The yellow "dot" indicates the (dry) year with worst $KGE_{ss}$ skill.

**Figure 6**: Results of adding a learnable a *Mass Relaxation* gate to the $MC\{O_\sigma L_\sigma^{con}\}$ architecture. Subplot (a) shows box and whisker plots of the 40-year distributions of annual $KGE_{ss}$ performance metric values for various single-node MCP architectures with *MR* gating. Subplots (b) to (f) show detailed results for the $MC\{O_\sigma L_\sigma^{con} M_\sigma^R\}$ architecture with unconstrained $c_{MR}$, for WY 1952 (a dry year). Subplot (b) shows the form of the MR gating function. Subplot (c) shows the precipitation input & evaporative loss fluxes. Subplot (d) shows the trajectory of the internal cell state. Subplot (e) shows the (unobserved) mass flux passing through the MR gate (into the cell state, for the case of this dry year). Subplot (f) shows, using a log-streamflow plot, how the (unobserved) mass flux entering the system improves the representation of baseflow throughout the year.

**Figure 7:** Benchmark comparison of four of the best interpretable single-node mass-conserving MCP-based models with several non-mass-conserving purely data-based models, including a linear time-delay *ARX* model, three non-linear time-delay *ANN-based* models with increasing numbers of hidden layer nodes ($N_H = $ *1, 2 & 6*), three *RNN-based* models with $N_H = $ *1, 2 & 7*, and three gated *LSTM-based* models with $N_H = $ *1, 2 & 5*. In each case, the largest value of $N_H$ shown for the case with best test period $KGE_{SS}$ when varying $N_H$ from *1* to *10*. To ensure fair comparison, all models use exactly the same input drivers (current observed values of precipitation ($U_t$) and potential evapotranspiration ($D_t$), while the *ARX* and *ANN-based* models use one lagged simulated discharge ($O_{t-1}^{sim}$) to provide indirect information about the current state of the system (note that *MCP, RNN* and *LSTM* track the system state internally). Box and whisker plots show the 40-year distributions of annual $KGE_{SS}$ performance metric values for each model; the yellow "dot" indicates the (dry) year with worst $KGE_{SS}$ skill.

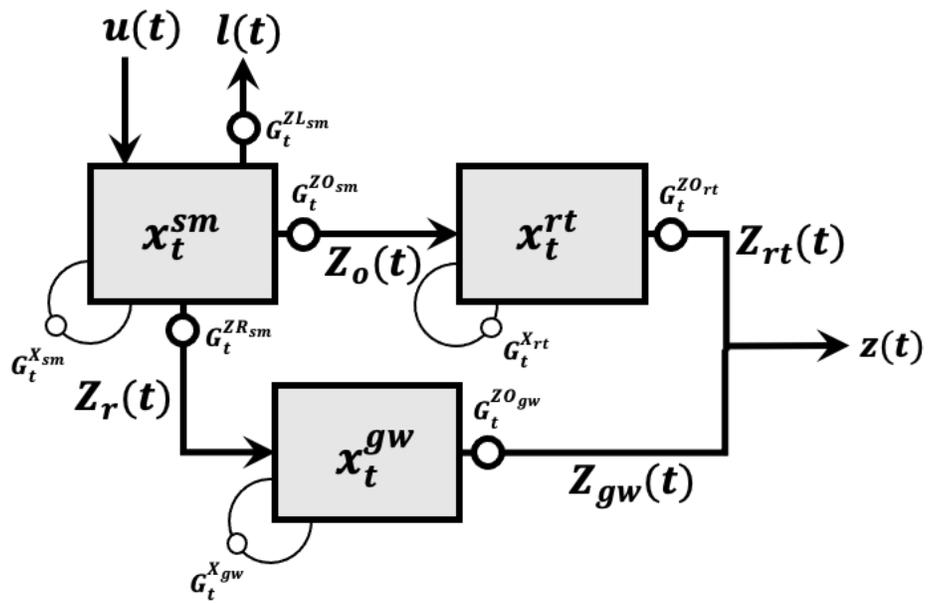

**Figure 1:** The illustration of HyMod-Like architecture

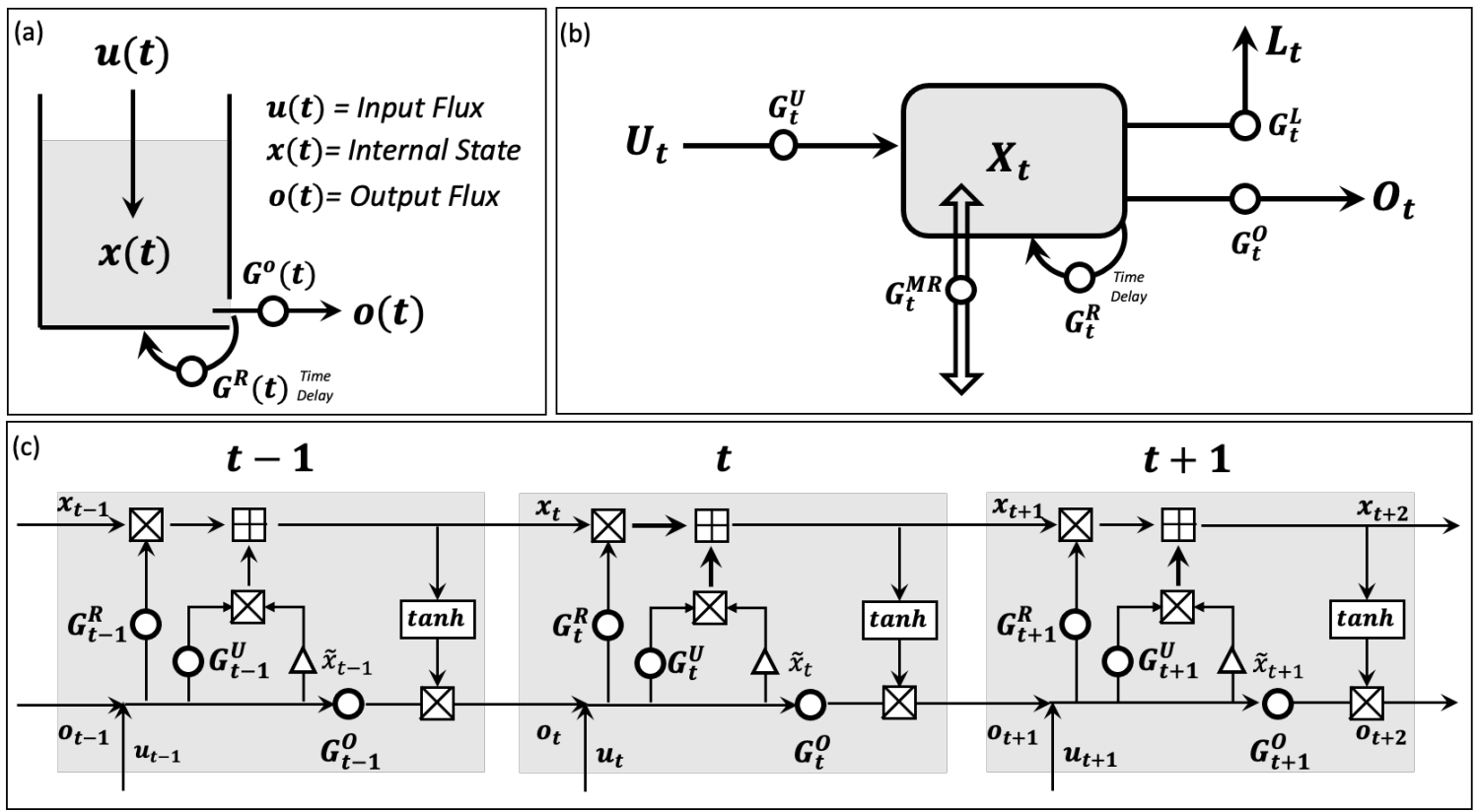

**Figure 2:** Direct-graph representation for: a) a simple physically-based "*bucket*" model; b) the generic Mass Conserving *Perceptron* (MCP); and c) a *Long-Short-Term-Memory* network (LSTM)

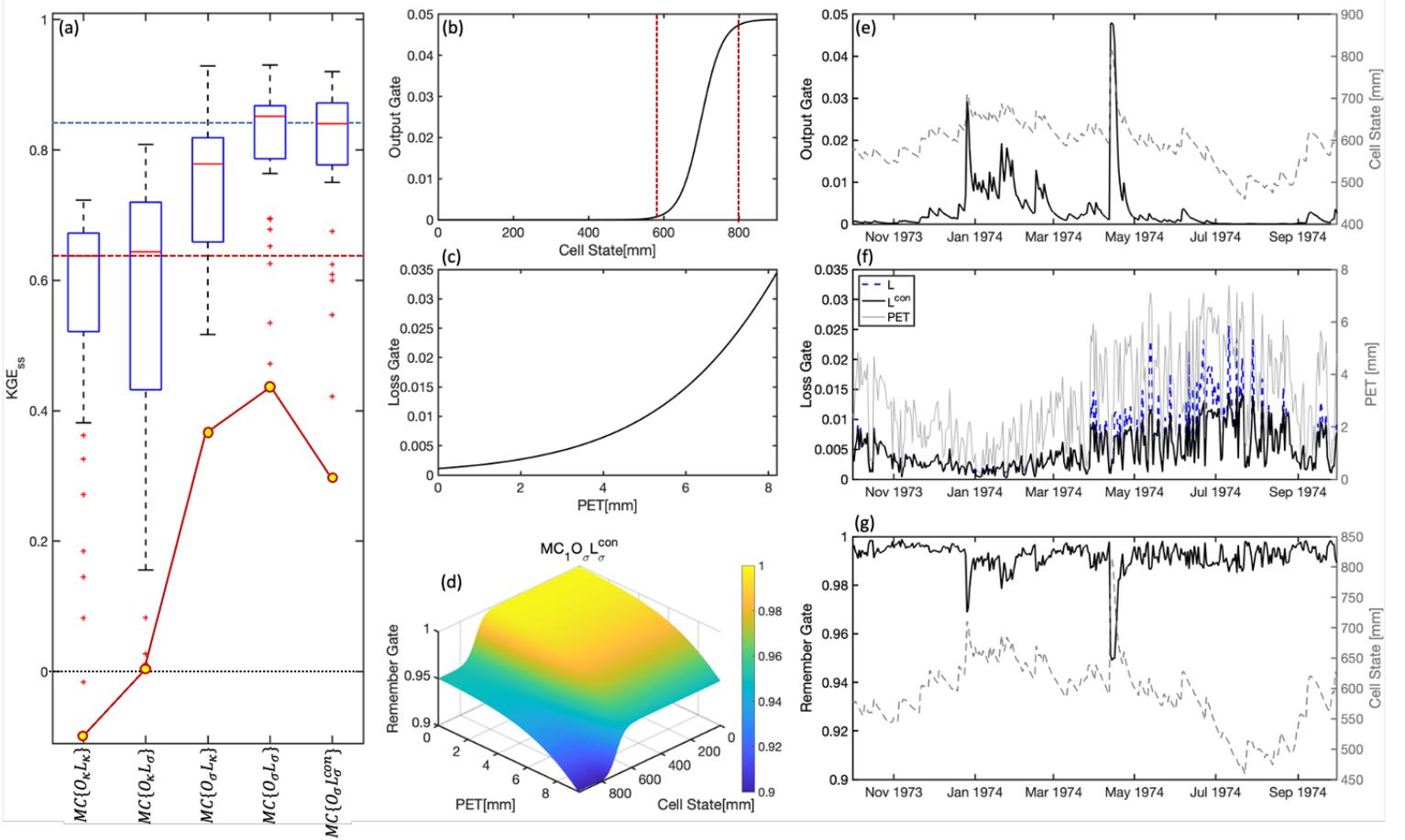

**Figure 3:** a) Box and whisker plots of the 40-year distributions of annual $KGE_{ss}$ performance metric values for various single-node MCP architectures ($O$ and $L$ indicate *Output* and *Loss* gates respectively; subscripts $\kappa$ and $\sigma$ indicate *constant* and context-dependent *Sigmoid* gating respectively; superscript $con$ indicates that the *actual Loss* is constrained to not exceed *Potential Loss*). Each "box" delimits the 25th and 75th percentiles, central red line indicates the median, whiskers indicate the 5th and 95th percentiles, red "+" symbols indicate outlier years with poor $KGE_{ss}$ skill relative to other years, and the yellow "dot" indicates the year with worst $KGE_{ss}$ skill. Subplots (b) through (g) show results for the $MC\{O_\sigma L_\sigma^{con}\}$ model architecture only. Subplots (b) through (d) show the learned forms of the *Output*, *Loss* and *Remember* gating functions respectively. Subplot (e) shows time-evolution plots of the *Output* gate (solid black line) and corresponding cell state (dashed grey line). Subplot (f) shows time-evolution plots of the *PET-constrained-Loss* gate (solid black line), *unconstrained-Loss* gate (dashed blue line), and corresponding Potential Loss (solid grey line). Subplot (g) shows time-evolution plots of the *Remember* gate (solid black line) and corresponding cell state (dashed grey line).

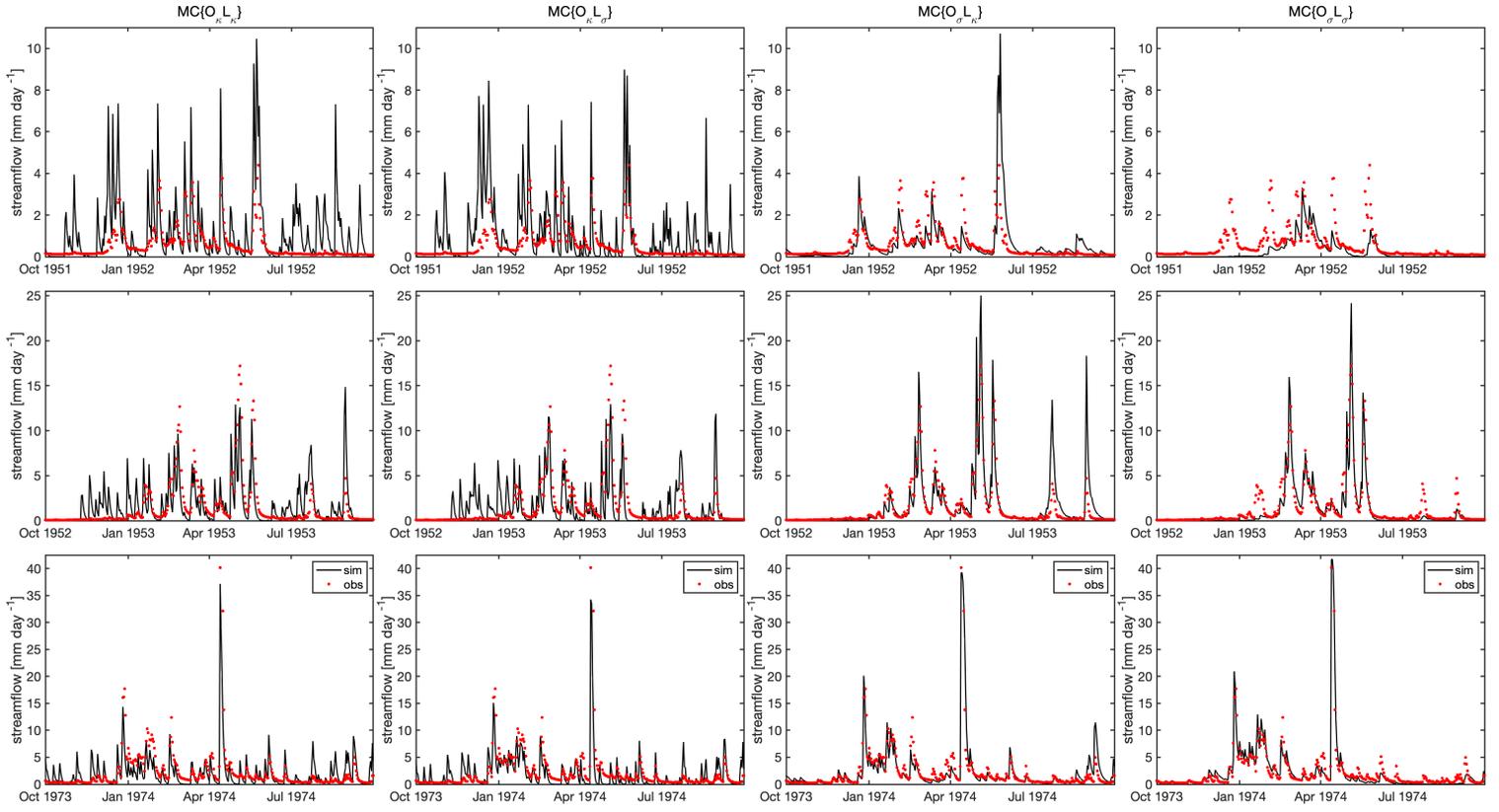

**Figure 4:** Observed (red dots) and model-simulated (solid black lines) hydrographs for three selected years. Top row shows a dry year (WY 1952), middle row shows a median year (WY 1953), and bottom row shows a wet year (WY 1974) based on annual flow peak. From left to right the columns show results for the $MC\{O_\kappa L_\kappa\}$, $MC\{O_\kappa L_\sigma\}$, $MC\{O_\sigma L_\kappa\}$ and $MC\{O_\sigma L_\sigma\}$ model architectures respectively.

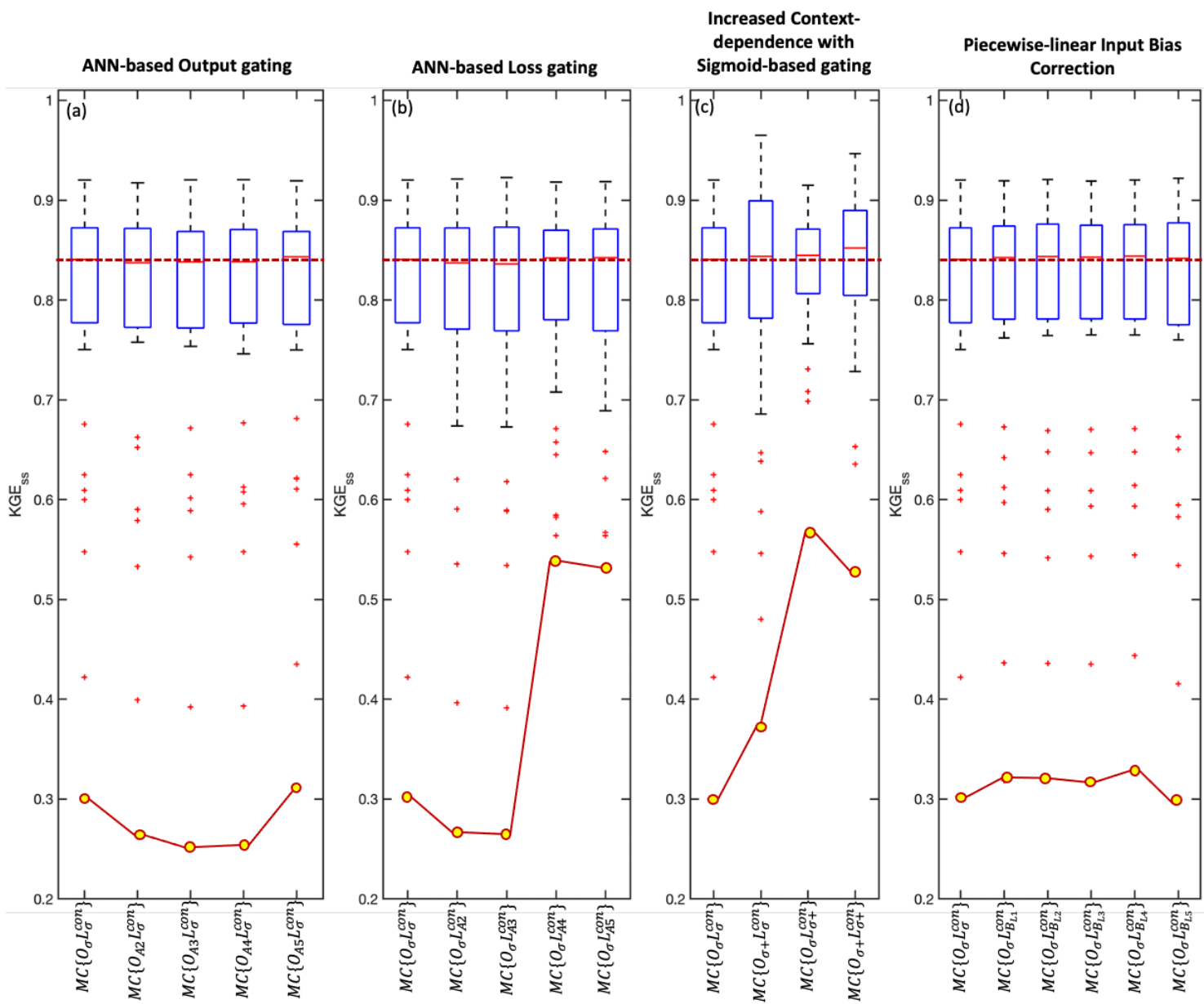

**Figure 5:** Box and whisker plots of the 40-year distributions of annual $KGE_{ss}$ performance metric values for various single-node MCP architectures. In each subplot, the leftmost boxplot represents the baseline architecture $\boldsymbol{MC\{O_\sigma L_\sigma^{con}\}}$ that we are trying to improve upon. Subplots (a) & (b) show use of progressively more flexible (ANN-based) forms for the *Output* and *Loss* gating functions respectively, achieved by increasing the number $N_H$ of hidden nodes. Subplot (c) shows use of increased context dependence with simple *Sigmoid-based* gating. The $\boldsymbol{MC\{O_{\sigma+} L_\sigma^{con}\}}$ architecture uses both current and past *cell states* as contextual information for the *Output* gate, but only current *Potential Loss* for the *Loss* gate. The $\boldsymbol{MC\{O_\sigma L_{\sigma+}^{con}\}}$ architecture uses current *Potential Loss* and cell state as contextual information for the *Loss* gate. The $\boldsymbol{MC\{O_{\sigma+} L_{\sigma+}^{con}\}}$ architecture uses both current and past *cell states* as contextual information for the *Output* gate, and current *Potential Loss* and *cell state* as contextual information for the *Loss* gate. Subplot (d) shows use of piecewise-linear input bias correction, with number of piece-wise segments indicated by the notation $B_{Ln}$ where $n$ is the number of segments. The yellow "dot" indicates the (dry) year with worst $KGE_{ss}$ skill.

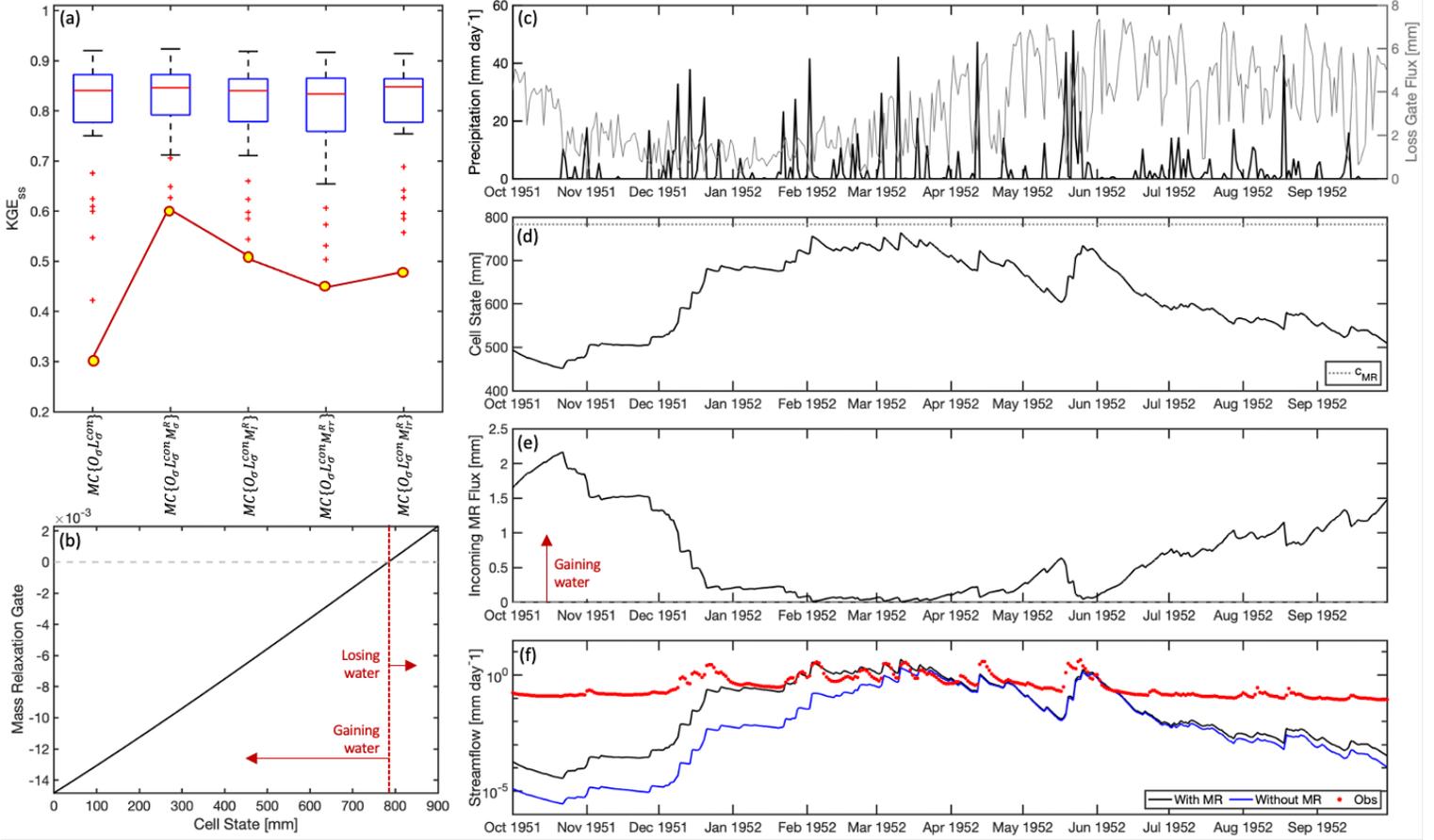

**Figure 6**: Results of adding a learnable a *Mass Relaxation* gate to the $MC_1\{O_\sigma L_\sigma^{con}\}$ architecture. Subplot (a) shows box and whisker plots of the 40-year distributions of annual $KGE_{SS}$ performance metric values for various single-node MCP architectures with *MR* gating. Subplots (b) to (f) show detailed results for the $MC\{O_\sigma L_\sigma^{con} M_\sigma^R\}$ architecture with unconstrained $c_{MR}$, for WY 1952 (a dry year). Subplot (b) shows the form of the MR gating function. Subplot (c) shows the precipitation input & evaporative loss fluxes. Subplot (d) shows the trajectory of the internal cell state. Subplot (e) shows the (unobserved) mass flux passing through the MR gate (into the cell state, for the case of this dry year). Subplot (f) shows, using a log-streamflow plot, how the (unobserved) mass flux entering the system improves the representation of baseflow throughout the year.

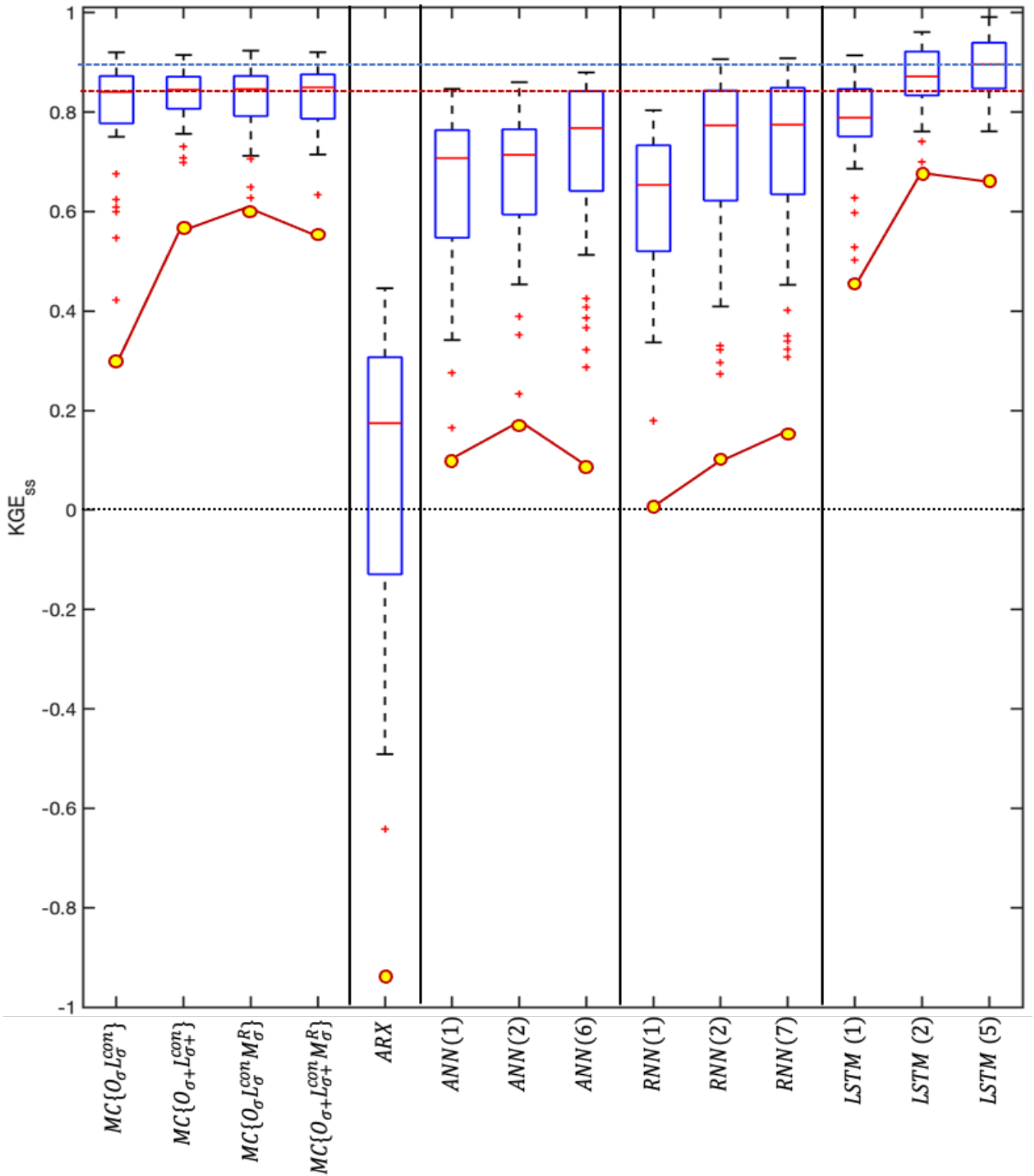

**Figure 7:** Benchmark comparison of four of the best interpretable single-node mass-conserving MCP-based models with several non-mass-conserving purely data-based models, including a linear time-delay *ARX* model, three non-linear time-delay *ANN-based* models with increasing numbers of hidden layer nodes ($N_H = 1, 2$ & $6$), three *RNN-based* models with $N_H = 1, 2$ & $7$, and three gated *LSTM-based* models with $N_H = 1, 2$ & $5$. In each case, the largest value of $N_H$ shown for the case with best test period $KGE_{SS}$ when varying $N_H$ from *1* to *10*. To ensure fair comparison, all models use exactly the same input drivers (current observed values of precipitation ($U_t$) and potential evapotranspiration ($D_t$), while the *ARX* and *ANN-based* models use one lagged simulated discharge ($O_{t-1}^{sim}$) to provide indirect information about the current state of the system (note that *MCP, RNN* and *LSTM* track the system state internally). Box and whisker plots show the 40-year distributions of annual $KGE_{SS}$ performance metric values for each model; the yellow "dot" indicates the (dry) year with worst $KGE_{SS}$ skill.

Supporting Information for

A Mass-Conserving-Perceptron for Machine-Learning-Based Modeling of Geoscientific Systems

Yuan-Heng Wang[1], Hoshin V. Gupta[1]

[1] Department of Hydrology and Atmospheric Science, The University of Arizona, Tucson, AZ

Contact: yhwang0730@gmail.com | hoshin@arizona.edu

Contents of this file

**Introduction**

This Supporting Information provides 10 supplementary table, and 10 supplementary figures to support the discussions in the main text. The contents of these supplementary materials are as follows.

Figure S4. The results of Hydrograph on the selected dry year (WY 1963), median year (WY 1972) and wet year (WY 1980) based on annual total flow volume.

Figure S5. Box and whisker plots of the distributions of annual $KGE_{ss}$ values for $MC\{O_\sigma L_\sigma^{con}\}$ single-node architectures using progressively more flexible (ANN-based) forms of the Output and Loss gating functions by varying $N_H$ from 1 to 5.

Figure S6 Box and whisker plots of the distributions of annual $KGE_{ss}$ values for $MC\{O_\sigma L_\sigma^{con}\}$ architecture using progressively more flexible (ANN-based) forms of the Output and Loss gating functions by varying N from 1 to 5 with multisource of contextual input in both output (cell state at present and previous timestep) and loss gates (cell state and PET at current time step

Figure S7. Cumulative flux for input precipitation, output discharge and the flux generated through loss gate, MR gate, and output gate based on $MC\{O_\sigma L_\sigma^{con} M_\sigma^R\}$ where the discharge is compared against the baseline $MC\{O_\sigma L_\sigma^{con}\}$ case.

Figure S8. Box and whisker plots of the distributions of annual $KGE_{ss}$ values for various single-node ($MC$) architectures when piecewise-quadratic Input Bias Correction is enabled, indicated by the notation $B_{Ln}$ where subscript $n$ refers to the number of piecewise linear sections.

Figure S9. Plots showing the shapes of the learned Input Bias Correction functions for increasing numbers (from 1 to 5) of piecewise linear and quadratic segments.

Figure S10. Box and whisker plots of the distributions of annual $KGE_{ss}$ values for the single node $MC\{O_\sigma L_\sigma^{con}\}$ and the long short-term memory network (LSTM) with different sequence length day.

Table S1: Information of model description, total parameters used, and parameters inheritance for all model cases.

| Model Name | Model Inheritance | Parameter Inheritance | Random Seeds No. | Epoch | Input Standardization for Output/Loss Gate |
|---|---|---|---|---|---|
| $MC\{O_\kappa L_\kappa\}$ | Scratch | NA | 10 | 5000 | No/No |
| $MC\{O_\kappa L_\sigma\}$ | Scratch | NA | 10 | 5000 | No/Yes |
| $MC\{O_\sigma L_\kappa\}^*$ | Scratch | NA | 10 | 5000 | No/No |
| $MC\{O_\sigma L_\kappa\}$ | $MC\{O_\sigma L_\kappa\}^*$ | All Weights and Bias+ | 10 | 5000 | Yes/No |
| $MC\{O_\sigma L_\sigma\}^*$ | Scratch | NA | 10 | 5000 | No/No |
| $MC\{O_\sigma L_\sigma\}$ | $MC\{O_\sigma L_\sigma\}^*$ | All Weights and Bias+ | 10 | 5000 | Yes/Yes |
| $MC\{O_\sigma L_\sigma^{con}\}$ | $MC\{O_\sigma L_\sigma\}$ | All Weights and Bias | NA | 5000 | Yes/Yes |
| $MC\{O_{A1} L_\sigma^{con}\}$ | $MC\{O_\sigma L_\sigma^{con}\}$ | All Weights and Bias+ | 10 | 1000 | Yes/Yes |
| $MC\{O_{AN} L_\sigma^{con}\}$ $(1 < N \leq 5)$ | $MC\{O_{AN-1} L_\sigma^{con}\}$ | All Weights and Bias+ | 10 | 100 | Yes/Yes |
| $MC\{O_\sigma L_{A1}^{con}\}$ | $MC\{O_\sigma L_\sigma^{con}\}$ | All Weights and Bias+ | 10 | 1000 | Yes/Yes |
| $MC\{O_\sigma L_{AM}^{con}\}(1 < M \leq 5)$ | $MC\{O_\sigma L_{AM-1}^{con}\}$ | All Weights and Bias+ | 10 | 100 | Yes/Yes |
| $MC\{O_{AN} L_{AM}^{con}\}$ $(1 \leq N \leq 5; 1 \leq M \leq 5)$ | $MC\{O_\sigma L_\sigma^{con}\}$ | All Weights and Bias except output/loss gate | NA | 100 | Yes/Yes |
| | $MC\{O_{AN} L_\sigma^{con}\}$ | Output Gate Weights and Bias | | | |
| | $MC\{O_\sigma L_{AM}^{con}\}$ | Loss Gate Weights and Bias | | | |

Notice that "All Weights and Bias+" represents all the parameters but without including the parameters in the new added components for the target model.

Table S1: Information of model description, total parameters used, and parameters inheritance for all model cases (continue)

| Model Name | Model Inheritance | Parameter Inheritance | Random Seeds No. | Epoch | Input Standardization for Output/Loss Gate |
|---|---|---|---|---|---|
| $MC\{O_{\sigma+}L_\sigma^{con}\}$ | $MC\{O_\sigma L_\sigma^{con}\}$ | NA | 10 | 1000 | Yes/Yes |
| $MC\{O_\sigma L_{\sigma+}^{con}\}$ | $MC\{O_\sigma L_\sigma^{con}\}$ | NA | 10 | 1000 | Yes/Yes |
| $MC\{O_{\sigma+}L_{\sigma+}^{con}\}$ | $MC\{O_\sigma L_\sigma^{con}\}$ | All Weights and Bias except output/loss gate | NA | 1000 | Yes/Yes |
| | $MC\{O_{\sigma+}L_\sigma^{con}\}$ | Output Gate Weights and Bias | | | Yes/Yes |
| | $MC\{O_\sigma L_{\sigma+}^{con}\}$ | Loss Gate Weights and Bias | | | Yes/Yes |
| $MC\{O_{A1+}L_\sigma^{con}\}$ | $MC\{O_\sigma L_\sigma^{con}\}$ | All Weights and Bias+ | 10 | 1000 | Yes/Yes |
| $MC\{O_{A2+}L_\sigma^{con}\}$ | $MC\{O_{A1+}L_\sigma^{con}\}$ | All Weights and Bias+ | 10 | 500 | Yes/Yes |
| $MC\{O_{AN+}L_\sigma^{con}\}$ ($2 < N \leq 5$) | $MC\{O_{AN-1+}L_\sigma^{con}\}$ | All Weights and Bias+ | 10 | 100 | Yes/Yes |
| $MC\{O_\sigma L_{A1+}^{con}\}$ | $MC\{O_\sigma L_\sigma^{con}\}$ | All Weights and Bias+ | 10 | 1000 | Yes/Yes |
| $MC\{O_\sigma L_{A2+}^{con}\}$ | $MC\{O_\sigma L_{A1+}^{con}\}$ | All Weights and Bias+ | 10 | 500 | Yes/Yes |
| $MC\{O_\sigma L_{AM+}^{con}\}$ ($2 < M \leq 5$) | $MC\{O_\sigma L_{AM-1+}^{con}\}$ | All Weights and Bias+ | 10 | 100 | Yes/Yes |
| $MC\{O_{AN+}L_{AM+}^{con}\}$ ($1 \leq N \leq 5; 1 \leq M \leq 5$) | $MC\{O_\sigma L_\sigma^{con}\}$ | All Weights and Bias except output/loss gate | NA | 100 | Yes/Yes |
| | $MC\{O_{AN+}L_\sigma^{con}\}$ | Output Gate Weights and Bias | | | |
| | $MC\{O_\sigma L_{AM+}^{con}\}$ | Loss Gate Weights and Bias | | | |
| $MC\{O_\sigma L_\sigma^{con}\}M_\sigma^R$ | $MC\{O_\sigma L_\sigma^{con}\}$ | All Weights and Bias+ | 10 | 2000 | Yes/Yes |
| $MC\{O_\sigma L_\sigma^{con}\}M_I^R$ | $MC\{O_\sigma L_\sigma^{con}\}$ | All Weights and Bias+ | 10 | 2000 | Yes/Yes |
| $MC\{O_\sigma L_\sigma^{con}\}M_{\sigma r}^R$ | $MC\{O_\sigma L_\sigma^{con}\}$ | All Weights and Bias+ | 10 | 2000 | Yes/Yes |
| $MC\{O_\sigma L_\sigma^{con}\}M_{Ir}^R$ | $MC\{O_\sigma L_\sigma^{con}\}$ | All Weights and Bias+ | 10 | 2000 | Yes/Yes |
| $MC\{O_\sigma L_\sigma^{con}\}B_{L1}$ | $MC\{O_\sigma L_\sigma^{con}\}$ | All Weights and Bias+ | 10 | 1000 | Yes/Yes |
| $MC\{O_\sigma L_\sigma^{con}\}B_{LN}$ ($1 < N \leq 5$) | $MC\{O_\sigma L_\sigma^{con}\}B_{LN-1}$ | All Weights and Bias+ | 10 | 100 | Yes/Yes |
| $MC\{O_\sigma L_\sigma^{con}\}B_{Q1}$ | $MC\{O_\sigma L_\sigma^{con}\}$ | All Weights and Bias+ | 10 | 1000 | Yes/Yes |
| $MC\{O_\sigma L_\sigma^{con}\}B_{QM}$ ($1 < M \leq 5$) | $MC\{O_\sigma L_\sigma^{con}\}B_{QM-1}$ | All Weights and Bias+ | 10 | 100 | Yes/Yes |
| $MC\{O_{\sigma+}L_{\sigma+}^{con}\}M_\sigma^R$ | $MC\{O_{\sigma+}L_{\sigma+}^{con}\}$ | All Weights and Bias | NA | 2000 | Yes/Yes |
| | $MC\{O_\sigma L_\sigma^{con}\}M_\sigma^R$ | Mass Relaxation Gate Weights and Bias | | | |

Notice that "All Weights and Bias+" represents all the parameters but without including the parameters in the new added components for the target model.

Table S1: Information of model description, total parameters used, and parameters inheritance for all model cases listed (continue)

| Model Name | Model Description | Model Parameters |
|---|---|---|
| $MC\{O_\kappa L_\kappa\}$ | MCP architecture with constant output gate and constant loss gate, no input standardization on output/loss gate | $\kappa_O, \kappa_L, \kappa_R$ |
| $MC\{O_\kappa L_\sigma\}$ | MCP architecture with constant output gate and time-variable (sigmoid) loss gate, with input standardization for loss gate | $\kappa_O, \kappa_L, \kappa_R, a_L, b_L$ |
| $MC\{O_\sigma L_\kappa\}*$ | MCP architecture with time-variable (sigmoid) output gate and constant loss gate, no input standardization on output/loss gate | $\kappa_O, \kappa_L, \kappa_R, a_O, b_O$ |
| $MC\{O_\sigma L_\kappa\}$ | MCP architecture with time-variable (sigmoid) output gate and constant loss gate, with input standardization for output gate | $\kappa_O, \kappa_L, \kappa_R, a_O, b_O$ |
| $MC\{O_\sigma L_\sigma\}*$ | MCP architecture with time-variable (sigmoid) output gate and time-variable (sigmoid) loss gate, no input standardization on output/loss gate | $\kappa_O, \kappa_L, \kappa_R, a_O, b_O, a_L, b_L$ |
| $MC\{O_\sigma L_\sigma\}$ | MCP architecture with time-variable (sigmoid) output gate and time-variable (sigmoid) loss gate, with input standardization on output/loss gate | $\kappa_O, \kappa_L, \kappa_R, a_O, b_O, a_L, b_L$ |
| $MC\{O_\sigma L_\sigma^{con}\}$ | MCP architecture with time-variable (sigmoid) output gate and time-variable (sigmoid) loss gate, with constraint of loss gate flux to be less than PET, with input standardization on output/loss gate | $\kappa_O, \kappa_L, \kappa_R, a_O, b_O, a_L, b_L$ |
| $MC\{O_{AN} L_\sigma^{con}\}$ $(1 \leq N \leq 5)$ | MCP architecture with time-variable (single layer ANN with 1 to 5 nodes) output gate and time-variable (sigmoid) loss gate, with constraint of loss gate flux to be less than PET, with input standardization on output/loss gate | $\kappa_O, \kappa_L, \kappa_R,$ $a_O, b_O^i, b_O^{*i}$ $(1 \leq i \leq 5),$ $a_L, b_L$ |
| $MC\{O_\sigma L_{AM}^{con}\}(1 \leq M \leq 5)$ | MCP architecture with time-variable (sigmoid) output gate and time-variable (single layer ANN with 1 to 5 nodes) loss gate, with constraint of loss gate flux to be less than PET, with input standardization on output/loss gate | $\kappa_O, \kappa_L, \kappa_R,$ $a_O, b_O,$ $a_L, b_L^j, b_L^{*j} (1 \leq j \leq 5)$ |
| $MC\{O_{AN} L_{AM}^{con}\}$ $(1 \leq N \leq 5; 1 \leq M \leq 5)$ | MCP architecture with time-variable (single layer ANN with 1 to 5 nodes) output gate and time-variable (single layer ANN with 1 to 5 nodes) loss gate, with constraint of loss gate flux to be less than PET, with input standardization on output/loss gate | $\kappa_O, \kappa_L, \kappa_R,$ $a_O, b_O^i, b_O^{*i}$ $(1 \leq i \leq 5),$ $a_L, b_L^j, b_L^{*j} (1 \leq j \leq 5)$ |

Table S1: Information of model description, total parameters used, and parameters inheritance for all model cases listed (continue)

| Model Name | Model Description | Model Parameters |
|---|---|---|
| $MC\{O_{\sigma+}L_\sigma^{con}\}$ | MCP architecture with time-variable (sigmoid) output gate and time-variable (sigmoid) loss gate, with constraint of loss gate flux to be less than PET, with cell state at previous timestep as additional input used in output gate, with input standardization on output/loss gate | $\kappa_O, \kappa_L, \kappa_R, a_O, b_O, b_{O+}, a_L, b_L$ |
| $MC\{O_\sigma L_{\sigma+}^{con}\}$ | MCP architecture with time-variable (sigmoid) output gate and time-variable (sigmoid) loss gate, with constraint of loss gate flux to be less than PET, with cell state at current timestep as additional input used in loss gate, with input standardization on output/loss gate | $\kappa_O, \kappa_L, \kappa_R, a_O, b_O, a_L, b_L, b_{L+}$ |
| $MC\{O_{\sigma+}L_{\sigma+}^{con}\}$ | MCP architecture with time-variable (sigmoid) output gate and time-variable (sigmoid) loss gate, with constraint of loss gate flux to be less than PET, with cell state at previous timestep as additional input used in output gate and cell state at current timestep as additional input used in loss gate, with input standardization on output/loss gate | $\kappa_O, \kappa_L, \kappa_R, a_O, b_O, b_{O+}, a_L b_L, b_{L+}$ |
| $MC\{O_{AN+}L_\sigma^{con}\}$ $(1 \leq N \leq 5)$ | MCP architecture with time-variable (single layer ANN with 1 to 5 nodes) output gate and time-variable (sigmoid) loss gate, with constraint of loss gate flux to be less than PET, with cell state at previous timestep as additional input used in output gate and cell state at current timestep as additional input used in loss gate, with input standardization on output/loss gate | $\kappa_O, \kappa_L, \kappa_R, a_O,$ $b_O^i, b_O^{*i} (1 \leq i \leq 5), a_L, b_L,$ $a_O^\times, b_O^{1k}, b_O^{2k} (1 \leq k \leq 5)$ |
| $MC\{O_\sigma L_{AM+}^{con}\}$ $(1 \leq M \leq 5)$ | MCP architecture with time-variable (sigmoid) output gate and time-variable (single layer ANN with 1 to 5 nodes) loss gate, with constraint of loss gate flux to be less than PET, with cell state at previous timestep as additional input used in output gate and cell state at current timestep as additional input used in loss gate, with input standardization on output/loss gate | $\kappa_O, \kappa_L, \kappa_R,$ $a_O, b_O,$ $a_L, b_L^j, b_L^{*j} (1 \leq j \leq 5),$ $a_L^\times, b_L^{1k}, b_L^{2k} (1 \leq k \leq 5)$ |
| $MC\{O_{AN+}L_{AM+}^{con}\}$ $(1 \leq N \leq 5; 1 \leq M \leq 5)$ | MCP architecture with time-variable (single layer ANN with 1 to 5 nodes) output gate and time-variable (single layer ANN with 1 to 5 nodes) loss gate, with constraint of loss gate flux to be less than PET, with cell state at previous timestep as additional input used in output gate and cell state at current timestep as additional input used in loss gate, with input standardization on output/loss gate | $\kappa_O, \kappa_L, \kappa_R,$ $a_O, b_O^{1i}, b_O^{*1i}, b_O^{2i}, b_O^{*2i} (1 \leq i \leq 5),$ $a_L, b_L^{1j}, b_L^{*1j}, b_L^{2j}, b_L^{*2j} (1 \leq j \leq 5),$ $a_O^\times, a_L^\times, b_O^{1k}, b_O^{2k}, b_L^{1k}, b_L^{2k} (1 \leq k \leq 5)$ |

Note: Following the convention in the ML community, an additional bias term ($a_O^\times$ or $a_L^\times$) is employed in the case of multiple sources of inputs in the output and the loss gate (for converting two inputs into the gate dimension using the layer with parameters of $b_{O/L}^1, b_{O/L}^2$), as compared to the formulation presented in the main text section 4.4.

Table S1: Information of model description, total parameters used, and parameters inheritance for all model cases listed (continue)

| Model Name | Model Description | Model Parameters |
|---|---|---|
| $MC\{O_\sigma L_\sigma^{con}\}M_\sigma^R$ | MCP architecture with time-variable (sigmoid) output gate and time-variable (sigmoid) loss gate, with constraint of loss gate flux to be less than PET, with input standardization on output/loss gate, with incorporation of cell-state-dependent mass-relaxation gate | $\kappa_O, \kappa_L, \kappa_R, a_O, b_O, a_L, b_L,$ $\kappa_{MR}, a_{MR}, \tilde{c}_{MR}$ |
| $MC\{O_\sigma L_\sigma^{con}\}M_I^R$ | MCP architecture with time-variable (sigmoid) output gate and time-variable (sigmoid) loss gate, with constraint of loss gate flux to be less than PET, with input standardization on output/loss gate, with incorporation of cell-state-independent mass-relaxation gate | $\kappa_O, \kappa_L, \kappa_R, a_O, b_O, a_L, b_L,$ $\kappa_{MR}, \tilde{c}_{MR}$ |
| $MC\{O_\sigma L_\sigma^{con}\}M_{\sigma r}^R$ | MCP architecture with time-variable (sigmoid) output gate and time-variable (sigmoid) loss gate, with constraint of loss gate flux to be less than PET, with input standardization on output/loss gate, with incorporation of cell-state-dependent mass-relaxation gate and the relaxation of positivity constraint on the threshold value ($\tilde{c}_{MR}$) | $\kappa_O, \kappa_L, \kappa_R, a_O, b_O, a_L, b_L,$ $\kappa_{MR}, a_{MR}, \tilde{c}_{MR}$ |
| $MC\{O_\sigma L_\sigma^{con}\}M_{Ir}^R$ | MCP architecture with time-variable (sigmoid) output gate and time-variable (sigmoid) loss gate, with constraint of loss gate flux to be less than PET, with input standardization on output/loss gate, with incorporation of cell-state-independent mass-relaxation gate and the relaxation of positivity constraint on the threshold value ($\tilde{c}_{MR}$) | $\kappa_O, \kappa_L, \kappa_R, a_O, b_O, a_L, b_L,$ $\kappa_{MR}, \tilde{c}_{MR}$ |
| $MC\{O_\sigma L_\sigma^{con}\}B_{LN}$ ($1 \leq N \leq 5$) | MCP architecture with time-variable (sigmoid) output gate and time-variable (sigmoid) loss gate, with constraint of loss gate flux to be less than PET, with input standardization on output/loss gate, with piece-wise linear input-bias correction gate with varying function degree from 1 to 5 | $\kappa_O, \kappa_L, \kappa_R, a_O, b_O, a_L, b_L,$ $1 \leq \omega_j, \gamma_j \leq 5$ |
| $MC\{O_\sigma L_\sigma^{con}\}B_{QM}$ ($1 \leq M \leq 5$) | MCP architecture with time-variable (sigmoid) output gate and time-variable (sigmoid) loss gate, with constraint of loss gate flux to be less than PET, with input standardization on output/loss gate, with piece-wise quadratic input-bias correction gate with varying function degree from 1 to 5 | $\kappa_O, \kappa_L, \kappa_R, a_O, b_O, a_L, b_L,$ $\gamma_0, 1 \leq \omega_j, \gamma_j \leq 5$ |
| $MC\{O_{\sigma+} L_{\sigma+}^{con}\}M_\sigma^R$ | MCP architecture with time-variable (sigmoid) output gate and time-variable (sigmoid) loss gate, with constraint of loss gate flux to be less than PET, with cell state at current timestep as additional input used in loss gate, with input standardization on output/loss gate, with incorporation of cell-state-dependent mass-relaxation gate and the relaxation of positivity constraint on the threshold value | $\kappa_O, \kappa_L, \kappa_R, a_O, b_O, b_{O+}, a_L, b_L, b_{L+},$ $\kappa_{MR}, a_{MR}, \tilde{c}_{MR}$ |

Table S2: $KGE_{ss}$ Statistics and parameter numbers for single node MCP architecture

|  | $MC\{O_\kappa L_\kappa\}$ | $MC\{O_\kappa L_\sigma\}$ | $MC\{O_\sigma L_\kappa\}$ | $MC\{O_\sigma L_\sigma\}$ | $MC\{O_\sigma L_\sigma^{con}\}$ |
|---|---|---|---|---|---|
| $KGE_{ss}^{worst}$ | -0.10 | 0.01 | 0.37 | 0.44 | 0.30 |
| $KGE_{ss}^{5\%}$ | 0.03 | 0.05 | 0.52 | 0.50 | 0.48 |
| $KGE_{ss}^{25\%}$ | 0.52 | 0.43 | 0.66 | 0.79 | 0.78 |
| $KGE_{ss}^{median}$ | 0.64 | 0.64 | 0.78 | 0.85 | 0.84 |
| $KGE_{ss}^{75\%}$ | 0.67 | 0.72 | 0.82 | 0.87 | 0.87 |
| $KGE_{ss}^{95\%}$ | 0.72 | 0.77 | 0.89 | 0.92 | 0.91 |
| Parameter No. | 3 | 5 | 5 | 7 | 7 |

Table S3: $KGE_{ss}$ Statistics and parameter numbers for single node MCP architecture with varying degree of functional complexity in output and loss gates

|  | $MC\{O_\sigma L_\sigma^{con}\}$ | $MC\{O_{A2} L_\sigma^{con}\}$ | $MC\{O_{A3} L_\sigma^{con}\}$ | $MC\{O_{A4} L_\sigma^{con}\}$ | $MC\{O_{A5} L_\sigma^{con}\}$ |
|---|---|---|---|---|---|
| $KGE_{ss}^{worst}$ | 0.30 | 0.26 | 0.25 | 0.25 | 0.31 |
| $KGE_{ss}^{5\%}$ | 0.48 | 0.47 | 0.47 | 0.47 | 0.50 |
| $KGE_{ss}^{25\%}$ | 0.78 | 0.77 | 0.77 | 0.78 | 0.78 |
| $KGE_{ss}^{median}$ | 0.84 | 0.84 | 0.84 | 0.84 | 0.84 |
| $KGE_{ss}^{75\%}$ | 0.87 | 0.87 | 0.87 | 0.87 | 0.87 |
| $KGE_{ss}^{95\%}$ | 0.91 | 0.91 | 0.91 | 0.91 | 0.91 |
| Parameter No. | 7 | 10 | 12 | 14 | 16 |
|  | $MC\{O_\sigma L_\sigma^{con}\}$ | $MC\{O_\sigma L_{A2}^{con}\}$ | $MC\{O_\sigma L_{A3}^{con}\}$ | $MC\{O_\sigma L_{A4}^{con}\}$ | $MC\{O_\sigma L_{A5}^{con}\}$ |
| $KGE_{ss}^{worst}$ | 0.30 | 0.26 | 0.54 | 0.53 | 0.47 |
| $KGE_{ss}^{5\%}$ | 0.48 | 0.46 | 0.57 | 0.55 | 0.51 |
| $KGE_{ss}^{25\%}$ | 0.78 | 0.77 | 0.78 | 0.77 | 0.74 |
| $KGE_{ss}^{median}$ | 0.84 | 0.84 | 0.84 | 0.84 | 0.84 |
| $KGE_{ss}^{75\%}$ | 0.87 | 0.87 | 0.87 | 0.87 | 0.87 |
| $KGE_{ss}^{95\%}$ | 0.91 | 0.91 | 0.90 | 0.91 | 0.92 |
| Parameter No. | 7 | 10 | 12 | 14 | 16 |

Table S4: $KGE_{ss}$ Statistics and parameter numbers for $MC\{O_\sigma L_\sigma^{con}\}$ single-node architectures using progressively more flexible (ANN-based) forms of the Output and Loss gating functions by varying N from 1 to 5.

|  | $MC\{O_{A1}L_{A1}^{con}\}$ | $MC\{O_{A1}L_{A2}^{con}\}$ | $MC\{O_{A1}L_{A3}^{con}\}$ | $MC\{O_{A1}L_{A4}^{con}\}$ | $MC\{O_{A1}L_{A5}^{con}\}$ |
|---|---|---|---|---|---|
| $KGE_{ss}^{worst}$ | 0.29 | 0.29 | 0.36 | 0.36 | 0.50 |
| $KGE_{ss}^{5\%}$ | 0.48 | 0.48 | 0.50 | 0.50 | 0.53 |
| $KGE_{ss}^{25\%}$ | 0.78 | 0.77 | 0.77 | 0.78 | 0.76 |
| $KGE_{ss}^{50\%}$ | 0.84 | 0.84 | 0.84 | 0.84 | 0.84 |
| $KGE_{ss}^{75\%}$ | 0.87 | 0.87 | 0.87 | 0.87 | 0.87 |
| $KGE_{ss}^{95\%}$ | 0.91 | 0.91 | 0.91 | 0.91 | 0.90 |
| Parameter No. | 9 | 11 | 13 | 15 | 17 |
|  | $MC\{O_{A2}L_{A1}^{con}\}$ | $MC\{O_{A2}L_{A2}^{con}\}$ | $MC\{O_{A2}L_{A3}^{con}\}$ | $MC\{O_{A2}L_{A4}^{con}\}$ | $MC\{O_{A2}L_{A5}^{con}\}$ |
| $KGE_{ss}^{worst}$ | 0.32 | 0.34 | 0.36 | 0.37 | 0.48 |
| $KGE_{ss}^{5\%}$ | 0.49 | 0.50 | 0.50 | 0.51 | 0.53 |
| $KGE_{ss}^{25\%}$ | 0.77 | 0.77 | 0.77 | 0.78 | 0.76 |
| $KGE_{ss}^{50\%}$ | 0.84 | 0.84 | 0.84 | 0.84 | 0.85 |
| $KGE_{ss}^{75\%}$ | 0.87 | 0.87 | 0.87 | 0.87 | 0.87 |
| $KGE_{ss}^{95\%}$ | 0.91 | 0.91 | 0.91 | 0.91 | 0.90 |
| Parameter No. | 11 | 13 | 15 | 17 | 19 |
|  | $MC\{O_{A3}L_{A1}^{con}\}$ | $MC_1\{O_{A3}L_{A2}^{con}\}$ | $MC_1\{O_{A3}L_{A3}^{con}\}$ | $MC_1\{O_{A3}L_{A4}^{con}\}$ | $MC_1\{O_{A3}L_{A5}^{con}\}$ |
| $KGE_{ss}^{worst}$ | 0.26 | 0.31 | 0.37 | 0.39 | 0.50 |
| $KGE_{ss}^{5\%}$ | 0.46 | 0.48 | 0.50 | 0.51 | 0.52 |
| $KGE_{ss}^{25\%}$ | 0.77 | 0.77 | 0.77 | 0.78 | 0.75 |
| $KGE_{ss}^{50\%}$ | 0.84 | 0.84 | 0.84 | 0.84 | 0.84 |
| $KGE_{ss}^{75\%}$ | 0.87 | 0.87 | 0.87 | 0.87 | 0.87 |
| $KGE_{ss}^{95\%}$ | 0.91 | 0.91 | 0.91 | 0.91 | 0.91 |
| Parameter No. | 13 | 15 | 17 | 19 | 21 |
|  | $MC\{O_{A4}L_{A1}^{con}\}$ | $MC\{O_{A4}L_{A2}^{con}\}$ | $MC\{O_{A4}L_{A3}^{con}\}$ | $MC\{O_{A4}L_{A4}^{con}\}$ | $MC\{O_{A4}L_{A5}^{con}\}$ |
| $KGE_{ss}^{worst}$ | 0.26 | 0.31 | 0.38 | 0.39 | 0.49 |
| $KGE_{ss}^{5\%}$ | 0.46 | 0.49 | 0.50 | 0.51 | 0.52 |
| $KGE_{ss}^{25\%}$ | 0.76 | 0.78 | 0.77 | 0.78 | 0.76 |
| $KGE_{ss}^{50\%}$ | 0.84 | 0.84 | 0.84 | 0.84 | 0.85 |
| $KGE_{ss}^{75\%}$ | 0.87 | 0.87 | 0.87 | 0.87 | 0.87 |
| $KGE_{ss}^{95\%}$ | 0.91 | 0.91 | 0.91 | 0.91 | 0.90 |
| Parameter No. | 15 | 17 | 19 | 21 | 23 |
|  | $MC_1\{O_{A5}L_{A1}^{con}\}$ | $MC_1\{O_{A5}L_{A2}^{con}\}$ | $MC_1\{O_{A5}L_{A3}^{con}\}$ | $MC_1\{O_{A5}L_{A4}^{con}\}$ | $MC_1\{O_{A5}L_{A5}^{con}\}$ |
| $KGE_{ss}^{worst}$ | 0.28 | 0.31 | 0.39 | 0.40 | 0.49 |
| $KGE_{ss}^{5\%}$ | 0.47 | 0.49 | 0.50 | 0.51 | 0.52 |
| $KGE_{ss}^{25\%}$ | 0.77 | 0.77 | 0.77 | 0.77 | 0.76 |
| $KGE_{ss}^{50\%}$ | 0.84 | 0.84 | 0.84 | 0.84 | 0.85 |
| $KGE_{ss}^{75\%}$ | 0.87 | 0.87 | 0.87 | 0.87 | 0.87 |
| $KGE_{ss}^{95\%}$ | 0.91 | 0.91 | 0.91 | 0.91 | 0.90 |
| Parameter No. | 17 | 19 | 21 | 23 | 25 |

Table S5: $KGE_{ss}$ Statistics and parameter numbers for single node MCP architecture with multi-source of input information in output and loss gates

|  | $MC\{O_\sigma L_\sigma^{con}\}$ | $MC\{O_{\sigma+} L_\sigma^{con}\}$ | $MC\{O_\sigma L_{\sigma+}^{con}\}$ | $MC\{O_{\sigma+} L_{\sigma+}^{con}\}$ |
|---|---|---|---|---|
| $KGE_{ss}^{worst}$ | 0.30 | 0.37 | 0.57 | 0.53 |
| $KGE_{ss}^{5\%}$ | 0.48 | 0.51 | 0.70 | 0.64 |
| $KGE_{ss}^{25\%}$ | 0.78 | 0.78 | 0.81 | 0.80 |
| $KGE_{ss}^{median}$ | 0.84 | 0.84 | 0.84 | 0.85 |
| $KGE_{ss}^{75\%}$ | 0.87 | 0.90 | 0.87 | 0.89 |
| $KGE_{ss}^{95\%}$ | 0.91 | 0.93 | 0.91 | 0.93 |
| Parameter No. | 7 | 8 | 8 | 9 |
|  | $MC\{O_{A2} L_\sigma^{con}\}$ | $MC\{O_{A3} L_\sigma^{con}\}$ | $MC\{O_{A4} L_\sigma^{con}\}$ | $MC\{O_{A5} L_\sigma^{con}\}$ |
| $KGE_{ss}^{worst}$ | 0.49 | 0.45 | 0.46 | 0.47 |
| $KGE_{ss}^{5\%}$ | 0.52 | 0.48 | 0.49 | 0.49 |
| $KGE_{ss}^{25\%}$ | 0.77 | 0.77 | 0.77 | 0.77 |
| $KGE_{ss}^{median}$ | 0.86 | 0.86 | 0.86 | 0.86 |
| $KGE_{ss}^{75\%}$ | 0.90 | 0.90 | 0.90 | 0.90 |
| $KGE_{ss}^{95\%}$ | 0.93 | 0.94 | 0.94 | 0.94 |
| Parameter No. | 13 | 15 | 17 | 19 |
|  | $MC\{O_\sigma L_{A2}^{con}\}$ | $MC\{O_\sigma L_{A3}^{con}\}$ | $MC\{O_\sigma L_{A4}^{con}\}$ | $MC\{O_\sigma L_{A5}^{con}\}$ |
| $KGE_{ss}^{worst}$ | 0.53 | 0.53 | 0.50 | 0.53 |
| $KGE_{ss}^{5\%}$ | 0.69 | 0.69 | 0.61 | 0.64 |
| $KGE_{ss}^{25\%}$ | 0.79 | 0.79 | 0.75 | 0.78 |
| $KGE_{ss}^{median}$ | 0.84 | 0.84 | 0.84 | 0.84 |
| $KGE_{ss}^{75\%}$ | 0.87 | 0.87 | 0.88 | 0.87 |
| $KGE_{ss}^{95\%}$ | 0.91 | 0.92 | 0.91 | 0.91 |
| Parameter No. | 13 | 15 | 17 | 19 |

Table S6: $KGE_{ss}$ Statistics and parameter numbers for $MC\{O_\sigma L_\sigma^{con}\}$ single-node architectures using progressively more flexible (ANN-based) forms of the Output and Loss gating functions by varying N from 1 to 5 with multisource of contextual input in both output (cell state at present and previous timestep) and loss gates (cell state and PET at current time step).

|  | $MC\{O_{A1+}L_{A1+}^{con}\}$ | $MC\{O_{A1+}L_{A2+}^{con}\}$ | $MC\{O_{A1+}L_{A3+}^{con}\}$ | $MC\{O_{A1+}L_{A4+}^{con}\}$ | $MC\{O_{A1+}L_{A5+}^{con}\}$ |
|---|---|---|---|---|---|
| $KGE_{ss}^{worst}$ | 0.52 | 0.54 | 0.54 | 0.50 | 0.50 |
| $KGE_{ss}^{5\%}$ | 0.62 | 0.62 | 0.62 | 0.62 | 0.62 |
| $KGE_{ss}^{25\%}$ | 0.82 | 0.83 | 0.83 | 0.82 | 0.81 |
| $KGE_{ss}^{50\%}$ | 0.86 | 0.86 | 0.86 | 0.86 | 0.86 |
| $KGE_{ss}^{75\%}$ | 0.89 | 0.89 | 0.89 | 0.89 | 0.89 |
| $KGE_{ss}^{95\%}$ | 0.93 | 0.92 | 0.92 | 0.94 | 0.94 |
| Parameter No. | 15 | 19 | 23 | 27 | 31 |
|  | $MC\{O_{A2+}L_{A1+}^{con}\}$ | $MC\{O_{A2+}L_{A2+}^{con}\}$ | $MC\{O_{A2+}L_{A3+}^{con}\}$ | $MC\{O_{A2+}L_{A4+}^{con}\}$ | $MC\{O_{A2+}L_{A5+}^{con}\}$ |
| $KGE_{ss}^{worst}$ | 0.55 | 0.57 | 0.57 | 0.52 | 0.52 |
| $KGE_{ss}^{5\%}$ | 0.65 | 0.66 | 0.65 | 0.65 | 0.64 |
| $KGE_{ss}^{25\%}$ | 0.83 | 0.83 | 0.83 | 0.82 | 0.82 |
| $KGE_{ss}^{50\%}$ | 0.86 | 0.86 | 0.86 | 0.86 | 0.86 |
| $KGE_{ss}^{75\%}$ | 0.89 | 0.89 | 0.89 | 0.89 | 0.89 |
| $KGE_{ss}^{95\%}$ | 0.92 | 0.92 | 0.92 | 0.93 | 0.93 |
| Parameter No. | 19 | 23 | 27 | 31 | 35 |
|  | $MC\{O_{A3+}L_{A1+}^{con}\}$ | $MC\{O_{A3+}L_{A2+}^{con}\}$ | $MC\{O_{A3+}L_{A3+}^{con}\}$ | $MC\{O_{A3+}L_{A4+}^{con}\}$ | $MC\{O_{A3+}L_{A5+}^{con}\}$ |
| $KGE_{ss}^{worst}$ | 0.53 | 0.55 | 0.56 | 0.51 | 0.51 |
| $KGE_{ss}^{5\%}$ | 0.63 | 0.63 | 0.63 | 0.63 | 0.63 |
| $KGE_{ss}^{25\%}$ | 0.83 | 0.83 | 0.83 | 0.82 | 0.82 |
| $KGE_{ss}^{50\%}$ | 0.86 | 0.86 | 0.86 | 0.86 | 0.86 |
| $KGE_{ss}^{75\%}$ | 0.89 | 0.89 | 0.89 | 0.89 | 0.89 |
| $KGE_{ss}^{95\%}$ | 0.93 | 0.92 | 0.92 | 0.93 | 0.93 |
| Parameter No. | 23 | 27 | 31 | 35 | 39 |
|  | $MC\{O_{A4+}L_{A1+}^{con}\}$ | $MC\{O_{A4+}L_{A2+}^{con}\}$ | $MC\{O_{A4+}L_{A3+}^{con}\}$ | $MC\{O_{A4+}L_{A4+}^{con}\}$ | $MC\{O_{A4+}L_{A5+}^{con}\}$ |
| $KGE_{ss}^{worst}$ | 0.55 | 0.57 | 0.57 | 0.52 | 0.52 |
| $KGE_{ss}^{5\%}$ | 0.66 | 0.65 | 0.65 | 0.64 | 0.64 |
| $KGE_{ss}^{25\%}$ | 0.83 | 0.83 | 0.83 | 0.82 | 0.83 |
| $KGE_{ss}^{50\%}$ | 0.86 | 0.86 | 0.86 | 0.86 | 0.86 |
| $KGE_{ss}^{75\%}$ | 0.88 | 0.89 | 0.89 | 0.89 | 0.88 |
| $KGE_{ss}^{95\%}$ | 0.92 | 0.92 | 0.92 | 0.93 | 0.93 |
| Parameter No. | 27 | 31 | 35 | 39 | 43 |
|  | $MC\{O_{A5+}L_{A1+}^{con}\}$ | $MC\{O_{A5+}L_{A2+}^{con}\}$ | $MC\{O_{A5+}L_{A3+}^{con}\}$ | $MC\{O_{A5+}L_{A4+}^{con}\}$ | $MC\{O_{A5+}L_{A5+}^{con}\}$ |
| $KGE_{ss}^{worst}$ | 0.57 | 0.59 | 0.59 | 0.53 | 0.53 |
| $KGE_{ss}^{5\%}$ | 0.67 | 0.67 | 0.67 | 0.65 | 0.65 |
| $KGE_{ss}^{25\%}$ | 0.82 | 0.83 | 0.82 | 0.82 | 0.83 |
| $KGE_{ss}^{50\%}$ | 0.86 | 0.86 | 0.86 | 0.86 | 0.86 |
| $KGE_{ss}^{75\%}$ | 0.88 | 0.88 | 0.88 | 0.88 | 0.88 |
| $KGE_{ss}^{95\%}$ | 0.92 | 0.92 | 0.92 | 0.93 | 0.93 |
| Parameter No. | 31 | 35 | 39 | 43 | 47 |

Table S7: $KGE_{ss}$ Statistics and parameter numbers for single node MCP architecture with relaxation of mass-conserving constraint

|  | $MC\{O_\sigma L_\sigma^{con}\}$ | $MC\{O_\sigma L_\sigma^{con}\}M_\sigma^R$ | $MC\{O_\sigma L_\sigma^{con}\}M_I^R$ | $MC\{O_\sigma L_\sigma^{con}\}M_{\sigma r}^R$ | $MC\{O_\sigma L_\sigma^{con}\}M_{Ir}^R$ |
|---|---|---|---|---|---|
| $KGE_{ss}^{worst}$ | 0.30 | 0.60 | 0.51 | 0.46 | 0.48 |
| $KGE_{ss}^{5\%}$ | 0.48 | 0.64 | 0.56 | 0.52 | 0.57 |
| $KGE_{ss}^{25\%}$ | 0.78 | 0.79 | 0.78 | 0.76 | 0.78 |
| $KGE_{ss}^{median}$ | 0.84 | 0.85 | 0.84 | 0.83 | 0.85 |
| $KGE_{ss}^{75\%}$ | 0.87 | 0.87 | 0.86 | 0.87 | 0.86 |
| $KGE_{ss}^{95\%}$ | 0.91 | 0.91 | 0.91 | 0.91 | 0.91 |
| Parameter No. | 7 | 10 | 9 | 10 | 9 |

Table S8: $KGE_{ss}$ Statistics and parameter numbers for input bias-correction cases including $MC\{O_\sigma L_\sigma^{con} B_{LN}\}$ with N vary from 1 to 5.

|  | $MC\{O_\sigma L_\sigma^{con}\}B_{L1}$ | $MC\{O_\sigma L_\sigma^{con}\}B_{L2}$ | $MC\{O_\sigma L_\sigma^{con}\}B_{L3}$ | $MC\{O_\sigma L_\sigma^{con}\}B_{L4}$ | $MC\{O_\sigma L_\sigma^{con}\}B_{L5}$ |
|---|---|---|---|---|---|
| $KGE_{ss}^{worst}$ | 0.32 | 0.32 | 0.32 | 0.33 | 0.30 |
| $KGE_{ss}^{5\%}$ | 0.49 | 0.49 | 0.49 | 0.49 | 0.48 |
| $KGE_{ss}^{25\%}$ | 0.78 | 0.78 | 0.78 | 0.78 | 0.78 |
| $KGE_{ss}^{median}$ | 0.84 | 0.84 | 0.84 | 0.84 | 0.84 |
| $KGE_{ss}^{75\%}$ | 0.87 | 0.88 | 0.87 | 0.88 | 0.88 |
| $KGE_{ss}^{95\%}$ | 0.92 | 0.91 | 0.91 | 0.91 | 0.91 |
| Parameter No. | 9 | 11 | 13 | 15 | 17 |

Table S9: $KGE_{ss}$ Statistics and parameter numbers for input bias-correction cases including $MC\{O_\sigma L_\sigma^{con} B_{QN}\}$ with N vary from 1 to 5.

|  | $MC\{O_\sigma L_\sigma^{con} B_{Q1}\}$ | $MC\{O_\sigma L_\sigma^{con} B_{Q2}\}$ | $MC\{O_\sigma L_\sigma^{con} B_{Q3}\}$ | $MC\{O_\sigma L_\sigma^{con} B_{Q4}\}$ | $MC\{O_\sigma L_\sigma^{con} B_{Q5}\}$ |
|---|---|---|---|---|---|
| $KGE_{ss}^{worst}$ | 0.36 | 0.43 | 0.33 | 0.40 | 0.37 |
| $KGE_{ss}^{5\%}$ | 0.45 | 0.50 | 0.44 | 0.49 | 0.46 |
| $KGE_{ss}^{25\%}$ | 0.76 | 0.75 | 0.76 | 0.76 | 0.76 |
| $KGE_{ss}^{50\%}$ | 0.84 | 0.84 | 0.85 | 0.85 | 0.85 |
| $KGE_{ss}^{75\%}$ | 0.88 | 0.87 | 0.88 | 0.88 | 0.88 |
| $KGE_{ss}^{95\%}$ | 0.91 | 0.92 | 0.91 | 0.91 | 0.91 |
| Parameter No. | 10 | 12 | 14 | 16 | 18 |

Table S10: KGE$_{ss}$ Statistics and parameter numbers for single node MCP architecture, physics-based and several ML-based benchmarks.

|  | $MC\{O_\sigma L_\sigma^{con}\}$ | $MC\{O_{\sigma+} L_{\sigma+}^{con}\}$ | $MC\{O_\sigma L_\sigma^{con} M_\sigma^R\}$ | $MC\{O_{\sigma+} L_{\sigma+}^{con} M_\sigma^R\}$ | $ARX$ | $ANN(1)$ | $ANN(2)$ |
|---|---|---|---|---|---|---|---|
| $KGE_{ss}^{worst}$ | 0.30 | 0.57 | 0.60 | 0.56 | -1.22 | 0.10 | 0.17 |
| $KGE_{ss}^{5\%}$ | 0.47 | 0.70 | 0.64 | 0.67 | -0.98 | 0.21 | 0.28 |
| $KGE_{ss}^{25\%}$ | 0.78 | 0.81 | 0.78 | 0.78 | -0.17 | 0.53 | 0.58 |
| $KGE_{ss}^{median}$ | 0.84 | 0.84 | 0.85 | 0.85 | 0.16 | 0.70 | 0.71 |
| $KGE_{ss}^{75\%}$ | 0.87 | 0.87 | 0.88 | 0.88 | 0.30 | 0.76 | 0.77 |
| $KGE_{ss}^{95\%}$ | 0.92 | 0.91 | 0.91 | 0.91 | 0.42 | 0.82 | 0.83 |
| Parameter No. | 7 | 9 | 10 | 12 | 4 | 6 | 10 |
|  | $ANN(6)$ | $RNN(1)$ | $RNN(2)$ | $RNN(7)$ | $LSTM(1)$ | $LSTM(2)$ | $LSTM(5)$ |
| $KGE_{ss}^{worst}$ | 0.09 | 0.01 | 0.10 | 0.15 | 0.46 | 0.68 | 0.66 |
| $KGE_{ss}^{5\%}$ | 0.30 | 0.18 | 0.28 | 0.31 | 0.51 | 0.72 | 0.78 |
| $KGE_{ss}^{25\%}$ | 0.61 | 0.51 | 0.60 | 0.63 | 0.75 | 0.83 | 0.85 |
| $KGE_{ss}^{median}$ | 0.77 | 0.64 | 0.77 | 0.77 | 0.78 | 0.87 | 0.90 |
| $KGE_{ss}^{75\%}$ | 0.84 | 0.73 | 0.84 | 0.85 | 0.84 | 0.92 | 0.93 |
| $KGE_{ss}^{95\%}$ | 0.88 | 0.78 | 0.88 | 0.89 | 0.90 | 0.95 | 0.98 |
| Parameter No. | 26 | 4 | 10 | 30 | 16 | 43 | 166 |

Note: The statistics of percentile year are calculated based on total 38 years since the LSTM networks of the maximum sequence length used in the experiments equals to 390 days

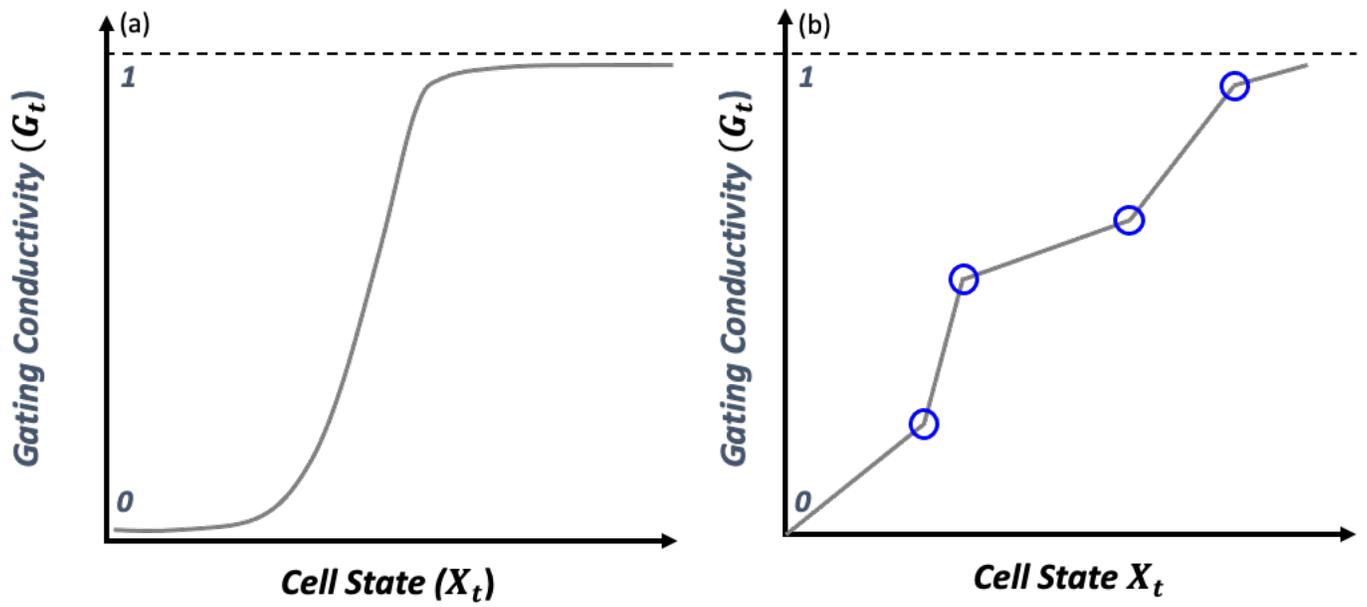

Figure S1: Illustration of a) simple monotonic-increasing gating function constructed using a single *Sigmoid* activation unit, and b) ANN-based piecewise-linear gating function constructed using multiple *ReLU* activation units in parallel.

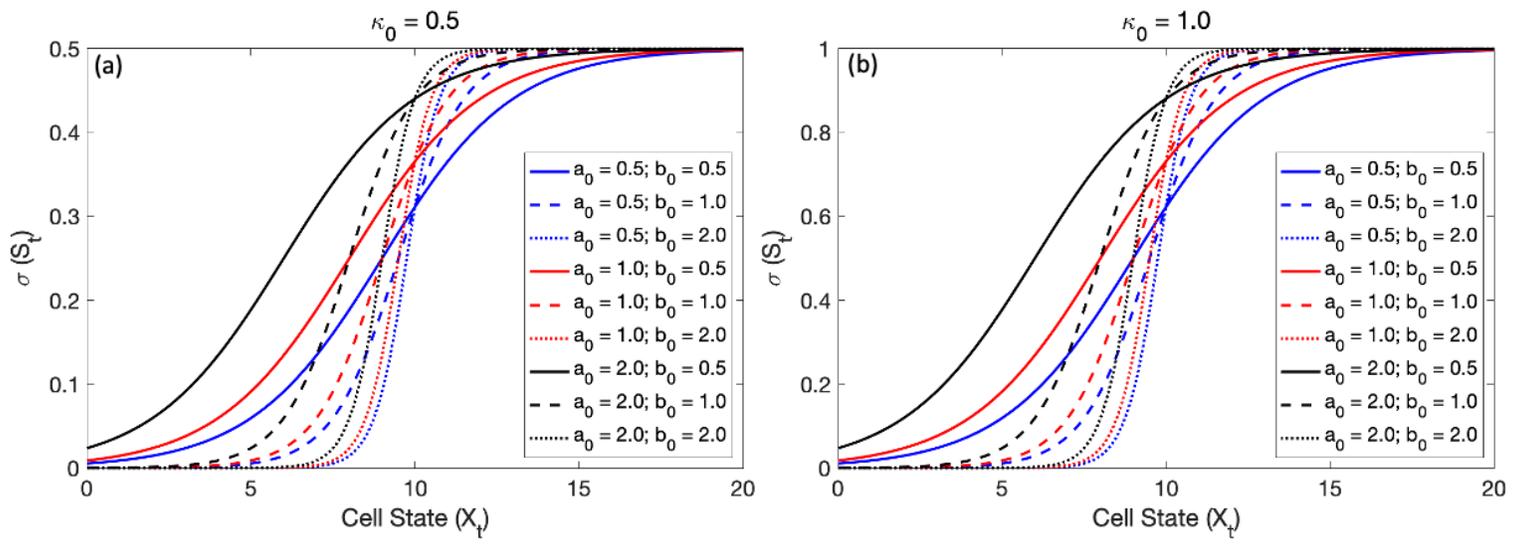

Figure S2: Illustration of the effect of parameter choices on the gating function shapes when using a Sigmoid-based activation function

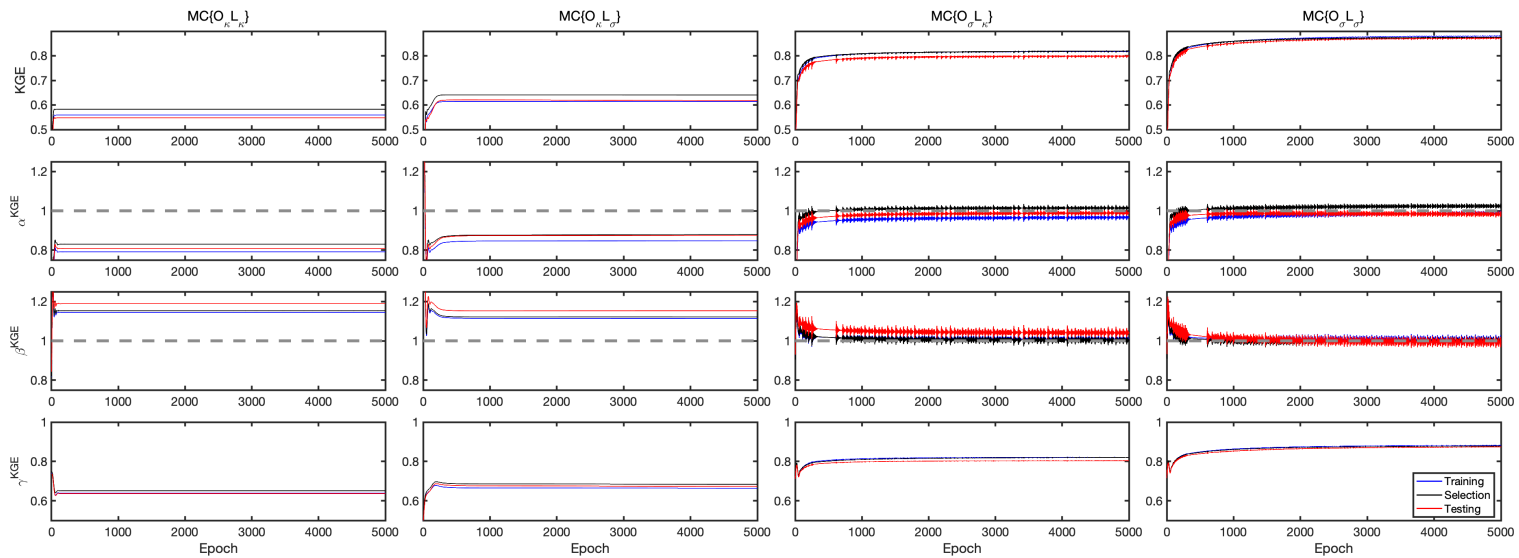

Figure S3: Evolution of training (blue), selection (black) and testing (red) period metrics with epoch for various MCP architectures

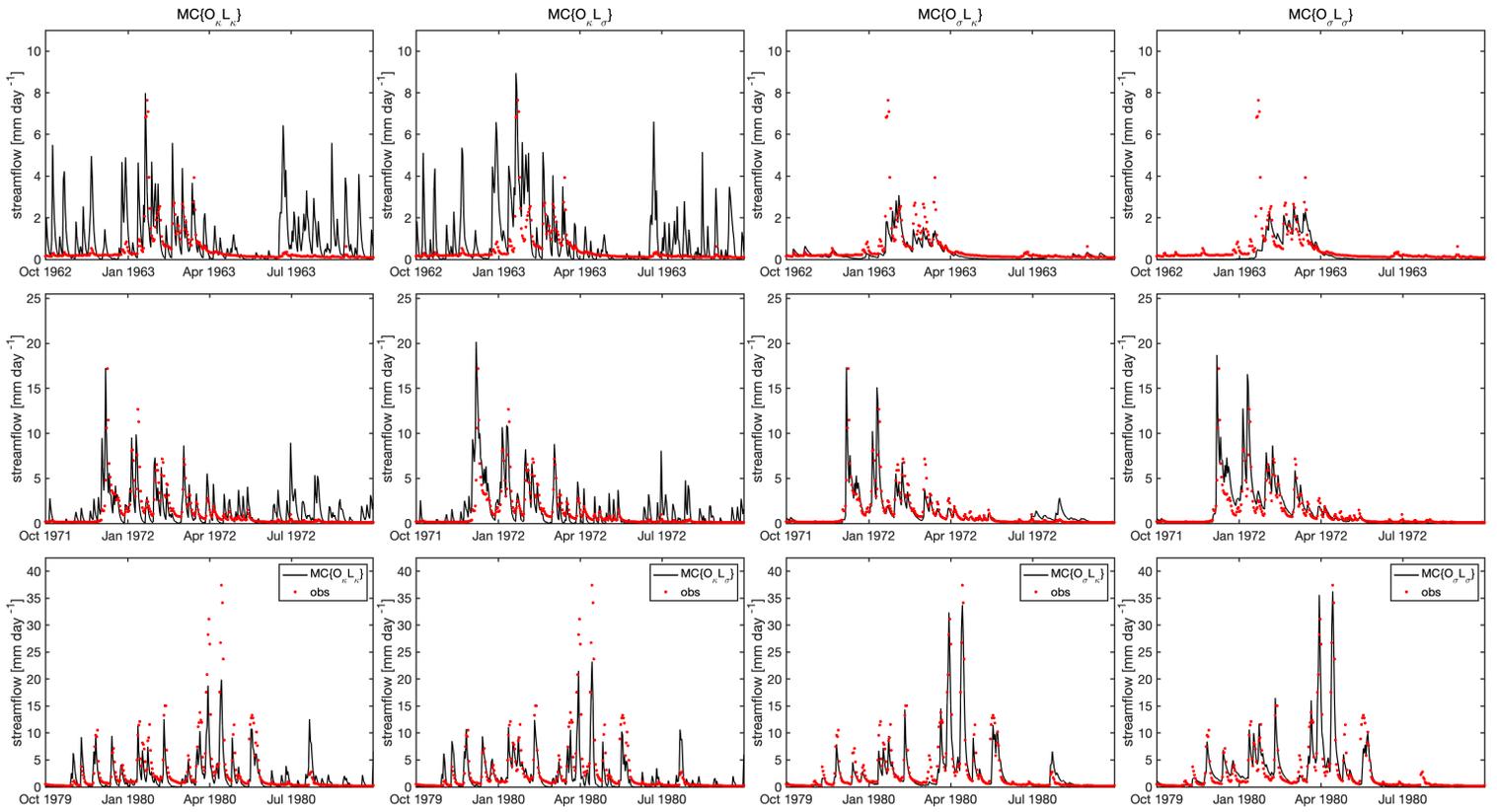

Figure S4: Observed (red dots) and model-simulated (solid black lines) hydrographs for three selected years. Top row shows a dry year (WY 1963), middle row shows a median year (WY 1972), and bottom row shows a wet year (WY 1980) based on annual flow volume. From left to right the columns show results for the $MC\{O_\kappa L_\kappa\}$, $MC\{O_\kappa L_\sigma\}$, $MC\{O_\sigma L_\kappa\}$ and $MC\{O_\sigma L_\sigma\}$ model architectures respectively.

Figure S5: Box and whisker plots of the distributions of annual $KGE_{ss}$ values for $MC\{O_\sigma L_\sigma^{con}\}$ single-node architectures using progressively more flexible (ANN-based) forms of the Output and Loss gating functions by varying N from 1 to 5.

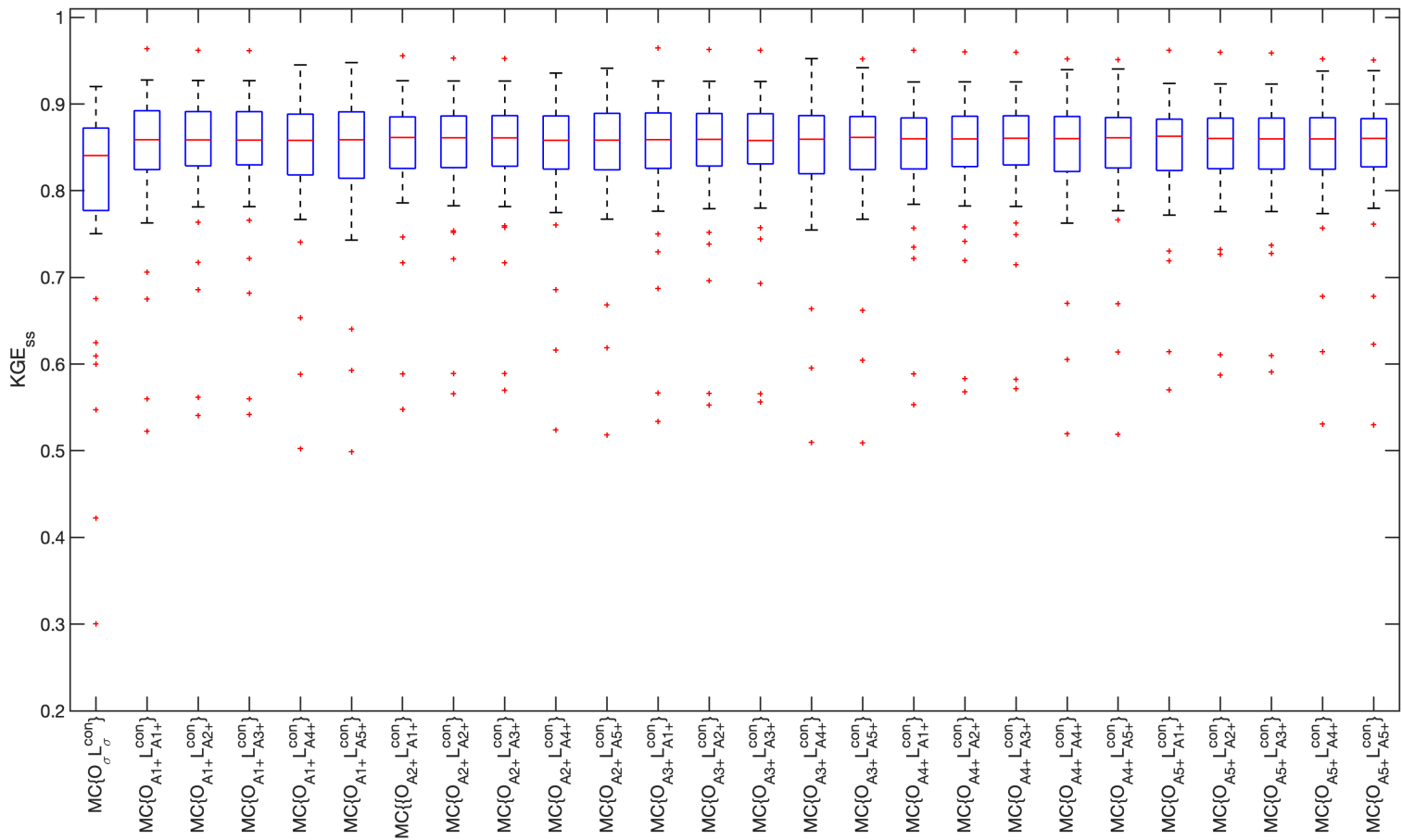

Figure S6: Box and whisker plots of the distributions of annual $KGE_{ss}$ values for $MC\{O_\sigma L_\sigma^{con}\}$ architecture using progressively more flexible (ANN-based) forms of the Output and Loss gating functions by varying N from 1 to 5 with multisource of contextual input in both output (cell state at present and previous timestep) and loss gates (cell state and PET at current time step).

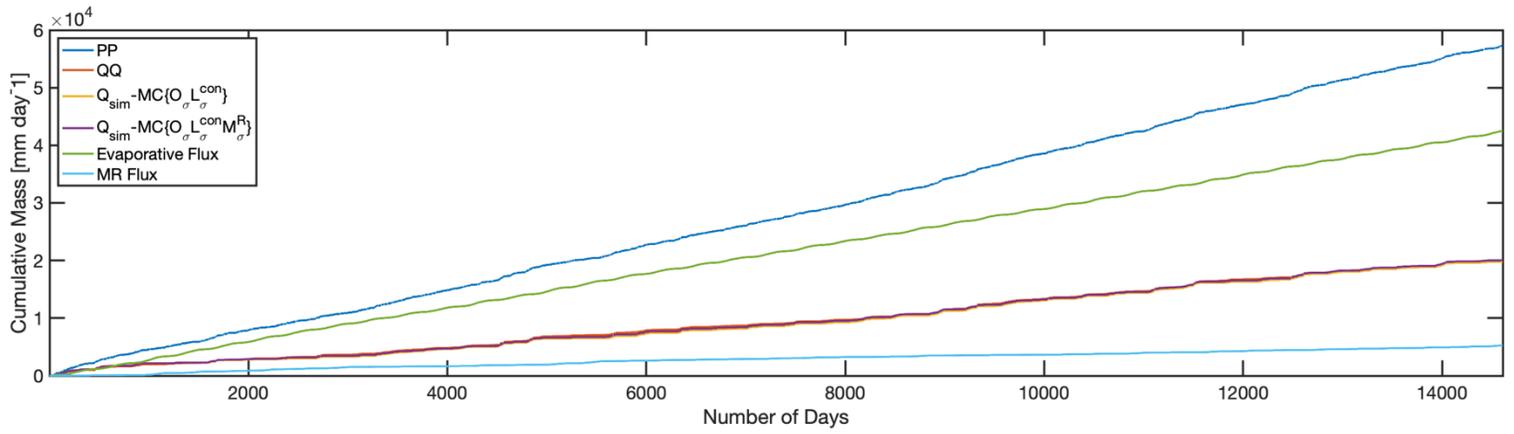

Figure S7: Cumulative flux for input precipitation, output discharge and the flux generated through loss gate, MR gate, and output gate based on $MC\{O_\sigma L_\sigma^{con} M_\sigma^R\}$ where the discharge is compared against the baseline $MC\{O_\sigma L_\sigma^{con}\}$ case.

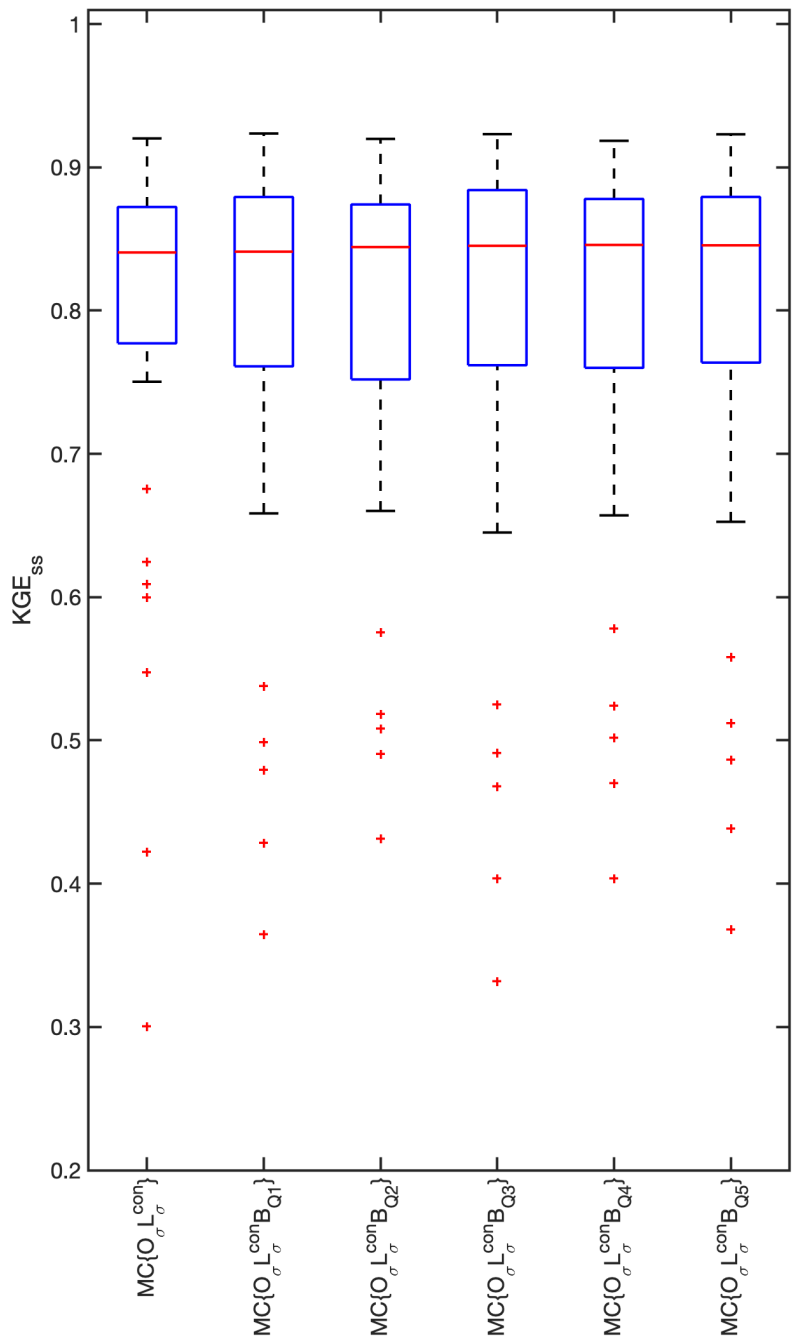

Figure S8: Box and whisker plots of the distributions of annual $KGE_{ss}$ values for various single-node ($MC$) architectures when piecewise-quadratic Input Bias Correction is enabled, indicated by the notation $B_{Qn}$ where subscript $n$ refers to number of piecewise linear sections.

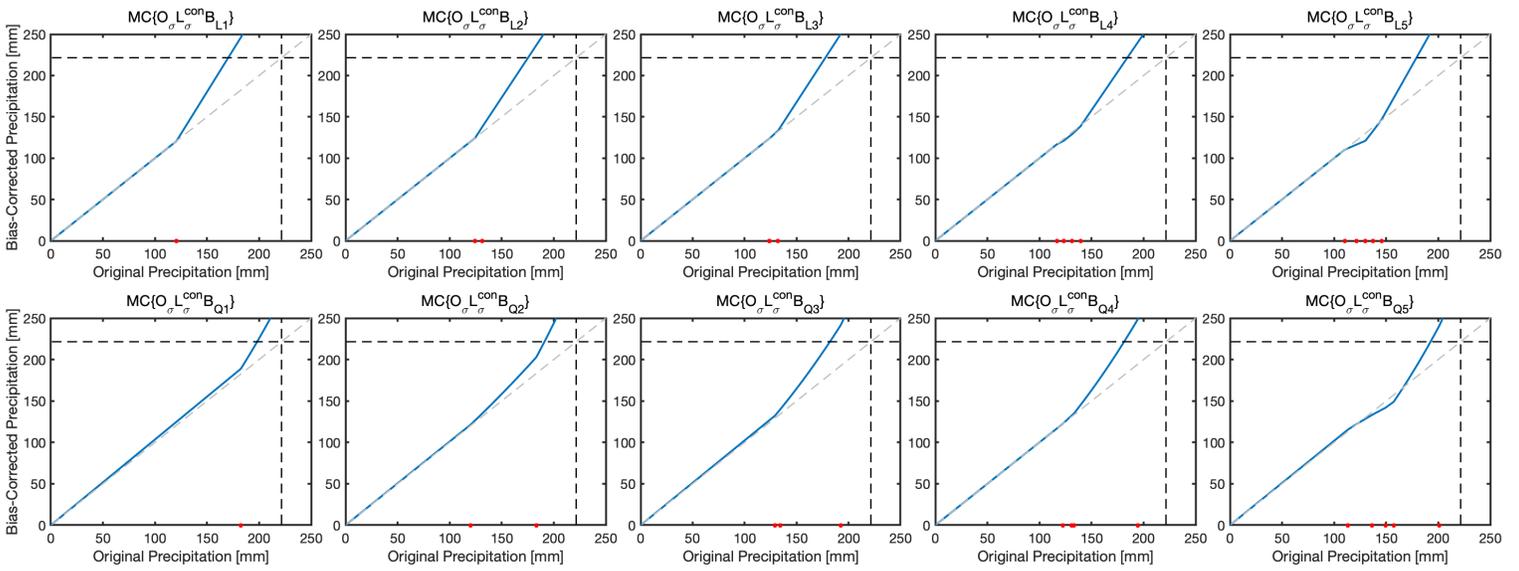

Figure S9: Plots showing the shapes of the learned Input Bias Correction functions for increasing numbers (from 1 to 5) of piecewise linear (top row) and quadratic (bottom row) segments. For reference, the grey dashed 1:1 line indicates no bias correction, and the horizontal/vertical dashed line indicates the observed maximum daily precipitation value over the 40-year period WY 1949 to 1988 – values larger extrapolate beyond the historical range and should be viewed with caution. In the top and bottom row, the red dots indicate the points where the piecewise linear ($B_{Ln}$) and quadratic functions ($B_{Qn}$) selected the break points for connecting two pieces of function

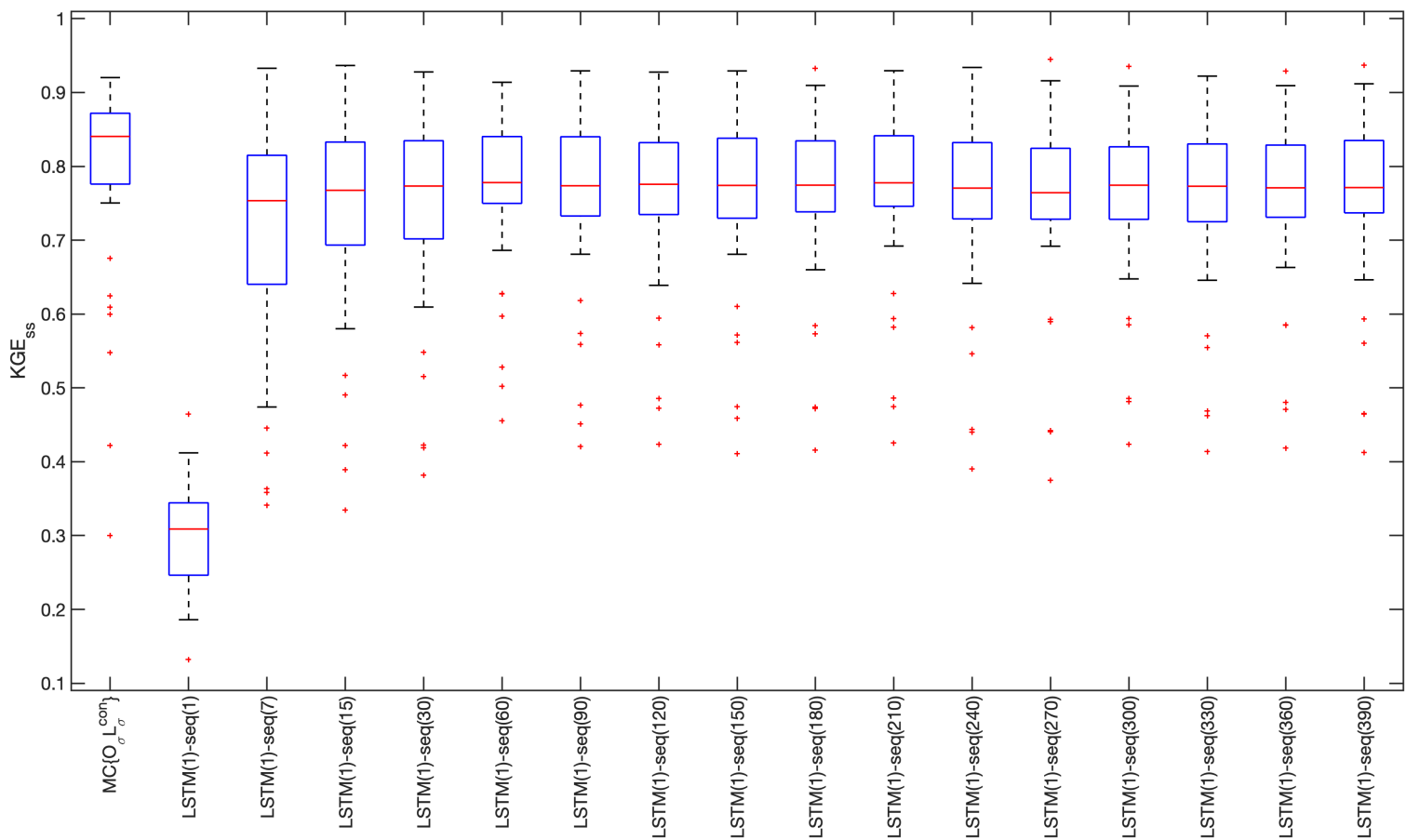

Figure S10: Box and whisker plots of the distributions of annual $KGE_{ss}$ values for the single node $MC\{O_\sigma L_\sigma^{con}\}$ and the long short-term memory network (LSTM) with different sequence length including 1, 7, 15, 30, 60, 90, 120, 150,180, 210, 240, 270, 300, 330, 360, and 390 days.